\documentclass{article}

\usepackage{microtype}
\usepackage{graphicx}
\usepackage{subcaption}
\usepackage{booktabs,multirow,makecell,array,longtable}
\usepackage{pifont}
\usepackage[nointegrals]{wasysym}
\usepackage[table]{xcolor}
\usepackage{float}          
\usepackage{placeins}       

\usepackage{hyperref}
\usepackage{xurl}           

\let\keepaddcontentsline\addcontentsline  
\usepackage[preprint]{icml2026}
\let\addcontentsline\keepaddcontentsline

\usepackage{amsmath,amssymb,mathtools,amsthm}
\usepackage{thm-restate}
\usepackage[capitalize,noabbrev]{cleveref}
\usepackage[most]{tcolorbox}
\usepackage{needspace}
\usepackage{algorithm,algorithmic}
\usepackage{tikz}
\usetikzlibrary{positioning,arrows.meta,calc,fit,backgrounds}
\usepackage{enumitem}
\setlist{nosep,leftmargin=*}

\definecolor{oursbg}{HTML}{E8F0FA}   
\definecolor{oursfg}{HTML}{2B6CB0}   
\definecolor{hdrbg}{HTML}{E1E7F0}    
\definecolor{negfg}{HTML}{B5473A}    
\definecolor{qcframe}{HTML}{C9D3DE}\definecolor{qcwrongbg}{HTML}{FDF1E4}  
\definecolor{qcwrongfg}{HTML}{C0712B}\definecolor{qcquotebg}{HTML}{F6F8FB}
\newcommand{\sig}[1]{\textcolor{oursfg}{\bfseries\boldmath #1}}     
\newcommand{\signeg}[1]{\textcolor{negfg}{\bfseries\boldmath #1}}   
\providecommand{\ci}[2]{{\scriptsize[#1, #2]}}                       
\newcommand{\grouprow}[3]{\multicolumn{#1}{c}{\rule{0pt}{3.4ex}\textbf{\textcolor{oursfg}{#2}}\ifx\relax#3\relax\else\ {\footnotesize(#3)}\fi}\\*[4pt]}
\newcommand{\fitwidth}[1]{\resizebox{\ifdim\width>\linewidth\linewidth\else\width\fi}{!}{#1}}  
\newcommand{\yes}{\textcolor{oursfg}{\ding{51}}}
\newcommand{\no}{\textcolor{naivefg}{\ding{55}}}
\newcommand{\partialicon}{\textcolor{basefg}{\LEFTcircle}}
\newcommand{\vh}[1]{\rotatebox{90}{\parbox{1.75cm}{\raggedright #1}}}  
\newcommand{\why}[1]{&&\text{\footnotesize\textcolor{naivefg}{[#1]}}}  
\definecolor{basefg}{HTML}{B06A1F}   
\definecolor{naivefg}{HTML}{6B7280}  
\definecolor{remfg}{HTML}{C0712B}
\newcommand{\inours}[1]{\textcolor{oursfg}{\textbf{#1}}}
\newcommand{\inbase}[1]{\textcolor{basefg}{\textbf{#1}}}
\newcommand{\innaive}[1]{\textcolor{naivefg}{\textbf{#1}}}
\newcommand{\hib}{{\scriptsize$\uparrow$}}
\newcommand{\lob}{{\scriptsize$\downarrow$}}

\definecolor{keybg}{HTML}{E4E8EE}
\newtcolorbox{resbox}{enhanced, breakable=false, colback=oursbg!45, colframe=oursfg!45, boxrule=0.5pt, arc=2pt,
  left=6pt, right=6pt, top=3pt, bottom=3pt, before skip=8pt, after skip=4pt}
\newtcolorbox{checkbox}{enhanced, breakable=false, colback=keybg!60, colframe=keybg, boxrule=0.4pt, arc=2pt,
  left=6pt, right=6pt, top=2pt, bottom=2pt, before skip=6pt, after skip=10pt, fontupper=\small}
\newtcolorbox{inshort}{enhanced, colback=oursbg!60, colframe=oursfg!40, boxrule=0.4pt, arc=2pt,
  left=5pt, right=5pt, top=3pt, bottom=3pt, fontupper=\small}
\newtcolorbox{examplebox}[1]{enhanced, breakable, colback=oursbg!45, colframe=oursfg!70, boxrule=0.5pt,
  arc=2pt, left=5pt, right=5pt, top=4pt, bottom=4pt, fonttitle=\bfseries\small, coltitle=oursfg,
  colbacktitle=oursbg, title={#1}}
\theoremstyle{plain}
\newtheorem{theorem}{Theorem}[section]
\newtheorem{proposition}[theorem]{Proposition}

\theoremstyle{definition}

\theoremstyle{remark}

\newcommand{\E}{\mathbb{E}}

\newcommand{\R}{\mathbb{R}}
\newcommand{\dW}{\Delta W}
\newcommand{\xb}{\bar x}
\newcommand{\Cset}{\mathcal{C}}
\newcommand{\Gr}{G_{\mathrm{rel}}}
\newcommand{\Go}{G_{\mathrm{off}}}
\newcommand{\Gm}{G_{\mathrm{rem}}}
\newcommand{\Ge}{G_{\mathrm{ext}}}
\newcommand{\Acc}{\mathrm{Acc}}
\newcommand{\bS}{\bar b^{S}}
\DeclareMathOperator{\corr}{corr}
\DeclareMathOperator{\Var}{Var}
\DeclareMathOperator{\Cov}{Cov}

\DeclareMathOperator*{\argmax}{arg\,max}

\newcommand{\methodname}[1]{#1}
\newcommand{\mSplit}{\methodname{Offset Split}}      
\newcommand{\mShared}{\methodname{Shared Offset}}    
\newcommand{\mBase}{\methodname{Base}}
\newcommand{\mRel}{\methodname{Release}}
\newcommand{\mOff}{\methodname{Offset}}
\newcommand{\mRem}{\methodname{Remainder}}
\newcommand{\mOwn}{\methodname{Own offset}}
\newcommand{\mLoo}{\methodname{Others' offset}}
\newcommand{\mPrior}{\methodname{Prior}}          
\newcommand{\mBias}{\methodname{Bias}}
\newcommand{\mPriorB}{\methodname{Prior-B}}
\newcommand{\mBiasB}{\methodname{Bias-B}}
\newcommand{\mLk}{\methodname{LoRA-8k}}
\newcommand{\mLv}{\methodname{LoRA-Varied}}
\newcommand{\mLc}{\methodname{LoRA-Clean}}
\newcommand{\mLx}{\methodname{LoRA-Exposed}}
\newcommand{\mLd}{\methodname{LoRA-Distill}}

\newtcolorbox{qualcard}[2]{enhanced, breakable=false, colback=white, colframe=qcframe, boxrule=0.6pt, arc=2pt,
  left=6pt, right=6pt, top=4pt, bottom=4pt, colbacktitle=hdrbg, coltitle=black, fonttitle=\small\bfseries,
  title={#1\hfill\normalfont\small #2}, before skip=8pt, after skip=8pt}
\newtcbox{\qbadge}[1][oursfg]{on line, colback=#1, colframe=#1, coltext=white, boxrule=0pt, arc=1.5pt,
  left=2.5pt, right=2.5pt, top=0.5pt, bottom=0.5pt, fontupper=\scriptsize\bfseries}
\newtcbox{\qok}{on line, colback=oursbg, colframe=oursfg, coltext=oursfg, boxrule=0.5pt, arc=1.5pt,
  left=2.5pt, right=2.5pt, top=0.5pt, bottom=0.5pt, fontupper=\footnotesize}
\newtcbox{\qbad}{on line, colback=qcwrongbg, colframe=qcwrongfg, coltext=qcwrongfg, boxrule=0.5pt, arc=1.5pt,
  left=2.5pt, right=2.5pt, top=0.5pt, bottom=0.5pt, fontupper=\footnotesize}

\newtcolorbox{excard}[4]{enhanced, breakable=false, colback=white, colframe=qcframe, boxrule=0.6pt, arc=2pt,
  left=6pt, right=6pt, top=5pt, bottom=5pt, colbacktitle=hdrbg, coltitle=black, fonttitle=\small\bfseries,
  title={\qbadge[#4]{#1}\enspace #3\hfill\normalfont\small #2}, before skip=9pt, after skip=9pt,
  sidebyside, sidebyside align=top, righthand width=5.3cm, sidebyside gap=10pt,
  segmentation style={solid, draw=qcframe, line width=0.4pt}, fontupper=\small, fontlower=\footnotesize}
\newcommand{\exrecord}[1]{\begin{tcolorbox}[enhanced, colback=qcquotebg, colframe=qcquotebg, boxrule=0pt, arc=1.5pt,
  left=4pt, right=4pt, top=2pt, bottom=2pt, nobeforeafter, fontupper=\footnotesize]\raggedright #1\end{tcolorbox}\par\vspace{3pt}}
\newcommand{\exgold}[1]{\textcolor{oursfg}{\textbf{#1}\,\ding{51}}}                 
\newcommand{\exhead}[1]{{\scriptsize\bfseries\color{black!60}\MakeUppercase{#1}}\par\vspace{1pt}}
\newcommand{\exbar}[2]{
  \begingroup\setlength{\fboxsep}{0pt}\colorbox{keybg}{\makebox[1.6cm][l]{\textcolor{oursfg}{\rule{\dimexpr 1.6cm*#1/#2\relax}{4pt}}}}\endgroup\ {\scriptsize #1/#2}}

\newtcolorbox{promptbox}[2]{enhanced, breakable, colback=white, colframe=qcframe, boxrule=0.6pt, arc=2pt,
  left=6pt, right=6pt, top=4pt, bottom=4pt, before skip=8pt, after skip=8pt,
  colbacktitle=hdrbg, coltitle=black, fonttitle=\small\bfseries, title={#1\hfill\normalfont\small #2},
  fontupper=\footnotesize, before upper={\raggedright\setlength{\parskip}{3pt}\setlength{\parindent}{0pt}}}
\tcbset{pb gray/.style={colback=qcquotebg, before upper={\raggedright{\scriptsize\bfseries\color{black!65}\pbtitle}\par\setlength{\parskip}{1pt}}},
  pb oursfg/.style={colback=oursbg!80, before upper={\raggedright{\scriptsize\bfseries\color{oursfg}\pbtitle}\par\setlength{\parskip}{1pt}}}}
\newtcolorbox{pblock}[2][gray]{enhanced, colframe=white, boxrule=0pt, arc=1.5pt, left=4pt, right=4pt, top=2pt,
  bottom=2pt, before skip=3pt, after skip=3pt, fontupper=\footnotesize, code={\def\pbtitle{#2}}, pb #1}
\newtcbox{\pslot}{on line, colback=oursbg, colframe=oursfg, coltext=oursfg, boxrule=0.5pt, arc=1.5pt,
  left=2pt, right=2pt, top=0.3pt, bottom=0.3pt, fontupper=\scriptsize\itshape}   
\newcommand{\ptag}[1]{{\ttfamily\scriptsize\bfseries\color{black!65}#1}}     

\newcommand{\pverdict}[1]{\par\smallskip{\footnotesize\textbf{Verdict format.}}\enspace{\ttfamily\scriptsize #1}}

\icmltitlerunning{One Offset, Many Models}

\begin{document}

\twocolumn[
  \icmltitle{One Offset, Many Models: A Weight-Space Audit of Decision Models}
  \icmlsetsymbol{equal}{*}
  \begin{icmlauthorlist}
    \icmlauthor{Quy-Anh Dang}{knovel}
    \icmlauthor{Chris Ngo}{knovel}
  \end{icmlauthorlist}
  \icmlaffiliation{knovel}{Knovel Engineering Lab, Singapore}
  \icmlcorrespondingauthor{Quy-Anh Dang}{quyanh.dang@knoveleng.com}
  \icmlcorrespondingauthor{Chris Ngo}{chris.ngo@knoveleng.com}
  \icmlkeywords{fine-tuning, model diffing, steering vectors, task vectors, evaluation}
  \begin{center}\small
  \href{https://github.com/knoveleng/one-offset}{\raisebox{-0.15em}{\includegraphics[height=1.05em]{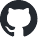}}\,\texttt{github.com/knoveleng/one-offset}}
  \hspace{2.5em}
  \href{https://knoveleng.github.io/one-offset/}{\raisebox{-0.15em}{\includegraphics[height=1.05em]{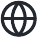}}\,\texttt{knoveleng.github.io/one-offset}}
  \end{center}
  \vskip 0.15in
]
\raggedbottom
\begin{NoHyper}
\printAffiliationsAndNotice{}  
\end{NoHyper}

\begin{abstract}
Dozens of groups release small ``decision models'': open language models fine-tuned to answer a question about a
record with one of a few labels. We ask what these separately trained releases have in common. Our \mSplit{}
replaces a release's weight update at every changed matrix by its mean effect, a constant offset that cannot depend
on the prompt, and keeps the input-dependent remainder apart. We audit 31 releases from 26 groups on three bases
(Qwen3.5-4B, Qwen3.5-2B, Gemma-4-E4B). The offset alone reaches a plateau: +4.5 to +10.7 points on the Decision
Index in 20 of the 22 releases that gain 8 to 25 points. It is sufficient, not necessary: without its offset, a
release keeps a median 94\% of its gain. Offsets of different groups point the same way, help the same benchmarks
and can be exchanged. Their average, a \mShared{} added to an untouched base, gains +8.8 to +10.5 points with no
training, where a fitted bias on the label logits gains at most +1.2. Fine-tunes for other purposes lack it, and our
label-only adapters reach it and little more. What the strongest releases add is input-dependent and specific to
each model.
\end{abstract}

\begin{figure}[t]
\centering
\includegraphics[width=\columnwidth]{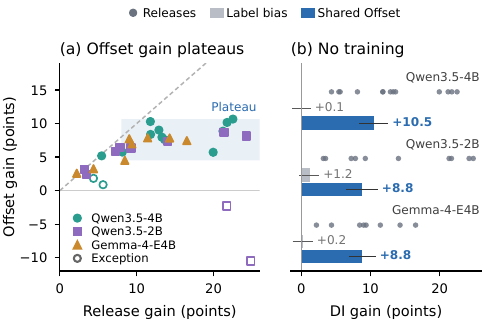}
\caption{\textbf{Released decision models share a constant offset that needs no training.} (a) Gain of each of
31 releases when its weight update is replaced by a constant (its \mOff), against the gain of the release (dashed:
equal; colour is the base in this panel only; hollow: the four exceptions). Above 8 points the offset gain stays in a
band. (b) Per base: gains of the releases, of a \innaive{bias} fitted on the base's label logits (\mBias), and of
the \inours{\mShared} added to the untouched base, with 95\% intervals. Values:
\cref{tab:summary,tab:releases,tab:calibration}.}
\label{fig:teaser}
\end{figure}

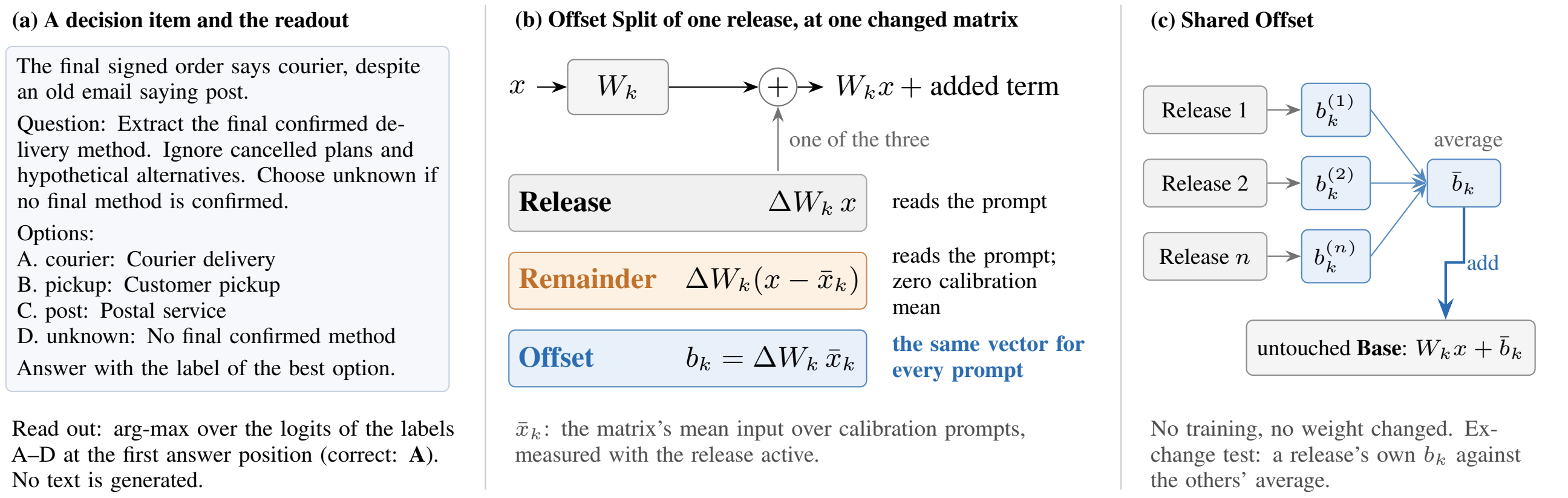
\begin{figure*}[t]
\centering
\begin{tikzpicture}[font=\footnotesize, >=Stealth,
  box/.style={draw=black!45, rounded corners=2pt, minimum height=6mm, minimum width=11mm, fill=black!4},
  opt/.style={rounded corners=2pt, minimum height=6.2mm, text width=37mm, inner sep=3pt, align=left},
  tag/.style={font=\scriptsize, text width=21mm, align=left},
  rel/.style={draw=black!40, rounded corners=2pt, fill=black!5, minimum height=5.2mm, minimum width=13.5mm, font=\scriptsize},
  vec/.style={draw=oursfg!80, rounded corners=2pt, fill=oursbg, minimum height=5.2mm, minimum width=7.5mm, font=\scriptsize},
  ttl/.style={anchor=west, font=\scriptsize\bfseries}]
  \node[ttl] at (0,0.62) {(a) A decision item and the readout};
  \node[draw=qcframe, fill=qcquotebg, rounded corners=2pt, anchor=north west, text width=46mm, inner sep=3.5pt,
        font=\scriptsize, align=left] (item) at (0.05,0.35)
       {The final signed order says courier, despite an old email saying post.\\[2pt]
        Question: Extract the final confirmed delivery method. Ignore cancelled plans and hypothetical
        alternatives. Choose unknown if no final method is confirmed.\\[2pt]
        Options:\\
        A. courier: Courier delivery\\
        B. pickup: Customer pickup\\
        C. post: Postal service\\
        D. unknown: No final confirmed method\\[2pt]
        Answer with the label of the best option.};
  \node[anchor=north west, font=\scriptsize, text width=49mm, align=left] at (0.0,-3.62)
       {Read out: arg-max over the logits of the labels A--D at the first answer position
        (correct: \textbf{A}). No text is generated.};
  \draw[black!20] (5.3,0.8) -- (5.3,-4.5);
  \begin{scope}[shift={(5.5,0)}]
  \node[ttl] at (0,0.62) {(b) Offset Split of one release, at one changed matrix};
  \node (x) at (0.15,-0.1) {$x$};
  \node[box] (W) at (1.25,-0.1) {$W_k$};
  \node[draw=black!55, circle, inner sep=0.8pt] (plus) at (3.0,-0.1) {$+$};
  \node[anchor=west] (y) at (3.5,-0.1) {$W_kx\,+$ added term};
  \draw[->] (x) -- (W); \draw[->] (W) -- (plus); \draw[->] (plus) -- (y);
  \node[opt, fill=black!6, draw=black!35, anchor=north west] (rel) at (0.05,-1.05)
       {\textbf{\mRel}\hfill $\dW_k\,x$};
  \node[opt, fill=qcwrongbg, draw=qcwrongfg!80, anchor=north west] (rem) at (0.05,-1.9)
       {\textbf{\textcolor{qcwrongfg}{\mRem}}\hfill $\dW_k(x-\xb_k)$};
  \node[opt, fill=oursbg, draw=oursfg!80, anchor=north west] (off) at (0.05,-2.75)
       {\textbf{\textcolor{oursfg}{\mOff}}\hfill $b_k=\dW_k\,\xb_k$};
  \node[tag, anchor=west] at ($(rel.east)+(1.5mm,0)$) {reads the prompt};
  \node[tag, anchor=west] at ($(rem.east)+(1.5mm,0)$) {reads the prompt; zero calibration mean};
  \node[tag, anchor=west, text=oursfg] at ($(off.east)+(1.5mm,0)$) {\textbf{the same vector for every prompt}};
  \draw[->, black!55] (3.0,-1.02) -- (plus.south) node[midway, right, font=\scriptsize, text=black!60] {one of the three};
  \node[anchor=north west, font=\scriptsize, text width=58mm, align=left, text=black!70] at (0.0,-3.62)
       {$\xb_k$: the matrix's mean input over calibration prompts, measured with the release active.};
  \end{scope}
  \draw[black!20] (12.25,0.8) -- (12.25,-4.5);
  \begin{scope}[shift={(12.45,0)}]
  \node[ttl] at (0,0.62) {(c) Shared Offset};
  \node[rel] (r1) at (0.72,-0.35) {Release 1};
  \node[rel] (r2) at (0.72,-1.15) {Release 2};
  \node[rel] (r3) at (0.72,-1.95) {Release $n$};
  \node[vec] (b1) at (2.15,-0.35) {$b^{(1)}_k$};
  \node[vec] (b2) at (2.15,-1.15) {$b^{(2)}_k$};
  \node[vec] (b3) at (2.15,-1.95) {$b^{(n)}_k$};
  \foreach \i in {1,2,3} {\draw[->, black!55] (r\i) -- (b\i);}
  \node[vec, minimum width=8mm] (avg) at (3.55,-1.15) {$\bar b_k$};
  \foreach \i in {1,2,3} {\draw[->, oursfg!80] (b\i.east) -- (avg.west);}
  \node[font=\scriptsize, text=black!60] at (3.6,-0.72) {average};
  \node[box, minimum width=30mm, align=center, font=\scriptsize] (base) at (2.75,-2.95)
       {untouched \textbf{\mBase}: $W_kx+\bar b_k$};
  \draw[->, oursfg, thick] (avg.south) -- ++(0,-0.6) -| ([xshift=6mm]base.north)
       node[pos=0.25, right, font=\scriptsize, text=oursfg] {add};
  \node[anchor=north west, font=\scriptsize, text width=43mm, align=left, text=black!70] at (0.0,-3.62)
       {No training, no weight changed. Exchange test: a release's own $b_k$ against the others' average.};
  \end{scope}
\end{tikzpicture}
\caption{\textbf{The \mSplit{} and the \mShared.} (a) A decision item as every model sees it, and the readout.
(b) A changed matrix adds $\dW_k x$ to its output. The \inbase{\mRem{}} model adds only the input-dependent part,
$\dW_k(x-\xb^{\mathrm{all}}_k)$; the \inours{\mOff{}} model adds only the constant $b_k=\dW_k\xb_k$. (c) Offsets of releases on one
base live in the same space; the \mShared{} adds their average to the untouched base.}
\label{fig:schematic}
\end{figure*}

\section{Introduction}\label{sec:intro}

Small open language models are now routinely fine-tuned into \emph{decision models}: given a record (a ticket, a
contract clause, a tool call) and a typed question, they return one of a few labels, to route a request, flag a
claim or pick a tool. Dozens of groups publish such models on the same open bases, each with its own data and
recipe and a benchmark number. What the tuning installed is not described. A user choosing among them, or deciding
whether to fine-tune at all, cannot tell whether two releases learned different skills or the same shift.

Benchmark scores cannot answer this, and the usual weight-space tools answer a different question. Low-rank
analyses count how many directions an update writes \citep{aghajanyan2020intrinsic,hu2022lora}; task arithmetic
adds and averages whole updates \citep{ilharco2023task}; steering methods learn or derive vectors on one model
\citep{turner2023actadd,panickssery2023caa}. None separates the part of an update that depends on the prompt from the
part that does not, and none compares that part across models trained by different groups for the same purpose.

The question can be asked directly in the weights. A changed matrix adds $\dW x$ to its output, a function of its
input $x$. Replace $x$ by its average over many prompts and the addition becomes one constant vector per matrix: a
steering vector, read off the released weights instead of learned. If a release keeps a gain after this replacement,
that gain does not need the update to read the prompt. We call this the \emph{\mSplit} (\cref{fig:schematic}): every release yields an
\mOff{} model (the constant part only) and a \mRem{} model (the input-dependent rest). Because offsets of releases
on one base live in the same space, they can also be averaged and exchanged; the \emph{\mShared} is the average
offset of several groups' releases, added to an untouched base.

We audit 31 releases from 26 groups on three bases (Qwen3.5-4B, Qwen3.5-2B, Gemma-4-E4B), with twelve fine-tunes
built for other purposes as controls and five adapters of our own, on 36 benchmarks of the Decision Index and on
JevBench. Our contribution is the split itself, a training-free and exact decomposition of any released update
(\cref{sec:method}), and the answers it gives to four questions:
\begin{itemize}[leftmargin=1.1em,itemsep=2pt]
\item \textbf{RQ1: How much of a release's gain can a constant reach, and does the release need it?} A plateau: in 20 of
the 22 releases that gain 8 to 25 points, the offset alone gives \sig{+4.5 to +10.7} points (\cref{fig:teaser}a).
Yet with its offset removed, a release keeps a median \sig{94\%} of its gain: the constant is sufficient for this
part, not necessary (\cref{sec:rq1}).
\item \textbf{RQ2: Do releases of different groups share one offset?} Yes. Offsets can be exchanged between releases, and
the \mShared{} adds \sig{+8.8 to +10.5} points to the untouched base with no training, against at most $+1.2$ for
a bias fitted on the label logits (\cref{fig:teaser}b, \cref{sec:rq2}).
\item \textbf{RQ3: Where does the offset come from?} From decision tuning. Offsets of other fine-tunes give at most
\sig{+2.0} points, and our label-only adapters stay at $+9.0$ to $+11.3$, on the plateau (\cref{sec:rq3}).
\item \textbf{RQ4: What do the strongest releases add?} An input-dependent part, different in each release, worth $+11.6$
to $+16.1$ points beyond their offset (\cref{sec:rq4}).
\end{itemize}

\section{Related Work}\label{sec:related}

\textbf{Parameter-efficient and bias-only adaptation.} Fine-tuning updates are known to be low-dimensional
\citep{aghajanyan2020intrinsic}, and low-rank adapters \citep{hu2022lora}, even of rank one
\citep{schulman2025lora}, match full fine-tuning in many settings. Tuning bias terms only \citep{benzaken2022bitfit},
one learned additive vector per layer \citep{sinii2025bias}, learned prompts or activation scalings
\citep{lester2021prompt,liu2022ia3}, or interventions on hidden representations
\citep{wu2024reft,yin2024lofit,wu2025improved} can come close to full tuning. These methods train a constrained model
and show what it can reach; they do not tell what an unconstrained release that someone else trained actually
uses.

\textbf{Activation steering and its transfer.} Adding a fixed vector to the residual stream changes behaviour, with
vectors obtained from contrast pairs, learned, or read from in-context examples
\citep{subramani2022extracting,turner2023actadd,panickssery2023caa,li2023iti,zou2023repe,arditi2024refusal,
hendel2023task,todd2024function,stickland2024steering}. Vectors can carry over between a base and its own fine-tune
\citep{stolfo2025instruction,cao2024bipo}, and directions derived from a base and from a reasoning fine-tune agree
\citep{ward2025reasoning}; weight differences can also steer \citep{fierro2026weight}. Each vector comes from a model
pair the authors built or chose, for one behaviour, and is not tested against models trained by other groups.

\textbf{Task arithmetic and model merging.} Fine-tuned minus base weights form task vectors that can be added,
negated and averaged \citep{ilharco2023task,wortsman2022soups,matena2022fisher,yadav2023ties,ortiz2023tangent,
song2026merging}, and task vectors of different tasks split into shared and task-specific parts
\citep{damirchi2025decomposing,mcclendon2026merging,kawamoto2025crossmodel}. These methods operate on whole updates and
on different tasks; the shared part is a geometric property of the weights, not a functional one tested by
exchanging it between models.

\begin{table}[t]
\centering
\small
\setlength{\tabcolsep}{2.2pt}
\caption{\textbf{Positioning.} What each line of work examines. \yes{}: yes; \partialicon{}: in part; \no{}: no.
\emph{Exchange test}: sameness is tested by moving a part from one model to another.}
\label{tab:positioning}
\begin{tabular}{@{}lcccccc@{}}
\toprule
\rowcolor{hdrbg} Line of work & \vh{Third-party weights} & \vh{Many releases, one purpose} & \vh{Prompt-independent part} & \vh{Exchange test} & \vh{Other-purpose controls} & \vh{No training}\\
\midrule
Bias-only adaptation & \no & \no & \yes & \no & \no & \no\\
Activation steering & \no & \no & \yes & \no & \partialicon & \partialicon\\
Steering transfer & \partialicon & \no & \yes & \partialicon & \no & \partialicon\\
Task arithmetic, merging & \yes & \partialicon & \no & \no & \no & \yes\\
Model diffing & \yes & \no & \no & \no & \partialicon & \no\\
\midrule
\rowcolor{oursbg} \textbf{Our work} & \yes & \yes & \yes & \yes & \yes & \yes\\
\bottomrule
\end{tabular}
\end{table}

\textbf{What fine-tuning changes.} Mechanistic studies find that fine-tuning often leaves a model's computation in
place and adds a thin layer on top of it or sharpens existing circuits
\citep{jain2024mechanistically,prakash2024finetuning}, and that separately trained networks converge on similar
representations \citep{huh2024platonic}. Crosscoders compare a base with one fine-tune through a learned dictionary
\citep{minder2025crosscoders,kassem2026delta}. These studies examine one fine-tune, usually the authors' own, and
need trained probes or dictionaries.

\textbf{Calibration and evaluation hygiene.} Answers to multiple-choice prompts depend on option order and label
priors \citep{zhao2021calibrate,holtzman2021surface,lu2022fantastically,robinson2023leveraging,zheng2024selectors,
pezeshkpour2023order}, which output-level corrections reduce without touching the model
\citep{zhao2021calibrate,zhou2024batch}; benchmark text also leaks into training corpora
\citep{sainz2023contamination,golchin2024time,oren2023proving}. Both effects can masquerade as skill in an audit, and
a constant offset could in principle be no more than such a correction; we test this in \cref{sec:rq2}.

\textbf{Our approach in context.} We extract the input-independent part of released weights, for any update, with no
training and no contrast data, and test it the way merging and steering work does not: across 31 releases from 26
groups tuned for the same purpose, by exchanging it between them, against output-level corrections, against
non-decision fine-tunes of the same bases, and against our own controlled tuning (\cref{tab:positioning}).


\section{The Offset Split}\label{sec:method}

\subsection{Setting}\label{sec:setting}

\textbf{Decision items and the readout.} An item is a record $s$, a question $q$ and options $o_1,\dots,o_K$ with
one correct option $y$ (\cref{fig:schematic}a shows one from JevBench).
Every model sees the same prompt (record, question, lettered options and a fixed instruction,
\cref{app:prompts}) and answers with the label whose token has the highest logit at the first answer position,
$\hat y=\argmax_{k\le K}\langle h(s,q,o), e_{\ell_k}\rangle$, where $h$ is the final hidden state and $e_{\ell_k}$
the output embedding of label $\ell_k$ (``A'', ``B'', \dots). No text is generated, so every model is read the same
way, whatever format its authors trained it to emit.

\textbf{Releases and updates.} A \emph{release} is a decision model published with its weights, fine-tuned from a
public \emph{base}. Each matrix $W_k$ of the base (attention, linear-attention and MLP projections) becomes
$W'_k=W_k+\dW_k$ in the release. A LoRA adapter merged into the weights is a $\dW_k$ of low rank, a full fine-tune
a dense one; the split below treats both the same way. Dense fine-tunes also change norm gains and embeddings; we
keep these at the release's values in every variant and measure their effect separately (it is small,
\cref{sec:ablations}).

\subsection{Splitting an update}\label{sec:split}

At every position $t$ of a prompt, a changed matrix computes $W'_k x_{k,t} = W_k x_{k,t} + \dW_k x_{k,t}$: the
release adds a vector that depends on the input $x_{k,t}$. Replacing $x_{k,t}$ by a fixed vector removes that
dependence. We take the mean input of the matrix over the tokens of a calibration set $\Cset$ of 300 public training
prompts (never evaluation items), measured with the release active, because these are the inputs at which the
release's update actually operates. The answer is read at the last position, whose inputs differ systematically, so
we use two means, over last positions $T_c$ and over the others:
\begin{equation}
\xb^{\mathrm{last}}_k=\frac{1}{|\Cset|}\sum_{c} x^{(c)}_{k,T_c},\quad
\xb^{\mathrm{rest}}_k=\frac{1}{N_{\Cset}-|\Cset|}\sum_{c}\sum_{t<T_c} x^{(c)}_{k,t}.
\label{eq:offset}
\end{equation}
The \emph{offset} of matrix $k$ is $b_k=\dW_k\xb_k$, with $\xb_k$ the first mean at the answer position and the
second elsewhere; $\xb^{\mathrm{all}}_k$ is one mean over all tokens. The split defines four models that differ only in what is added to $W_k x$ at each changed matrix:
\begin{center}
\small
\begin{tabular}{@{}l@{\hspace{6pt}}l@{\hspace{6pt}}c@{}}
\toprule
Model & Added to $W_kx$ & Reads the prompt\\
\midrule
\mBase & $0$ & --\\
\mRel & $\dW_k x$ & yes\\
\mRem & $\dW_k (x-\xb^{\mathrm{all}}_k)$ & yes\\
\rowcolor{oursbg}\mOff & $b_k$ & no\\
\bottomrule
\end{tabular}
\end{center}
With $\Acc$ the accuracy averaged over benchmarks, each model has a gain over the base: the \emph{release gain}
$\Gr=\Acc_{\mRel}-\Acc_{\mBase}$, the \emph{offset gain} $\Go=\Acc_{\mOff}-\Acc_{\mBase}$ and the \emph{remainder
gain} $\Gm=\Acc_{\mRem}-\Acc_{\mBase}$. Two derived quantities are used throughout: the \emph{offset share}
$\sigma=\Go/\Gr$, and the \emph{extra gain} $\Ge=\Gr-\Go$, the part of the release gain that its offset does not
reach. Splitting one release costs one pass over the calibration prompts and one evaluation pass per model; nothing is
trained.

\subsection{What the split says, and what it does not}\label{sec:props}

Three elementary properties fix how every experiment is read. Proofs are in \cref{app:theory}.

\begin{restatable}[Properties of the split]{proposition}{propsplit}\label{prop:exact}
\textup{(a) Exact.} For every matrix $k$ and input $x$, $\dW_k x=b_k+\dW_k(x-\xb_k)$, and the second term averages
to zero over the calibration tokens. \textup{(b) Best constant.} Among all constant vectors, $b_k$ is the one closest
to the release's added term $\dW_k x$ in mean squared error over the calibration tokens. \textup{(c) No item
information.} The \mOff{} model adds the same vectors to every prompt, chosen by position index only; any
difference between its answers on two items is computed by the base's own weights.
\end{restatable}

By (c), an offset gain cannot come from the update recognising anything in the item: not a memorised test item,
not a feature of the record. It does \emph{not} follow that the offset is a cosmetic correction of the output. A
constant passes through every nonlinearity after it and can change what the base computes, which is why bias-only
tuning can approach full tuning \citep{benzaken2022bitfit}; whether the offset amounts to an output correction is an
empirical question (\cref{sec:rq2}). The split also differs from rank. A rank-1 update $\dW_k=u v^\top$ adds
$u\,(v^\top x)$, a direction times a scalar that \emph{depends on the input}; rank counts the directions an update
writes, the split asks whether what it writes depends on what it reads.

\textbf{Two counterfactuals, not two addends.} A transformer is not linear and accuracy is a step function of the
logits, so the gains of the two parts add up to the release gain only if the logits are affine in the added terms
(\cref{prop:additive}). We therefore read the split through separate questions, each with its own quantity:
\begin{center}
\small
\begin{tabular}{@{}l@{\hspace{3pt}}l@{\hspace{3pt}}l@{}}
\toprule
Quantity & Question & If large\\
\midrule
$\Go$ & Sufficiency: what does a & a constant reaches\\
 & constant reach? & the gain\\
$\Gr-\Gm$ & Necessity: what is lost & the release relies\\
 & without the constant? & on its constant\\
$\Ge=\Gr-\Go$ & What is out of reach & much of the gain\\
 & of the constant? & needs the prompt\\
\bottomrule
\end{tabular}
\end{center}
Sufficiency and necessity can disagree, and in our audit they do (\cref{sec:rq1}): a gain can be reachable by the
constant and still not depend on it. The extra gain $\Ge$ is a lower bound on what needs the prompt, not the size of
a separate mechanism.

\subsection{Comparing the offsets of different releases}\label{sec:compare}

Offsets of releases on one base live in the same space, one vector per matrix, so they can be compared and
exchanged (\cref{fig:schematic}c). We use three tests, from the weakest to the only causal one.

\textbf{Same benchmarks.} Let $o_i(b)$ be the offset gain of release $i$ on benchmark $b$, its \emph{profile}. A
plain correlation of $o_i$ and $o_j$ over benchmarks is inflated, because both are measured against the same base
run and share its sampling error (\cref{prop:sharedbase}). We evaluate $i$ on a random half of each benchmark's
items and $j$ on the other half, average over 200 splits, and compare the result with the same estimate for $i=j$,
the \emph{reliability} of one profile. Two profiles that are the same up to noise reach the geometric mean of the
two reliabilities, the \emph{noise ceiling} (\cref{prop:attenuation}). We report agreement as a fraction of it, for
the offset profile, for the extra-gain profile, and for the release profile as a reference.

\textbf{Same direction.} We compare the mean change of the residual state at the answer position, \mOff{} minus
\mBase{} (the \emph{offset shift}), between releases by cosine, on 96 held-back training prompts; the same for
\mRel{} minus \mOff{} (the \emph{extra shift}).

\textbf{Interchangeable.} For a set $S$ of releases, the \emph{\mShared{}} adds $\bS_k=\frac{1}{|S|}\sum_{i\in S}b^{(i)}_k$
to the untouched base (a matrix a release did not change contributes zero). For a release $i$ we compare, both on
the untouched base, its \emph{\mOwn{}} $b^{(i)}_k$ with the \emph{\mLoo{}}, the same average over $S\setminus\{i\}$:
if offsets were specific to each release, the offsets of other groups would not do the work of a release's own.
Averaging offsets is not averaging weights: $\bS_k$ differs from the offset of the averaged update by a covariance
between how releases differ in $\dW_k$ and how they differ in $\xb_k$ (\cref{prop:average}), so the test needs no
assumption that the weight updates themselves agree, and a dense fine-tune can exchange offsets with a low-rank
adapter.

\begin{table*}[t]
\centering
\small
\setlength{\tabcolsep}{2.9pt}
\caption{\textbf{What a constant reaches, per base (gain over \mBase{} on the DI average, points).} Left:
median (range) over the base's $n$ releases of the gain of the \mRel, of its \mRem{} (update minus offset) and of
its \mOff{} (offset only). Right: constants added to the untouched base, with 95\% intervals. \mBias{}
and \mBiasB{} act on the label logits only (a bias fitted on gold labels, and the same bias fitted per benchmark;
\cref{tab:calibration} adds the label-free \mPrior); the \mShared{} is the average offset of several releases, added inside the network (\cref{app:impl}).
Acc.: accuracy of \mBase{} in percent. \sig{Blue bold} (\signeg{red bold}): 95\% interval above (below) zero. \hib: higher is better. Per release: \cref{tab:releases}.}
\label{tab:summary}
\begin{tabular}{@{}lrr r@{\,}l r@{\,}l r@{\,}l r@{\,}l r@{\,}l >{\columncolor{oursbg}[\tabcolsep][0pt]}r@{}>{\columncolor{oursbg}[0pt][\tabcolsep]\,}l@{}}
\toprule
\rowcolor{hdrbg} & & & \multicolumn{6}{c}{Releases: median (range)} & \multicolumn{6}{c}{Constant added to the untouched base}\\
\cmidrule(lr){4-9}\cmidrule(l){10-15}
\rowcolor{hdrbg} Base & Acc. & $n$ & \multicolumn{2}{c}{\mRel{} \hib} & \multicolumn{2}{c}{\mRem{} \hib} & \multicolumn{2}{c}{\mOff{} \hib} & \multicolumn{2}{c}{\mBias{} \hib} & \multicolumn{2}{c}{\mBiasB{} \hib} & \multicolumn{2}{c}{\cellcolor{hdrbg}\mShared{} \hib}\\
\midrule
Qwen3.5-4B & 50.4 & 13 & \sig{+12.9} & {\scriptsize(+4.4 to +22.5)} & \sig{+12.1} & {\scriptsize(+2.5 to +22.1)} & \sig{+8.0} & {\scriptsize(+0.9 to +10.7)} & +0.1 & \ci{$-$1.3}{+1.4} & \sig{+5.0} & \ci{+2.8}{+7.0} & \sig{+10.5} & \ci{+8.4}{+12.6}\\
Qwen3.5-2B & 43.8 & 10 & \sig{+11.7} & {\scriptsize(+3.2 to +24.9)} & \sig{+12.0} & {\scriptsize(+0.3 to +23.8)} & \sig{+6.1} & {\scriptsize($-$10.5 to +8.7)} & +1.2 & \ci{$-$0.3}{+2.6} & \sig{+2.9} & \ci{+0.7}{+4.9} & \sig{+8.8} & \ci{+6.5}{+11.1}\\
Gemma-4-E4B & 51.2 & 8 & \sig{+9.2} & {\scriptsize(+2.2 to +16.5)} & \sig{+7.1} & {\scriptsize(+1.7 to +16.2)} & \sig{+7.2} & {\scriptsize(+2.6 to +7.9)} & +0.2 & \ci{$-$1.2}{+1.7} & +1.3 & \ci{$-$0.5}{+3.1} & \sig{+8.8} & \ci{+6.9}{+10.8}\\
\bottomrule
\end{tabular}
\end{table*}

\begin{figure}[t]
\centering
\includegraphics[width=\columnwidth]{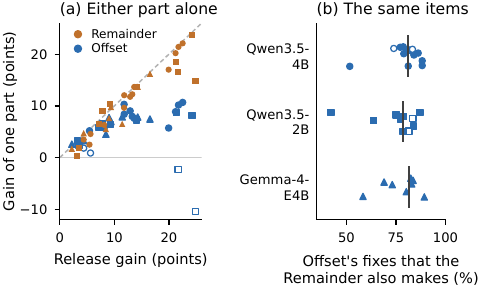}
\caption{\textbf{The offset is sufficient for part of the gain, and the release does not need it.} (a) Gain of
each part alone against the release gain, 31 releases: the \inbase{\mRem} (update minus offset) stays near the
diagonal, the \inours{\mOff} plateaus (hollow: exceptions; marker shape: base). (b) Of the items the \mOff{} model
fixes, the percentage that the \mRem{} model also fixes; one marker per release, bar: median. Values:
\cref{tab:releases,tab:redundancy}.}
\label{fig:rq1}
\end{figure}

\begin{figure*}[t]
\centering
\includegraphics[width=\textwidth]{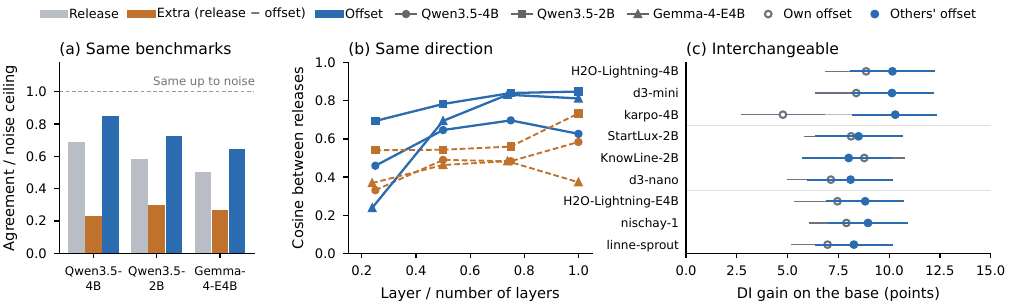}
\caption{\textbf{Offsets of different releases are one object; what releases reach beyond them is not.} (a)
Agreement of per-benchmark profiles between regular releases, as a fraction of the noise ceiling (1: the same up to
noise): whole releases, the extra gain, and the \inours{offset}. (b) Mean cosine between releases' residual shifts
at the answer position, by relative depth: \inours{offset shift} (solid) and \inbase{extra shift} (dashed). (c) DI
gain of the untouched base plus a release's own offset (hollow) or the average offset of the other releases
(\inours{blue}), with 95\% intervals. Values: \cref{tab:profiles,tab:direction,tab:shared}.}
\label{fig:rq2}
\end{figure*}

\section{Experiments}\label{sec:experiments}

\subsection{Setup}\label{sec:setup}

\textbf{Bases and releases.} Three bases: Qwen3.5-4B and Qwen3.5-2B \citep{qwen2026qwen35}, hybrids of linear and full attention, and
Gemma-4-E4B \citep{gemma2026gemma4}, a standard transformer. Releases were taken by a rule fixed before the second and third bases were run:
Hub models tagged \texttt{decision-model} that are fine-tunes of the base and load with its architecture, one per
group and base in download order (on Qwen3.5-4B, the first base, every such model that loaded, which includes two
variants from one group). This gives 31 usable releases from 26 groups (13, 10 and 8 per base), low-rank adapters
and dense fine-tunes alike. The three bases come from two families (Qwen3.5-2B, Qwen3.5-4B, Gemma-4-E4B). Seven more releases were split but
kept out of the analysis (\cref{tab:controls}): two gain nothing under the common readout, so there is no gain to
split; four, three of them versions of one release, fall 7 to 15 points below their base under the common prompt; and
one leaves the text weights unchanged (\cref{app:models}).

\textbf{How separate are the releases?} We read the 31 model cards (\cref{tab:cards}). Twenty describe their
training data, with different sources, teachers, scales (up to 6.2M examples) and input formats; seven name public
datasets, and two name the same public corpus of typed decisions \citep{sileo2023tasksource}. One release,
zils-mini, starts from a weight average of two others (d3-mini and torchcast-4B); it is in none of the averages or
exchange tests below. Eleven cards give no training information, so shared data cannot be ruled out for them.

\textbf{Controls.} The \emph{non-decision fine-tunes} are twelve popular fine-tunes of the same bases built for
other purposes (code, reasoning distillation, refusal removal, a medical model, a language adaptation, a general
assistant), four per base, run through the same split. Their offsets are constants of the same kind and natural
size as a release's, from tuning that had another purpose.

\textbf{Our adapters.} To ask what \emph{label-only tuning} produces, we train rank-64 LoRA adapters on Qwen3.5-4B
with the cross-entropy of the gold label under the same readout, on public training splits only, every item checked
against every evaluation item for a shared 13-word sequence. \mLk{} uses 8k items from 14 tasks. Four variants
change one thing each: \mLv{} (32k items, 38 tasks), \mLc{} (32k items from 668 tasks of the public decision
corpus, none of them an evaluated benchmark family), \mLx{} (\mLc{} with 4,675 items from 11 evaluated benchmark
families), and \mLd{} (\mLc{} items labelled with the answer distributions of the strongest permissive release,
H2O-Lightning-4B; \citealp{h2o2026h2olightning4b}). Recipes are in \cref{app:impl}; every model is cited in \cref{tab:licenses}.

\textbf{Evaluation.} Two public collections of decision benchmarks, the Decision
Index (DI, public suite 0.3; \citealp{decisionindex}) and JevBench \citep{jevbench}, and a private two-task classification set
(reported in \cref{app:figure-results} only). The \emph{DI average} is the macro accuracy over 36 DI benchmarks;
three more (ANLI, HellaSwag, WinoGrande; \citealp{nie2020anli,zellers2019hellaswag,sakaguchi2020winogrande}) are left out of it because our adapters train on their training splits,
and three single-class slices are excluded everywhere (\cref{app:datasets}). Released models and controls are
evaluated on a fixed sample, up to 60 items per DI benchmark, all 223 JevBench items and 300 items per private
task, 3,557 evaluations per model and variant; our adapters and seven headline releases also on the full set of
7,356. On four benchmarks whose correct option sits mostly in one position, every item is also evaluated with the
options reversed and scored by the mean of the two orders. Accuracy is in percent, differences in points;
intervals are 95\% paired bootstraps over items within benchmarks (1,000 draws, \cref{app:metrics}).

\textbf{Compared models.} For every release, the split gives \mBase, \mRel, \mRem{} (the update minus its one-mean offset)
and \mOff{} (the update replaced by its offset). Comparing releases adds the \mShared{} (the average offset of
several releases on the untouched base) and, for one release, its \mOwn{} against the \mLoo{} (the average over
the other releases). Four baselines correct the \emph{output} of the untouched base and change nothing inside it.
\mPrior{} subtracts from each label position its mean log-probability over items, a label-free correction in the
manner of \citet{zhao2021calibrate} and \citet{zhou2024batch}; \mBias{} adds a bias per label position fitted on gold
labels by maximum likelihood. Both use one vector per number of options, the most a constant at the output can
express. \mPriorB{} and \mBiasB{} estimate the same corrections separately for every benchmark, which no single
constant can do. Each is estimated on one half of the evaluation items and applied to the other. The \mSplit{} and
the \mShared{} are ours, drawn in \inours{blue}; \cref{tab:summary} gives the overview per base.

\subsection{RQ1: How much can a constant reach, and does the release need it?}\label{sec:rq1}

\textbf{A constant reaches a plateau: the \mOff{} alone gives \sig{+4.5 to +10.7} points in 20 of the 22
releases that gain 8 to 25 points.} \Cref{fig:teaser}a plots the offset gain against the release gain for all 31
releases. Releases that gain little are almost all offset: the 9 releases with a gain below 8 points keep 68 to
118\% of it, with two exceptions. Above 8 points the offset gain stops following the release gain: the 13 regular
releases that gain 8 to 15 points have a mean offset gain of $+7.5$, the 7 that gain 15 to 25 points $+8.5$, so
the offset share falls to 29--47\% for the six regular releases with the largest gains ($+20.0$ to $+24.3$
points). The same holds on all three bases (\cref{tab:summary,tab:releases}). On JevBench, the \mOff{} reaches a
median 81\% of the release's gain (26 regular releases with a JevBench gain above 3 points).

\textbf{Four exceptions.} We call a release an \emph{exception} if its offset gain is below half of $\min(\Gr, 8)$;
the other 27 are the \emph{regular} releases. The rule is descriptive and was set after seeing the data, so we also
give counts that do not use it: the offset gain is significantly above zero for 28 of the 31 releases. The four
exceptions sit at both ends of the size of a dense update. The two with the smallest weight changes of all (median
relative change 0.07\% and 0.15\%, against at least 0.9\% for every other surveyed release) gain 5.6 and 4.4
points, of which the offset reaches 16\% and 42\%. The two with the largest dense updates (5.5\% and 9.7\%),
karpo-2B and joey-1, gain 21.7 and 24.9 points while their offset loses 2.3 and 10.5. Scaling these two offsets
down removes the loss but yields no significant gain at any strength (\cref{sec:ablations}), so they are not the
common offset at a larger size. Two of the four strongest releases are thus outside the pattern.

\textbf{Sufficient, not necessary: without its offset, a release keeps a median \sig{94\%} of its gain.} The
\mRem{} model removes the (one-mean) offset from the release and keeps only the mean-zero, input-dependent part. It stays close
to the release (\cref{fig:rq1}a): it reproduces at least 75\% of the release gain in 25 of the 31 releases, and a
median 94\% in the 22 that gain 8 points or more. Removing the offset costs a regular release a median $0.8$ points
($-1.2$ to $+4.9$; the interval is above zero for 10 of the 27). The two parts are therefore not addends but
redundant routes: of the items the \mOff{} model fixes, the \mRem{} model also fixes a median 80\% (42--89\% over
the regular releases, \cref{fig:rq1}b, \cref{tab:redundancy}). The plateau is what a constant can do on a base, and
the input-dependent part of a release does that and more. The offset remains the part that can be compared across
releases, because it lives in a common space; the next question is whether it is the same.

\subsection{RQ2: Do releases of different groups share one offset?}\label{sec:rq2}

\textbf{Yes: offsets of different groups can be exchanged, and their average adds \sig{+8.8 to +10.5} points to an
untouched base, with no training.} On the untouched base, the average offset of the \emph{other} releases does at
least as well as a release's own: in all nine releases tested on the three bases, \mLoo{} minus \mOwn{} lies between
$-0.8$ and $+5.5$ points, and no interval lies below zero (\cref{fig:rq2}c). The \mShared{} of eight releases on
Qwen3.5-4B adds $+10.5$ \ci{8.4}{12.6} on the DI average and $+10.6$ on JevBench; six releases give $+8.8$ on
Qwen3.5-2B and on Gemma-4-E4B (\cref{tab:shared}). The averaged sets are eight regular releases from eight groups on Qwen3.5-4B and the six regular
releases with the largest DI gains on the other two bases (\cref{app:impl}); this choice used DI results, so JevBench
and random draws of one to four releases (\cref{tab:kcurve}) are the checks that do not depend on it. Averaging is
not what produces the gain: the offsets of
H2O-Lightning-4B and rycerzes-4B alone give $+8.9$ and $+10.3$, and averaging adds at most 1.6 points to these
(\cref{tab:kcurve}).

\textbf{The offsets point the same way and help the same benchmarks} (\cref{fig:rq2}a,b). At the middle layer, the
offset shifts of two regular releases have a cosine of 0.65, 0.78 and 0.69 on the three bases, against 0.49, 0.54
and 0.46 for their extra shifts; prompt by prompt, 0.62--0.76 against 0.33--0.44 (releases whose states were
stored, \cref{tab:direction}). Across benchmarks, the offset profiles of two regular releases agree at 64--85\% of
the noise ceiling. This is more than whole releases agree with each other (51--69\%), and far more than their
extra-gain profiles (23--30\%): what releases share is concentrated in the offset. The benchmarks it helps most are
fixed-option reasoning and classification tasks (\cref{tab:benchgain}).

\textbf{It is not a correction of the output.} If the offset only removed a label prior, a constant at the output
would do the same. It does not (\cref{tab:summary,tab:calibration}, \cref{fig:teaser}b): \mPrior{} gains $+0.6$ to $+1.8$ points
and \mBias, fitted on gold labels, $+0.1$ to $+1.2$. Even \mBiasB, fitted per benchmark, reaches $+1.3$ to $+5.0$,
and the \mShared{} is \sig{5.5 to 7.5} points above it on every base, each interval above zero
(\cref{tab:calibration}). The offset does remove label skew, as a by-product: the distance between the predicted and
the gold label distribution falls from 0.29--0.36 for the base to 0.16--0.22 with the \mShared, and \mPriorB{}
applied on top of the \mShared{} adds at most 0.4 points. A constant added \emph{inside} the network changes what
the base computes from the item; by \cref{prop:exact}(c) that computation is the base's own.

\begin{figure}[t]
\centering
\includegraphics[width=\columnwidth]{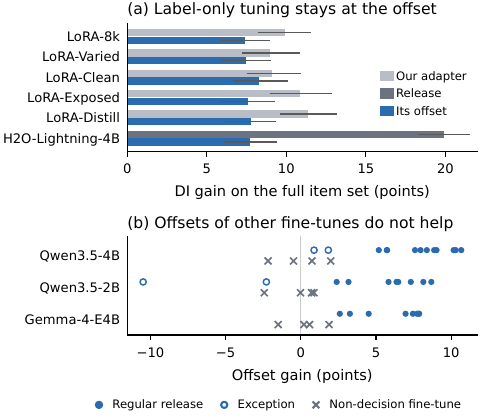}
\caption{\textbf{The offset comes from decision tuning, and label-only tuning reaches little more.} (a) Our
rank-64 LoRAs on Qwen3.5-4B, each changing one thing in the data, and H2O-Lightning-4B, on the full item set: gain
of the model (gray) and of its \inours{offset}, with 95\% intervals. (b) Offset gains of the releases and of
twelve non-decision fine-tunes of the same bases. Values: \cref{tab:adapters,tab:controls}.}
\label{fig:rq3}
\end{figure}

\subsection{RQ3: Where does the offset come from?}\label{sec:rq3}

\textbf{From decision tuning: offsets of fine-tunes for other purposes do not help.} The twelve non-decision
fine-tunes have offset gains between $-2.4$ and $+2.0$ points (\cref{fig:rq3}b), significantly positive for 2 of
the 12, against 28 of the 31 releases; 27 releases have an offset gain above the largest control's. A constant taken
from an arbitrary fine-tune of the same base is thus not enough. Their offset directions are unlike the releases'
at the middle layer on Qwen3.5-4B and Gemma-4-E4B (cosine at most 0.33 and 0.41), but not on Qwen3.5-2B, where two
of them reach 0.61 and 0.75 without a significant gain; direction is therefore supporting evidence, and the gain
and exchange tests carry the claim.

\textbf{Label-only tuning on public data reaches the plateau and little more.} Our own adapters land where the
offset does (\cref{fig:rq3}a). \mLk{} gains $+9.9$ points on the full item set, of which its offset reaches 75\%.
Four times the data, about 3 to 48 times the tasks, exposure to 11 evaluated benchmark families and the answer
distributions of H2O-Lightning-4B move the gain by at most 1.5 points, to $+9.0$ to $+11.3$, and keep the offset
share at 68--91\%. Another group's adapter trained on the same public corpus, rycerzes-4B, sits at the same point
($+11.8$, 87\%, on the sample), and the \mShared{} of eight releases, with no training at all, gains as much
($+10.5$). On Qwen3.5-4B and two further bases (Nemotron-3-Nano-4B, LFM3-VL-2B; \citealp{taghibakhshi2025elastic,liquidai2026lfm3}), one direction per matrix
(rank 1) is within 1.6 points of rank 64 on the held-out benchmarks, and training the offset alone, one bias vector
per matrix, reaches 63--81\% of the rank-1 adapter's held-out gain (\cref{tab:mechanism}). These are five recipes
with one seed each on one base; they show that scale, variety, exposure and distillation of labels do not leave the
plateau, not that no label-only recipe can.

\begin{figure}[t]
\centering
\includegraphics[width=\columnwidth]{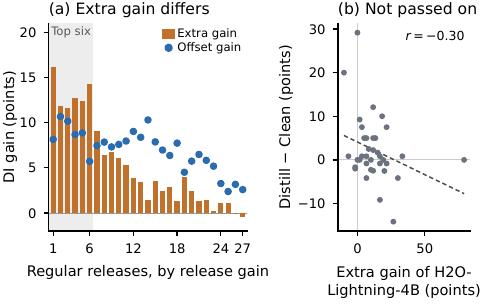}
\caption{\textbf{What separates releases is the extra gain, and distillation does not pass it on.} (a) The 27
regular releases, ordered by release gain: \inours{offset gain} and extra gain $\Ge$. (b) Per DI benchmark:
gain of \mLd{} over \mLc{} against H2O-Lightning-4B's extra gain. Values for (a): \cref{tab:releases}.}
\label{fig:rq4}
\end{figure}

\subsection{RQ4: What do the strongest releases add?}\label{sec:rq4}

\textbf{An input-dependent part that differs from release to release, with an extra gain of up to \sig{+16}
points that none of our manipulations reproduces.} The six regular releases with the largest gains have an extra
gain $\Ge$ of $+11.6$ to $+16.1$ points (\cref{fig:rq4}a, \cref{tab:releases}): at least this much of their gain needs the prompt.
This part is reliable within a release but not shared: the extra-gain profiles agree across releases at 23--30\% of
the noise ceiling (RQ2). It is partly concentrated on the seven benchmarks whose items share a 13-word sequence
with the public training corpus in at least 20\% of cases: there the extra gain is larger for 21 of the 27 regular
releases, $+6.9$ against $+5.0$ points on average (\cref{tab:overlap}); most of it lies on the other benchmarks.
Training on H2O-Lightning-4B's answers does not pass it on: the per-benchmark gain of \mLd{} over \mLc{} is
unrelated to H2O-Lightning-4B's extra-gain profile ($r=-0.30$ over 36 benchmarks, \cref{fig:rq4}b). Merging releases does not add their gains
either (\cref{app:merges}). For a user, the extra gain over
the \mShared{} is the number that separates one release from another; the gain over the base mostly is not.
\Cref{app:qualitative} shows items of both kinds.

\textbf{Using the \mShared{} as a reference.} For a new release on an audited base, add the \mShared{} to the
untouched base (no weight changes, one evaluation pass) and report the release's gain over it next to its gain over
the base. On a new base, the offset of a single existing release already gives most of the reference ($+8.9$ and
$+10.3$ against $+10.5$ from eight, \cref{tab:kcurve}).

\subsection{Ablations}\label{sec:ablations}

\begin{enumerate}
\item \textbf{One mean or two.} A single mean input for all positions gives offsets up to 4.2 points weaker than the
position-aware one used throughout (median 1.5 over the regular releases); no conclusion changes (\cref{tab:releases-variants}).
\item \textbf{Norms and embeddings.} The release's non-matrix parameters alone reach between $-3$\% and $+5$\% of
the gain of the 22 releases that gain 8 points or more.
\item \textbf{Sample and full set.} On all 7,356 evaluations, the release and offset gains of the seven
releases run on both item sets move by at most 1.1 and 1.0 points (\cref{tab:fullset}).
\item \textbf{How many releases.} Averages of one, two, four and eight releases give $+5.9$ to $+10.5$ points; the
lowest are the pairs that include karpo-4B, the least aligned of the eight averaged releases (\cref{tab:kcurve}).
\item \textbf{Scaling the exceptions.} joey-1's own offset on the untouched base (without its norms and embeddings) at full, half and quarter strength gives $-11.2$, $-3.8$
and $-0.9$ points, karpo-2B's $-2.9$, $+0.7$ \ci{$-$1.6}{+2.9} and $-0.1$: no strength gives a significant gain.
\item \textbf{Where a label-only adapter acts.} Active only on the options and the fixed instruction, our rank-1
adapter keeps 81--97\% of its held-out gain on three bases; active only on the record and question, 22--45\%
(\cref{tab:mechanism}).
\item \textbf{Option order.} Scoring the four position-skewed benchmarks in both orders instead of the original order alone changes the
list of regressions of at least 5 points for 11 of the 31 releases, in both directions (3 removed, 8 added;
\cref{tab:protocol}).
\end{enumerate}

\section{Limitations}\label{sec:limitations}

\textbf{Exchange on the base only.} The exchange test adds offsets to the untouched base, for nine of the 31
releases. We did not place another group's offset \emph{inside} a release (its \mRem{} plus the \mLoo), which
would test interchangeability where the offset was trained.

\textbf{Constant controls are real fine-tunes.} Our controls for ``any constant helps'' are the offsets of twelve
fine-tunes built for other purposes and the output-level corrections; we did not add random constants of matched
norm at the same matrices.

\textbf{One prompt, and groups that are not provably separate.} Every model is read through one prompt and its
label logits; four releases fall below their base under it and are excluded, and the offset we find is the one
visible under this prompt. Eleven model cards give no training information, so a shared offset may partly reflect
shared public data and a shared task format (\cref{tab:cards}); our own adapters, trained on public data only, reach
it too.

\section{Conclusion}\label{sec:conclusion}

Replacing a released update by its mean effect at every matrix shows that released decision models from different
groups share one input-independent offset. It reaches about the same gain whatever the release's strength, it
points the same way and can be exchanged across groups, it does more than any correction of the output, and it is
absent from fine-tunes for other purposes. The releases do not depend on it: their input-dependent part reaches the
same items and, for the strongest, many more. The average offset adds $+8.8$ to $+10.5$ points to an untouched base
with no training, as much as our label-only adapters. We recommend reporting any new decision release, and any
label-only tuning recipe, against this \mShared{} as well as against the base: a release that does not clear it has
shown nothing that a constant cannot do.

\section*{Impact Statement}

This work audits released models; it trains no new release. The \mSplit{} lets anyone with a base and a release's
weights check how much of a claimed gain survives without reading the prompt, and the \mShared{} gives a
training-free reference point for new releases, which can discourage inflated claims and wasted fine-tuning compute.
The same tool can extract and transplant the constant part of any fine-tune, including one tuned for unwanted
behaviour; the effect we find is bounded (at most about 11 points on the DI average) and gives no new
capability. All evaluated models and benchmarks are public except a private two-task classification set, used for
evaluation only, of which no item is shown. Third-party models are used under their licenses
(\cref{app:licenses}).

\FloatBarrier
\bibliography{main}
\bibliographystyle{icml2026}

\newpage
\appendix
\onecolumn
\setcounter{tocdepth}{2}
\tableofcontents
\newpage
\renewcommand{\arraystretch}{1.15}
\section{Glossary}\label{app:glossary}

An index of the names used in the text, tables and figures. Symbols are in \cref{tab:notation}.

\begin{table}[H]
\centering\small
\caption{\textbf{Glossary.}}
\label{tab:glossary}
\begin{tabular}{@{}p{3.6cm}p{8.6cm}p{3.6cm}@{}}
\toprule
\rowcolor{hdrbg} Name & Meaning & Defined in\\
\midrule
\grouprow{3}{Models and variants}{}
Base & the untuned public model a release was fine-tuned from & \cref{sec:setting}\\
Release & a decision model published with its weights by a third party; also the variant with its full update & \cref{sec:setting}\\
\mSplit{} (ours) & splitting a release's update into its offset and its remainder & \cref{sec:split}\\
\mOff{} & the base plus, at every changed matrix, the release's offset $b_k$ (position-aware) & \cref{sec:split}\\
One-mean offset & the offset with one mean input for all positions & \cref{sec:ablations}\\
\mRem{} & the base plus the release's update minus its one-mean offset (a model, like the \mOff) & \cref{sec:split}\\
Non-matrix only & the base matrices with the release's norms and embeddings & \cref{sec:ablations}\\
\mShared{} (ours) & the untouched base plus the average offset of several releases & \cref{sec:compare}\\
\mOwn{}, \mLoo{} & the base plus a release's own offset; plus the average offset of the other releases & \cref{sec:compare}\\
Regular release, exception & a release whose offset gain is at least / below half of $\min(\Gr,8)$ points & \cref{sec:rq1}\\
\mPrior, \mBias & corrections of the base's label logits: label-free, or fitted on gold labels; one vector per number of options & \cref{sec:setup}\\
\mPriorB, \mBiasB & the same corrections estimated per benchmark & \cref{sec:setup}\\
Non-decision fine-tune & a fine-tune of the same base built for another purpose (the control) & \cref{sec:setup}\\
\mLk, \mLv, \mLc, \mLx, \mLd & our rank-64 label-only LoRA adapters on Qwen3.5-4B & \cref{sec:setup}, \cref{app:impl}\\
\midrule
\grouprow{3}{Measurements}{}
DI average & macro accuracy over 36 Decision Index benchmarks & \cref{sec:setup}, \cref{app:datasets}\\
Trained tasks & ANLI, HellaSwag, WinoGrande: left out of the DI average & \cref{sec:setup}\\
Release / offset / remainder gain & accuracy of the \mRel{} / \mOff{} / \mRem{} model minus the base's, in points & \cref{sec:split}\\
Extra gain & release gain minus offset gain: what the release reaches beyond its offset & \cref{sec:split}\\
Plateau & the band the offset gain stays in for releases that gain 8 points or more & \cref{sec:rq1}\\
Label skew & total-variation distance between predicted and gold label histograms of a benchmark & \cref{sec:rq2}\\
Offset share & offset gain as a share of the release gain & \cref{sec:split}\\
Profile, reliability, noise ceiling & per-benchmark gains; their split-half self-agreement; the geometric mean of two reliabilities & \cref{sec:compare}\\
Offset shift, extra shift & mean change of the residual state at the answer position: \mOff{} minus \mBase; \mRel{} minus \mOff & \cref{sec:compare}\\
Sample, full set & the 3,557- and 7,356-evaluation item sets & \cref{sec:setup}, \cref{app:datasets}\\
Position-skewed benchmark & a benchmark whose correct option sits mostly in one position; scored in both orders & \cref{app:datasets}\\
Single-class slice & a benchmark whose items all have the same correct label in our item set; excluded & \cref{app:datasets}\\
\bottomrule
\end{tabular}
\end{table}

\section{Theoretical Results and Proofs}\label{app:theory}

Every result is set in a shaded card; results first stated in the main text keep their numbers. Proofs are written
as numbered steps; a displayed chain has one transformation per line with its reason in brackets on the right. All
results are also checked in Lean~4 with Mathlib (\cref{app:llm}); the formal statements are about finite-dimensional
real vectors and finite probability spaces.

\begin{table}[H]
\centering\small
\caption{\textbf{Theory map.}}
\label{tab:theorymap}
\begin{tabular}{@{}llp{7.4cm}l@{}}
\toprule
\rowcolor{hdrbg} Result & Name & Statement (one line) & Used in\\
\midrule
\Cref{prop:exact}(a,b) & Exact and best constant & $\dW x=b+\dW(x-\xb)$; $b$ is the closest constant in mean square & \cref{sec:props}\\
\Cref{prop:exact}(c) & No item information & the \mOff{} model adds the same vectors to every prompt & \cref{sec:props,sec:rq2}\\
\Cref{prop:lowrank} & Low-rank offsets & for $\dW=BA$ the offset lies in the column space of $B$ & \cref{sec:props}\\
\Cref{prop:additive} & Additivity under linearity & the two parts add if the logits are affine in the added terms & \cref{sec:props}\\
\Cref{prop:average} & Averaged offset & average of offsets $=$ offset of the averaged update $+$ a covariance term & \cref{sec:compare}\\
\Cref{prop:sharedbase} & Shared-base bias & naive profile products carry the base's sampling variance; disjoint halves do not & \cref{sec:compare}\\
\Cref{prop:attenuation} & Noise ceiling & split-half correlation $=\rho\sqrt{\lambda_i\lambda_j}$; reliability $=\lambda_i$ & \cref{sec:compare,sec:rq2}\\
\Cref{prop:skew} & Option order & averaging two orders removes the gold-position distribution for $K=2$ & \cref{sec:setup,sec:ablations}\\
\Cref{prop:singleclass} & Single-class slices & accuracy on a single-class slice is a label rate & \cref{sec:setup}\\
\bottomrule
\end{tabular}
\end{table}

\subsection{Setting and notation}\label{app:notation}

\begin{table}[H]
\centering\small
\caption{\textbf{Notation} shared by the main text and the proofs. Symbols used in one proof only are defined
there.}
\label{tab:notation}
\begin{tabular}{@{}lp{12cm}@{}}
\toprule
\rowcolor{hdrbg} Symbol & Meaning\\
\midrule
\grouprow{2}{Items and readout}{}
$s,q,o_1..o_K,y$ & record, question, options, index of the correct option\\
$h,\ e_{\ell_k},\ \hat y$ & final hidden state at the answer position; output embedding of label $\ell_k$; predicted option\\
\midrule
\grouprow{2}{Updates and the split}{}
$W_k,\ W'_k,\ \dW_k$ & matrix $k$ of the base, of the release, and their difference ($m_k\times n_k$)\\
$x_{k,t}$ & input of matrix $k$ at token position $t$\\
$\Cset,\ N_{\Cset}$ & calibration prompts (300 public training items) and their number of tokens\\
$\xb^{\mathrm{last}}_k,\ \xb^{\mathrm{rest}}_k$ & mean input of matrix $k$ over the last positions and over the other positions of the calibration prompts, release active\\
$\xb_k,\ \xb^{\mathrm{all}}_k$ & $\xb^{\mathrm{last}}_k$ at the answer position and $\xb^{\mathrm{rest}}_k$ elsewhere (\mOff); one mean over all calibration tokens (\mRem)\\
$b_k=\dW_k\xb_k$ & the offset of matrix $k$\\
$\Acc_m$ & accuracy of model $m$, averaged over the benchmarks of a group\\
$\Gr,\ \Go,\ \Gm$ & gains of the \mRel, \mOff{} and \mRem{} models over the base\\
$\Ge$ & extra gain, $\Gr-\Go$\\
$\sigma=\Go/\Gr$ & the offset share\\
\midrule
\grouprow{2}{Comparing releases}{}
$S,\ \bS_k$ & a set of releases; their average offset (the \mShared)\\
$o_i(b),\ r_i(b)$ & offset gain and extra gain of release $i$ on benchmark $b$ (the profiles)\\
$\lambda_i,\ \rho_{ij}$ & reliability of release $i$'s profile; correlation of the noise-free profiles of $i$ and $j$\\
$d^{(i)}_\ell$ & mean change of the residual state after layer $\ell$ at the answer position, \mOff{} minus \mBase\\
\bottomrule
\end{tabular}
\end{table}

A changed matrix of a release computes $W'_k x=W_k x+\dW_k x$. The four models of \cref{sec:split} add, at every
changed matrix and position, $0$ (\mBase), $\dW_k x$ (\mRel), $b_k$ (\mOff) or $\dW_k(x-\xb^{\mathrm{all}}_k)$ (\mRem). Inputs
$x_{k,t}$ are those of the model being run, so they depend on what earlier matrices added.

\subsection{Properties of the split}

\Needspace{10\baselineskip}
\begin{resbox}\propsplit*\end{resbox}
\begin{proof}[Proof of (a) and (b)]
\emph{Step 1 (decomposition).} For any $x$,
\begin{align*}
\dW_k x &= \dW_k\big(\xb_k+(x-\xb_k)\big) \why{add and subtract $\xb_k$}\\
&= \dW_k\xb_k+\dW_k(x-\xb_k) \why{linearity of $\dW_k$}\\
&= b_k+\dW_k(x-\xb_k). \why{definition~\eqref{eq:offset}}
\end{align*}
\emph{Step 2 (the remainder has mean zero).} Write $\sum_{c,t}$ for the sum over calibration prompts and their tokens.
\begin{align*}
\frac{1}{N_{\Cset}}\sum_{c,t}\dW_k\big(x^{(c)}_{k,t}-\xb_k\big) &= \dW_k\Big(\frac{1}{N_{\Cset}}\sum_{c,t}x^{(c)}_{k,t}-\xb_k\Big) \why{linearity; $N_{\Cset}$ terms}\\
&= \dW_k(\xb_k-\xb_k)=0. \why{definition of $\xb_k$}
\end{align*}
\emph{Step 3 (best constant).} Let $y_{c,t}=\dW_k x^{(c)}_{k,t}$ and $\bar y$ its mean, so $\bar y=b_k$ by Step 2.
For any constant $v\in\R^{m_k}$,
\begin{align*}
\sum_{c,t}\lVert y_{c,t}-v\rVert^2 &= \sum_{c,t}\lVert (y_{c,t}-\bar y)+(\bar y-v)\rVert^2 \why{add and subtract $\bar y$}\\
&= \sum_{c,t}\lVert y_{c,t}-\bar y\rVert^2+2\Big\langle \sum_{c,t}(y_{c,t}-\bar y),\,\bar y-v\Big\rangle+N_{\Cset}\lVert\bar y-v\rVert^2 \why{expand the square}\\
&= \sum_{c,t}\lVert y_{c,t}-\bar y\rVert^2+N_{\Cset}\lVert\bar y-v\rVert^2 \why{the middle sum is zero, Step 2}\\
&\ge \sum_{c,t}\lVert y_{c,t}-b_k\rVert^2, \why{$\bar y=b_k$; a norm is nonnegative}
\end{align*}
with equality if and only if $v=b_k$. Steps 2 and 3 are written for one set of $N_{\Cset}$ calibration tokens and
its mean; they apply separately to the last positions with $\xb^{\mathrm{last}}_k$ ($|\Cset|$ tokens), to the other
positions with $\xb^{\mathrm{rest}}_k$ ($N_{\Cset}-|\Cset|$ tokens), and to all tokens with $\xb^{\mathrm{all}}_k$
(the \mRem{} model); (a) holds for any of these means.
\end{proof}

\begin{proof}[Proof of (c)]
By the definition of the \mOff{} model, the vector added at matrix $k$ and position $t$ of a prompt of length $T$ is
$b^{\mathrm{last}}_k$ if $t=T$ and $b^{\mathrm{rest}}_k$ otherwise. Both are computed from the calibration set and
the release before any evaluation item is seen, so for two prompts the added vectors at matching position classes
are equal. All remaining computation is the base's weights applied to the prompt.
\end{proof}

\Needspace{10\baselineskip}
\begin{resbox}
\begin{proposition}[Low-rank offsets]\label{prop:lowrank}
If $\dW_k=B_kA_k$ with $B_k\in\R^{m\times r}$ and $A_k\in\R^{r\times n}$, then $b_k=B_k(A_k\xb_k)$ lies in the
column space of $B_k$. For rank one, $\dW_k=u_kv_k^\top$, the release adds $u_k\,(v_k^\top x)$ and the offset adds
$u_k\,(v_k^\top\xb_k)$: the split separates the mean of one scalar gate per matrix from its variation.
\end{proposition}
\end{resbox}
\begin{proof}
\begin{align*}
b_k &= (B_kA_k)\xb_k \why{definition~\eqref{eq:offset}}\\
&= B_k(A_k\xb_k). \why{associativity of matrix products}
\end{align*}
The right-hand side is $B_k$ times a vector of $\R^r$, hence in its column space. For rank one,
$u_kv_k^\top x=u_k(v_k^\top x)$ by associativity, and the same with $\xb_k$.
\end{proof}

\Needspace{10\baselineskip}
\begin{resbox}
\begin{proposition}[Additivity under linearity]\label{prop:additive}
Suppose the label logits of a prompt are an affine function of the vectors added at the changed matrices,
$z(\delta)=z(0)+L\delta$ for a linear map $L$. Then
$z_{\mRel}-z_{\mBase}=(z_{\mOff}-z_{\mBase})+(z_{\mRem}-z_{\mBase})$.
\end{proposition}
\end{resbox}
\begin{proof}
\emph{Step 1 (the added terms add).} The three models add $\delta^{\mathrm{rel}}_{k,t}=\dW_kx_{k,t}$,
$\delta^{\mathrm{off}}_{k,t}=b_k$ and $\delta^{\mathrm{rem}}_{k,t}=\dW_kx_{k,t}-b_k$, so
$\delta^{\mathrm{rel}}=\delta^{\mathrm{off}}+\delta^{\mathrm{rem}}$ by \cref{prop:exact}(a). Under the affine
assumption the inputs $x_{k,t}$ are those of the base (their dependence on earlier additions is absorbed into $L$).
The identity needs $b_k$ to be the same in $\delta^{\mathrm{off}}$ and $\delta^{\mathrm{rem}}$; for the reported
models (position-aware \mOff, one-mean \mRem) the two differ, by $\dW_k(\xb^{\mathrm{last}}_k-\xb^{\mathrm{all}}_k)$
at the answer position and $\dW_k(\xb^{\mathrm{rest}}_k-\xb^{\mathrm{all}}_k)$ elsewhere.

\emph{Step 2 (apply the affine map).}
\begin{align*}
z_{\mRel}-z_{\mBase} &= L\,\delta^{\mathrm{rel}} \why{affine assumption; $z_{\mBase}=z(0)$}\\
&= L\,\delta^{\mathrm{off}}+L\,\delta^{\mathrm{rem}} \why{Step 1; linearity of $L$}\\
&= (z_{\mOff}-z_{\mBase})+(z_{\mRem}-z_{\mBase}). \why{affine assumption, twice}
\end{align*}
\end{proof}
A transformer is not affine in the added terms and accuracy is a step function of the logits, so gains need not add:
if either part alone moves the correct label to the top, adding the other changes nothing ($\Go+\Gm>\Gr$ is
then possible). In the audit this is the rule: the \mRem{} model alone reaches a median 94\% of the release gain of the releases that gain 8 points or more, while the \mOff{} model alone reaches +4.5 to +10.7 points in 20 of them (\cref{tab:releases,tab:redundancy}). This is why the two parts are read as two counterfactuals (\cref{sec:props}).

\Needspace{10\baselineskip}
\begin{resbox}
\begin{proposition}[Averaged offset]\label{prop:average}
For releases $i\in S$ with updates $\dW^{(i)}_k$ and mean inputs $\xb^{(i)}_k$, let
$\overline{\dW}_k=\frac{1}{|S|}\sum_i\dW^{(i)}_k$ and $\tilde x_k=\frac{1}{|S|}\sum_i\xb^{(i)}_k$. Then
$\bS_k=\overline{\dW}_k\tilde x_k+\frac{1}{|S|}\sum_{i\in S}\big(\dW^{(i)}_k-\overline{\dW}_k\big)\big(\xb^{(i)}_k-\tilde x_k\big)$.
\end{proposition}
\end{resbox}
\begin{proof}
Write $E_i=\dW^{(i)}_k-\overline{\dW}_k$ and $e_i=\xb^{(i)}_k-\tilde x_k$, so $\sum_iE_i=0$ and $\sum_ie_i=0$.
\begin{align*}
\bS_k &= \frac{1}{|S|}\sum_i\dW^{(i)}_k\xb^{(i)}_k \why{definition of $\bS_k$ and $b^{(i)}_k$}\\
&= \frac{1}{|S|}\sum_i(\overline{\dW}_k+E_i)(\tilde x_k+e_i) \why{definitions of $E_i$, $e_i$}\\
&= \overline{\dW}_k\tilde x_k+\overline{\dW}_k\frac{1}{|S|}\sum_ie_i+\Big(\frac{1}{|S|}\sum_iE_i\Big)\tilde x_k+\frac{1}{|S|}\sum_iE_ie_i \why{expand; linearity}\\
&= \overline{\dW}_k\tilde x_k+\frac{1}{|S|}\sum_iE_ie_i. \why{$\sum_ie_i=0$, $\sum_iE_i=0$}
\end{align*}
\end{proof}
The first term is what averaging the weight updates (task arithmetic) and then taking the offset would give. The
second couples how releases differ in their updates with how they differ in the inputs they induce, which is why
the \mShared{} is tested directly rather than through averaged weights.

\subsection{Statistical safeguards}

For release $i$ and benchmark $b$ the observed offset profile is
$\hat o_i(b)=o_i(b)+\varepsilon_i(b)-\varepsilon_0(b)$, where $\varepsilon_i$ and $\varepsilon_0$ are the sampling
errors (mean zero) of the release's and the base's accuracy on that benchmark's items.

\Needspace{10\baselineskip}
\begin{resbox}
\begin{proposition}[Shared-base bias]\label{prop:sharedbase}
If releases $i\ne j$ are evaluated on the same items against the same base run, then
$\E[\hat o_i(b)\hat o_j(b)]=o_i(b)o_j(b)+\E[\varepsilon_i\varepsilon_j]-\E[\varepsilon_i\varepsilon_0]-\E[\varepsilon_j\varepsilon_0]+\Var(\varepsilon_0)$,
and the term $\Var(\varepsilon_0)>0$ is present even when the true profiles are unrelated. If instead $i$ is
evaluated on one half of the items and $j$ on a disjoint half, each against the base on its own half, then
$\E[\hat o^{(1)}_i(b)\,\hat o^{(2)}_j(b)]=o_i(b)o_j(b)$.
\end{proposition}
\end{resbox}
\begin{proof}
\emph{Step 1 (same items).} Drop the argument $b$.
\begin{align*}
\E[\hat o_i\hat o_j] &= \E[(o_i+\varepsilon_i-\varepsilon_0)(o_j+\varepsilon_j-\varepsilon_0)] \why{definition of $\hat o$}\\
&= o_io_j+o_i\E[\varepsilon_j-\varepsilon_0]+o_j\E[\varepsilon_i-\varepsilon_0]+\E[(\varepsilon_i-\varepsilon_0)(\varepsilon_j-\varepsilon_0)] \why{expand; $o_i,o_j$ fixed}\\
&= o_io_j+\E[\varepsilon_i\varepsilon_j]-\E[\varepsilon_i\varepsilon_0]-\E[\varepsilon_j\varepsilon_0]+\E[\varepsilon_0^2] \why{zero means; expand}\\
&= o_io_j+\E[\varepsilon_i\varepsilon_j]-\E[\varepsilon_i\varepsilon_0]-\E[\varepsilon_j\varepsilon_0]+\Var(\varepsilon_0). \why{$\E\varepsilon_0=0$}
\end{align*}
\emph{Step 2 (disjoint halves).} The errors $(\varepsilon^{(1)}_i,\varepsilon^{(1)}_0)$ are functions of the first
half and $(\varepsilon^{(2)}_j,\varepsilon^{(2)}_0)$ of the second; items are drawn independently, so the two pairs
are independent with mean zero and
$\E[(\varepsilon^{(1)}_i-\varepsilon^{(1)}_0)(\varepsilon^{(2)}_j-\varepsilon^{(2)}_0)]=0$. Repeating the second
line of Step 1 with this term gives $o_io_j$.
\end{proof}

\Needspace{10\baselineskip}
\begin{resbox}
\begin{proposition}[Noise ceiling]\label{prop:attenuation}
Let the profile measured on a half be $\hat o^{(h)}_i=o_i+\eta^{(h)}_i$, with noise uncorrelated with the true
profiles and across halves, and let $\lambda_i=\Var(o_i)/(\Var(o_i)+\Var(\eta_i))$, the variance taken over
benchmarks. Then $\corr(\hat o^{(1)}_i,\hat o^{(2)}_j)=\rho_{ij}\sqrt{\lambda_i\lambda_j}$ and
$\corr(\hat o^{(1)}_i,\hat o^{(2)}_i)=\lambda_i$. Two profiles that are equal up to noise ($\rho_{ij}=1$) reach
$\sqrt{\lambda_i\lambda_j}$, the noise ceiling.
\end{proposition}
\end{resbox}
\begin{proof}
\emph{Step 1 (covariance and variance).} The noise terms are uncorrelated with everything else, so
$\Cov(\hat o^{(1)}_i,\hat o^{(2)}_j)=\Cov(o_i,o_j)$, and $\Var(\hat o^{(h)}_i)=\Var(o_i)+\Var(\eta_i)=\Var(o_i)/\lambda_i$.

\emph{Step 2 (correlation).}
\begin{align*}
\corr(\hat o^{(1)}_i,\hat o^{(2)}_j) &= \frac{\Cov(o_i,o_j)}{\sqrt{\Var(o_i)/\lambda_i\cdot\Var(o_j)/\lambda_j}} \why{Step 1}\\
&= \frac{\Cov(o_i,o_j)}{\sqrt{\Var(o_i)\Var(o_j)}}\sqrt{\lambda_i\lambda_j} \why{factor the square root}\\
&= \rho_{ij}\sqrt{\lambda_i\lambda_j}. \why{definition of $\rho_{ij}$}
\end{align*}
For $j=i$, $\rho_{ii}=1$ and both variances carry $\lambda_i$, so the value is $\lambda_i$.
\end{proof}
In \cref{tab:profiles} the ratio is the mean split-half correlation over pairs divided by the mean noise ceiling; it
estimates an average $\rho$ under these assumptions.

\Needspace{10\baselineskip}
\begin{resbox}
\begin{proposition}[Option order]\label{prop:skew}
Let $a_p(i)$ be the probability that a model answers item $i$ correctly when its correct option is shown at position
$p$. With $K=2$ options, the score averaged over the original and the reversed order is $\tfrac12(a_1(i)+a_2(i))$
whatever the original position of the correct option; a model that answers by position only scores exactly
$\tfrac12$.
\end{proposition}
\end{resbox}
\begin{proof}
Reversing two options moves the correct one from position $p_i$ to $3-p_i$, so the averaged score is
$\tfrac12(a_{p_i}(i)+a_{3-p_i}(i))$, and $\{p_i,3-p_i\}=\{1,2\}$. A position-only model picks position 1 with some
probability $\pi$ for every item, so $a_1(i)=\pi$, $a_2(i)=1-\pi$ and the average is $\tfrac12$.
\end{proof}
For $K>2$ the reversal pairs positions $p$ and $K+1-p$, which removes a first-versus-last preference but not an
arbitrary one. Three of the four skewed benchmarks are binary; the fourth has four options.

\Needspace{10\baselineskip}
\begin{resbox}
\begin{proposition}[Single-class slices]\label{prop:singleclass}
If every item of a benchmark has the same correct label $y^\star$, the accuracy of a model equals the rate at which
it answers $y^\star$, and the constant answer $y^\star$ scores 100\%.
\end{proposition}
\end{resbox}
\begin{proof}
Accuracy is the share of items with $\hat y(i)=y(i)$; with $y(i)=y^\star$ for every $i$ this is the share with
$\hat y(i)=y^\star$, which is 1 for the constant answer.
\end{proof}
Since the \mOff{} adds a constant vector, it can shift exactly such label rates; three benchmarks of our item set are
single-class (\cref{tab:protocol}) and are therefore excluded from every average.

\section{Datasets}\label{app:datasets}

\textbf{Evaluation sets.} The Decision Index public suite \citep{decisionindex} (version 0.3, hash-checked against the published suite)
asks one typed question per sampled row of each benchmark; we use the 39 benchmarks whose prompts fit 3,000 tokens.
The DI average covers 36 of them: ARC-Challenge and ARC-Easy~\citep{clark2018arc}, Amazon ESCI~\citep{reddy2022esci}, BANKING77~\citep{casanueva2020banking77}, BBH fixed-option tasks~\citep{suzgun2023bbh}, BFCL, BPoMP, CLINC150 with out-of-scope~\citep{larson2019clinc}, CLadder~\citep{jin2023cladder}, CRUXEval~\citep{gu2024cruxeval}, ChessBench~\citep{ruoss2024chessbench}, ContractNLI~\citep{koreeda2021contractnli}, FinEntity~\citep{tang2023finentity}, GPQA Diamond~\citep{rein2024gpqa}, GSM8K~\citep{cobbe2021gsm8k}, HLE~\citep{cais2026hle}, Habermas Machine, HoVer~\citep{jiang2020hover}, a home-appliance simulator, Humicroedit~\citep{hossain2019humicroedit}, MMLU~\citep{hendrycks2021mmlu}, MMLU-Pro~\citep{wang2024mmlupro}, MuSR~\citep{sprague2024musr}, NLI4CT~\citep{jullien2023nli4ct}, New Yorker caption matching~\citep{hessel2023newyorker}, POP909-CL (built on POP909 \citep{wang2020pop909}), PhishNChips, RAGTruth~\citep{niu2024ragtruth}, RouterBench~\citep{hu2024routerbench}, SATA-Bench~\citep{xu2026sata}, SGD and SGD-X~\citep{rastogi2020sgd,lee2022sgdx}, SimpleBench, VAST~\citep{allaway2020vast}, When2Call~\citep{ross2025when2call},
cfcolor and iSarcasmEval; benchmarks without a citation here are documented in the suite \citep{decisionindex}. The other three, ANLI~\citep{nie2020anli}, HellaSwag~\citep{zellers2019hellaswag} and WinoGrande~\citep{sakaguchi2020winogrande}, are the \emph{trained tasks}: our adapters
use their training splits, so they are reported separately. Three more benchmarks are single-class in this
construction (BRIGHT~\citep{su2025bright} and ToolRet~\citep{shi2025toolret}: every scored item is ``relevant''; ACOS: 99\% ``absent'',
\cref{tab:protocol}) and are excluded everywhere (\cref{prop:singleclass}). JevBench \citep{jevbench} has 223 items in three slices
(easy, hard, original; 8 of 231 exceed the length limit), macro-averaged over slices. The private set has two
classification tasks (topic, tax type) of 797 rows each; no item is shown.

\textbf{Item sets.} The \emph{sample} draws, with a fixed seed, up to 60 items per DI benchmark (fewer where a
benchmark has fewer), all JevBench items and 300 items per private task: 3,077 items. The four benchmarks whose
correct option sits mostly in one position (Amazon ESCI, RAGTruth, iSarcasmEval and private tax type) are also
evaluated with the options reversed, 480 more evaluations, 3,557 in total. The \emph{full set} uses every item of the
same benchmarks, 7,356 evaluations. Every release and control uses the sample; our adapters and seven headline
releases also the full set. The mechanism runs (\cref{tab:mechanism}) use a 6,551-item audit set in the
original option order only.

\textbf{Training and calibration data.} All training items are public training splits, with prompts of at most
1,600 tokens and 2 to 26 options; every item that shares a 13-word sequence with any evaluation item is dropped.
\mLk{} uses 5,000 items from 11 public sources plus 1,000 each from the training splits of WinoGrande, HellaSwag and
ANLI. A control mix of 8,000 items from the same sources without the three tasks supplies the calibration set (its
first 300 items) and the direction prompts (items 300 to 395). \mLv{} adds 24,000 items from further public tasks (38 tasks in all).
\mLc{} takes 32,000 items round-robin over 668 source tasks of a public decision corpus \citep{sileo2023tasksource},
none of them an evaluated benchmark family; \mLx{} replaces 4,675 of them by training items of 11 evaluated families
(461 each from ANLI, ARC, Amazon ESCI, ContractNLI, HellaSwag, HoVer, Humicroedit, NLI4CT, WinoGrande and
iSarcasmEval, 65 from RAGTruth); \mLd{} keeps the \mLc{} items with H2O-Lightning-4B's answer distributions as
targets (it agrees with the dataset label on 73\% of them).

\textbf{Corpus overlap.} Of all evaluation items, 441 share a 13-word sequence with a training row of the public
corpus, concentrated in seven benchmarks (\cref{tab:overlap}); a shared sequence is often a passage or a contract
clause, not the same question and label.

\section{Models}\label{app:models}

\textbf{Bases.} Qwen3.5-4B (32 layers, width 2,560) and Qwen3.5-2B (24 layers, width 2,048) mix linear-attention and
full-attention layers; Gemma-4-E4B (42 layers, width 2,560) is a standard transformer. The mechanism runs of
\cref{tab:mechanism} add Nemotron-3-Nano-4B (a Mamba-2 hybrid) and LFM3-VL-2B (a short-convolution hybrid). Weights
in bfloat16; one 20\,GB GPU.

\textbf{Inclusion of releases.} On each base: Hub models tagged \texttt{decision-model} whose card names the base as
the fine-tuning source, that load with the base's architecture, one per group, in download order; the rule was fixed
before the runs on Qwen3.5-2B and Gemma-4-E4B. On Qwen3.5-4B every such model that loaded was taken. All are listed
in \cref{tab:licenses}. One candidate was dropped before running because it is not on the stated base. Releases
shipped as unmerged LoRA adapters are merged as $\dW=\frac{\alpha}{r}BA$; one of them (Nullmega-LoRA) reads out
through its own joint head, which our readout ignores.

\textbf{What the model cards declare.} \Cref{tab:cards} lists, for each release, whether its card describes the
training data, the public datasets it names and any audited release it names as its starting point. Twenty of the
31 cards describe training data; seven name public datasets; rycerzes-4B and zils-mini name the same public corpus
of typed decisions \citep{sileo2023tasksource}, which is also the source of \mLc; zils-mini starts from a weight
average of d3-mini and torchcast-4B. One release has no card and one an empty template. Four groups release on two
bases (and one of them two variants on Qwen3.5-4B), which is why 31 releases come from 26 groups. The \mShared{} sets
and the exchange tests do not contain zils-mini.

\begingroup\small
\begin{longtable}{@{}llp{7.4cm}l@{}}
\caption{\textbf{What the model cards of the 31 releases declare} (read on 20 November 2026; card text and
repository metadata only). Data: whether the card describes the training data
(20 of 31). Named public datasets: those the card or its metadata names (7 of 31; at most four
shown). Derived from: another audited release the card names as its starting point. --: not described, or none
named.}\label{tab:cards}\\
\toprule
\rowcolor{hdrbg} Release & Data & Named public datasets & Derived from\\
\midrule
\endfirsthead
\toprule
\rowcolor{hdrbg} Release & Data & Named public datasets & Derived from\\
\midrule
\endhead
\grouprow{4}{Qwen3.5-4B}{}
H2O-Lightning-4B & described & -- & --\\
StartLux-4B & described & -- & --\\
StartLux-4B-AT & described & -- & --\\
karpo-4B & -- & -- & --\\
zils-mini & described & tasksource-jev-typed-decisions, base-exam-seed-v1, base-exam-seed-v2, base-exam-seed-v3 & d3-mini, torchcast-4B\\
d3-mini & -- & -- & --\\
EpochEcho-4B & described & -- & --\\
torchcast-4B & described & -- & --\\
rycerzes-4B & described & tasksource-jev-typed-decisions & --\\
nexus-4B & described & nexus-decision-data, banking77, clinc\_oos & --\\
akiray1-bbh & -- & -- & --\\
Nullmega-LoRA & -- & -- & --\\
Intern-Decision-4B & -- & -- & --\\
\midrule
\grouprow{4}{Qwen3.5-2B}{}
joey-1 & described & -- & --\\
StartLux-2B & described & -- & --\\
karpo-2B & -- & -- & --\\
KnowLine-2B & described & -- & --\\
d3-nano & -- & -- & --\\
fort-1-2B & described & MASSIVE, SQuAD 2.0, CLINC150, DBpedia14, \dots & --\\
Jeff-2B & described & WinoGrande, RAGTruth, ContractNLI, CUAD, \dots & --\\
unee-2B & described & bekko-system-one-dataset-v0, SQuAD 2.0, oasst1 & --\\
JevK5-2B & described & MMLU-Pro, WANLI, MultiNLI, BoolQ, \dots & --\\
classone-2B & -- & -- & --\\
\midrule
\grouprow{4}{Gemma-4-E4B}{}
H2O-Lightning-E4B & -- & -- & --\\
Smaug-Decisions & described & -- & --\\
nischay-1 & -- & -- & --\\
linne-sprout & -- & -- & --\\
cue-reflex & described & -- & --\\
basal-omni & described & -- & --\\
nirnaya-flash & described & -- & --\\
jez-large & described & -- & --\\
\bottomrule
\end{longtable}
\endgroup

\textbf{Excluded releases.} Seven releases were run and are not analysed (\cref{tab:controls}): two gain nothing
under the common readout, so there is nothing to split; four fall 7 to 15 points below their base under the common
prompt (their JevBench accuracy drops by more than 30 points), which says they need their own input format; one has
text weights identical to its base.

\textbf{Controls.} Four popular fine-tunes per base built for other purposes, chosen by download count among those
that load with the base's architecture (\cref{tab:controls}).

\section{Implementation Details}\label{app:impl}

Every setting, taken from the code and the configuration file. Statistics are in \cref{app:metrics}.

\begingroup\small
\begin{longtable}{@{}p{5.2cm}p{10.6cm}@{}}
\caption{\textbf{Settings.}}\label{tab:impl}\\
\toprule
\rowcolor{hdrbg} Setting & Value\\
\midrule
\endfirsthead
\toprule
\rowcolor{hdrbg} Setting & Value\\
\midrule
\endhead
\grouprow{2}{Readout}{}
Prompt & one user turn (\cref{app:prompts}) in the base's chat template, generation prompt added, thinking disabled\\
Answer & arg-max over the logits of the $K$ label tokens (A--Z, a--z, 0--9; single tokens) at the last prompt position; ties broken by the lowest index; no sampling, no generated text\\
Prompt length & at most 3,000 tokens (Qwen3.5 tokenizer, the same item set on every base)\\
Precision & bfloat16 weights; label log-probabilities in float32\\
\midrule
\grouprow{2}{Offset Split}{}
Changed matrices & every two-dimensional text-model weight whose values differ from the base; other changed tensors (norm gains, embeddings) take the release's values in every variant\\
Calibration & the first 300 items of the control mix (public training splits, no evaluation item), run with the release active\\
Means & position-aware, for \mOff: one mean input over last positions, one over all other positions, per matrix; one-mean, for the reported \mRem{} and the one-mean variant: a single mean over all tokens (\cref{tab:releases-variants})\\
Hooks & forward hooks on each changed matrix add the position-aware $b_k$ (\mOff) or subtract the one-mean $b_k$ from the release's output (\mRem); merged LoRA repositories: $\dW=\frac{\alpha}{r}BA$ from the adapter config\\
Item sets & sample: seed 0, up to 60 items per DI benchmark, all JevBench, 300 per private task, reversed copies for four skewed benchmarks (3,557 evaluations); full set: all items (7,356)\\
\midrule
\grouprow{2}{Comparing releases}{}
\mShared{} & average of the offsets of 8 (Qwen3.5-4B) or 6 releases (the others) matrix by matrix, both position means; a matrix a release does not change contributes zero; no other change to the base\\
Releases averaged & rule: regular releases only, at most one per group, none built on another audited release, read out through the model's own head. Left out on Qwen3.5-4B: StartLux-4B-AT (same group and training data as StartLux-4B), zils-mini (starts from the average of d3-mini and torchcast-4B), Nullmega-LoRA (its own joint head, \cref{app:models}). Qwen3.5-4B: all 8 that qualify (H2O-Lightning-4B, d3-mini, EpochEcho-4B, nexus-4B, torchcast-4B, StartLux-4B, rycerzes-4B, karpo-4B); Qwen3.5-2B and Gemma-4-E4B: 8 qualify, the 6 with the largest release gains are averaged (\cref{tab:shared})\\
Exchange tests & three releases per base (\cref{tab:shared})\\
Number averaged & $k\in\{1,2,4\}$, three draws each, \texttt{random.Random(0).sample} over the eight releases\\
Scaled offsets & strengths 1, 0.5, 0.25 for the two large-update exceptions\\
Direction & last-position residual state after every layer on 96 held-back prompts (control-mix items 300--395); layers at 1/4, 1/2, 3/4 and the end of the network\\
Profiles & 200 random splits of every benchmark's items into halves (seed 0); an item and its reversed copy stay together; profiles with reliability below 0.2 left out\\
Output-level baselines & on the base's label log-probabilities (normalised over the item's $K$ labels); \mPrior: subtract the mean log-probability of each label position; \mBias: add a bias per label position fitted by maximum likelihood on gold labels (gradient ascent, 2,000 steps); one vector per $K$ (\mPriorB, \mBiasB: per benchmark and $K$); two folds by a hash of the item id, estimated on one and applied to the other; an item and its reversed copy share a fold\\
Label skew & total-variation distance between the histogram of arg-max labels and of gold labels, per benchmark, averaged over the 36 DI benchmarks\\
Fixed items & an item counts as fixed by a model if its score is above the base's (scores of position-skewed benchmarks are means over two orders)\\
Exception rule & offset gain below $0.5\times\min(\Gr, 8\text{ points})$ on the DI average\\
\midrule
\grouprow{2}{Our adapters (rank-64 LoRA on Qwen3.5-4B)}{}
Target matrices & attention $q,k,v,o$; linear-attention input and output projections; MLP gate, up and down projections, in every layer\\
LoRA & rank 64, $\alpha=r$, no dropout, no bias; merged into the weights for evaluation\\
Loss & cross-entropy of the gold label under the softmax over the item's label-token logits at the answer position (\mLd: cross-entropy to the teacher's distribution)\\
Optimiser & AdamW, learning rate $10^{-4}$, no weight decay, linear warm-up over 5\% of the steps then linear decay to zero, gradient clipping at 1.0\\
Batches & length-bucketed (chunks of 256 items sorted by length), at most 12,000 padded tokens per batch; 2 epochs; seed 0; gradient checkpointing\\
Data & \cref{app:datasets}; prompts at most 1,600 tokens\\
\midrule
\grouprow{2}{Mechanism runs (\cref{tab:mechanism})}{}
Rank 1 & same recipe at rank 1, learning rate $10^{-3}$ (Qwen3.5-4B, LFM3-VL-2B) or $3\times10^{-4}$ (Nemotron-3-Nano-4B); a second seed at the same rate\\
Offset trained directly & one output-bias vector per adapted matrix, no weight update; learning rate $10^{-4}$ (Qwen3.5-4B, Nemotron-3-Nano-4B) or $10^{-5}$ (LFM3-VL-2B); rates of $10^{-3}$ and above collapse below the base\\
Token spans & the adapter active only on: the chat header, record and question; the options; the fixed instruction and answer prefix; or unions of these\\
\midrule
\grouprow{2}{Compute}{}
Hardware & one NVIDIA RTX A4500 (20\,GB)\\
Cost & a split of one release on the sample: one calibration pass and four evaluation passes over 3,557 prompts\\
\bottomrule
\end{longtable}
\endgroup

\section{Prompt Template}\label{app:prompts}

Every model, variant and item uses the one template below, rendered from the code that builds the prompts; the
record, question and options are inserted where shown, the options lettered A, B, \dots{} in their stored order
(reversed for the second order of the skewed benchmarks). The calibration and direction prompts use the same
template on training items.

\begin{promptbox}{P1 \,·\, every model, every item}{one forward pass; no text generated}
\ptag{[USER]}\par
\pslot{record}\par
\smallskip
Question: \pslot{question}\par
\smallskip
Options:\par
A. \pslot{option 1}\par
B. \pslot{option 2}\par
$\ell_K$. \pslot{option $K$}\par
\smallskip
Answer with the label of the best option.\par
\par\smallskip
\ptag{[ASSISTANT]}\enspace\textit{answer position}
\pverdict{argmax over the logits of the $K$ label tokens A, B, \dots{} at the answer position}
\end{promptbox}

\section{Metrics and Statistics}\label{app:metrics}

\textbf{Accuracy.} An item's score is 1 if the arg-max label is correct and 0 otherwise; on a position-skewed
benchmark it is the mean of the scores in the two option orders (\cref{prop:skew}). A benchmark's accuracy is the
mean item score; a group's accuracy (the DI average over 36 benchmarks, JevBench over its three slices, each private
task) is the unweighted mean over its benchmarks. Accuracy is in percent, differences in points; correlations and
cosines are decimals.

\textbf{Intervals.} 95\% paired bootstrap: 1,000 draws (seed 0); in every draw the items of each benchmark are
resampled with replacement, an item and its reversed copy together, and the same draw is used for every variant
stored in one run, so differences such as \mRel{} minus \mBase{} or \mLoo{} minus \mOwn{} are paired. A share
(offset gain over release gain) is computed in every draw; its interval is reported when the release gain's lower
bound exceeds 1 point. Comparisons between runs on different item sets (sample against full set) are not paired.

\textbf{Profiles and direction.} Split-half correlations follow \cref{sec:compare} and \cref{prop:attenuation};
direction cosines compare mean shifts over the 96 prompts, and per prompt (\cref{tab:direction}).

\textbf{Protocol checks.} Position skew: \cref{prop:skew} and \cref{tab:protocol}. Single-class slices:
\cref{prop:singleclass}. Text overlap: a shared sequence of 13 lower-cased word tokens between an evaluation item
(record, question and options) and a training row of the public corpus (\cref{tab:overlap}).

\section{Data Behind the Main-Text Figures}\label{app:figure-results}

\FloatBarrier
\subsection{Offset split of every release (\cref{fig:teaser}a, RQ1)}

The release gains span $+2.2$ to $+24.9$ points and the offset gains of the regular releases $+2.4$ to $+10.7$
(\cref{tab:releases}). The position-aware offset is never more than 0.3 points weaker than the one-mean offset on a
regular release and up to 4.2 points stronger (\cref{tab:releases-variants}); norms and embeddings alone change
little. On the private set the picture is mixed, with large release-specific losses on the topic task for some
releases that their offsets share (\cref{tab:private}). We report this set for completeness and base no claim on it: it
has two tasks from one source, and we cannot tell whether the losses come from the task's label set under the common
prompt or from the releases themselves.

\begingroup\small\setlength{\tabcolsep}{3pt}
\begin{longtable}{@{}lrr@{\,}l>{\columncolor{oursbg}[\tabcolsep][0pt]}r@{}>{\columncolor{oursbg}[0pt][\tabcolsep]\,}lr@{\,}lr@{\,}lrr@{}}
\caption{\textbf{The offset split of every release.} Gains over the base in points on the DI average (36
benchmarks) with 95\% intervals; share: offset gain as a percentage of the release gain (given when the release gain's
interval lies above 1 point, else --); remainder: the release's update minus its offset; JevBench: release and offset gains.
Rel.\ change: median over changed matrices of $\lVert\dW_k\rVert_F/\lVert W_k\rVert_F$ in percent (for the
adapter-only releases, Nullmega-LoRA and cue-reflex, $\dW_k$ is the adapter's $\tfrac{\alpha}{r}B_kA_k$). $^\dagger$: exception (offset gain below half of $\min(\Gr,8)$). Shaded column: the \mOff. \sig{Blue bold} (\signeg{red bold}): 95\% interval above (below) zero.
\hib\,/\,\lob: higher\,/\,lower is better.}\label{tab:releases}\\
\toprule
\rowcolor{hdrbg} & Rel. & \multicolumn{2}{c}{Release} & \multicolumn{2}{c}{\cellcolor{oursbg}Offset} & \multicolumn{2}{c}{Share} & \multicolumn{2}{c}{Remainder} & \multicolumn{2}{c}{JevBench}\\
\rowcolor{hdrbg} Release & change & \multicolumn{2}{c}{gain \hib} & \multicolumn{2}{c}{\cellcolor{oursbg}gain \hib} & \multicolumn{2}{c}{(\%)} & \multicolumn{2}{c}{gain \hib} & Rel. & Off.\\
\midrule
\endfirsthead
\toprule
\rowcolor{hdrbg} Release & Rel. & \multicolumn{2}{c}{Release} & \multicolumn{2}{c}{\cellcolor{oursbg}Offset} & \multicolumn{2}{c}{Share} & \multicolumn{2}{c}{Remainder} & Rel. & Off.\\
\midrule
\endhead
\grouprow{12}{Qwen3.5-4B (DI base 50.4, JevBench base 76.8)}{}
StartLux-4B & 2.53 & \sig{+22.5} & \ci{+20.3}{+24.9} & \sig{+10.7} & \ci{+8.7}{+12.7} & 47 & \ci{40}{55} & \sig{+22.1} & \ci{+19.8}{+24.5} & +15.3 & +12.1\\
StartLux-4B-AT & 4.40 & \sig{+21.8} & \ci{+19.6}{+24.0} & \sig{+10.2} & \ci{+8.1}{+12.4} & 47 & \ci{39}{54} & \sig{+21.4} & \ci{+19.1}{+23.7} & +13.8 & +11.4\\
H2O-Lightning-4B & 6.57 & \sig{+21.2} & \ci{+19.0}{+23.5} & \sig{+8.9} & \ci{+6.9}{+11.0} & 42 & \ci{34}{50} & \sig{+20.2} & \ci{+18.0}{+22.3} & +13.2 & +10.7\\
karpo-4B & 4.18 & \sig{+20.0} & \ci{+17.8}{+22.1} & \sig{+5.7} & \ci{+3.7}{+7.9} & 29 & \ci{19}{38} & \sig{+17.0} & \ci{+14.9}{+19.2} & +9.4 & +5.0\\
zils-mini & 2.09 & \sig{+13.6} & \ci{+11.4}{+15.8} & \sig{+7.6} & \ci{+5.7}{+9.6} & 56 & \ci{45}{67} & \sig{+13.7} & \ci{+11.3}{+15.9} & +15.0 & +10.2\\
d3-mini & 1.12 & \sig{+13.3} & \ci{+10.9}{+15.6} & \sig{+8.0} & \ci{+5.9}{+10.1} & 60 & \ci{49}{72} & \sig{+12.2} & \ci{+10.0}{+14.4} & +9.9 & +9.9\\
EpochEcho-4B & 2.76 & \sig{+12.9} & \ci{+10.6}{+15.2} & \sig{+9.0} & \ci{+6.9}{+11.0} & 70 & \ci{58}{83} & \sig{+11.8} & \ci{+9.6}{+14.0} & +9.1 & +7.8\\
torchcast-4B & 0.93 & \sig{+11.8} & \ci{+9.5}{+14.0} & \sig{+8.4} & \ci{+6.3}{+10.4} & 71 & \ci{59}{85} & \sig{+12.1} & \ci{+10.0}{+14.1} & +15.3 & +11.3\\
rycerzes-4B & 1.25 & \sig{+11.8} & \ci{+9.6}{+13.9} & \sig{+10.3} & \ci{+8.4}{+12.3} & 87 & \ci{76}{101} & \sig{+9.6} & \ci{+7.6}{+11.5} & +8.9 & +8.3\\
nexus-4B & 2.76 & \sig{+8.1} & \ci{+6.1}{+10.1} & \sig{+5.7} & \ci{+3.9}{+7.5} & 71 & \ci{57}{85} & \sig{+6.3} & \ci{+4.3}{+8.4} & +8.3 & +9.4\\
akiray1-bbh$^\dagger$ & 0.07 & \sig{+5.6} & \ci{+3.9}{+7.5} & +0.9 & \ci{$-$0.3}{+2.2} & 16 & \ci{$-$8}{36} & \sig{+4.6} & \ci{+2.9}{+6.3} & +9.3 & +3.6\\
Nullmega-LoRA & 0.40 & \sig{+5.4} & \ci{+3.9}{+7.2} & \sig{+5.2} & \ci{+3.6}{+6.7} & 95 & \ci{78}{118} & \sig{+2.5} & \ci{+1.2}{+3.8} & +6.4 & +7.2\\
Intern-Decision-4B$^\dagger$ & 0.15 & \sig{+4.4} & \ci{+2.6}{+6.2} & \sig{+1.8} & \ci{+0.5}{+3.3} & 42 & \ci{16}{71} & \sig{+3.5} & \ci{+1.9}{+5.1} & +10.1 & +5.0\\
\midrule
\grouprow{12}{Qwen3.5-2B (DI base 43.8, JevBench base 70.3)}{}
joey-1$^\dagger$ & 9.69 & \sig{+24.9} & \ci{+22.7}{+27.2} & \signeg{$-$10.5} & \ci{$-$13.5}{$-$7.6} & $-$42 & \ci{$-$57}{$-$29} & \sig{+14.9} & \ci{+12.3}{+17.2} & +12.8 & $-$35.6\\
StartLux-2B & 3.07 & \sig{+24.3} & \ci{+22.0}{+26.6} & \sig{+8.1} & \ci{+5.9}{+10.4} & 34 & \ci{25}{41} & \sig{+23.8} & \ci{+21.6}{+26.1} & +18.2 & +7.5\\
karpo-2B$^\dagger$ & 5.50 & \sig{+21.7} & \ci{+19.4}{+24.1} & $-$2.3 & \ci{$-$5.2}{+0.3} & $-$11 & \ci{$-$26}{1} & \sig{+16.6} & \ci{+14.0}{+19.2} & +11.5 & $-$4.5\\
KnowLine-2B & 2.53 & \sig{+21.4} & \ci{+19.1}{+23.7} & \sig{+8.7} & \ci{+6.6}{+10.7} & 41 & \ci{32}{49} & \sig{+18.4} & \ci{+16.3}{+20.6} & +12.6 & +7.6\\
d3-nano & 1.17 & \sig{+14.1} & \ci{+11.8}{+16.3} & \sig{+7.3} & \ci{+5.1}{+9.4} & 52 & \ci{39}{65} & \sig{+13.6} & \ci{+11.4}{+15.7} & +9.9 & +6.3\\
fort-1-2B & 2.19 & \sig{+9.3} & \ci{+6.9}{+11.4} & \sig{+6.4} & \ci{+4.3}{+8.3} & 69 & \ci{54}{85} & \sig{+10.4} & \ci{+8.1}{+12.7} & +5.9 & +5.2\\
Jeff-2B & 1.43 & \sig{+7.8} & \ci{+5.4}{+10.0} & \sig{+6.5} & \ci{+4.4}{+8.6} & 83 & \ci{61}{109} & \sig{+9.0} & \ci{+6.7}{+11.2} & +11.2 & +6.2\\
unee-2B & 6.04 & \sig{+7.2} & \ci{+5.1}{+9.6} & \sig{+5.8} & \ci{+3.6}{+8.1} & 81 & \ci{60}{107} & \sig{+6.4} & \ci{+4.2}{+8.8} & +7.2 & +3.1\\
JevK5-2B & 0.35 & \sig{+3.5} & \ci{+1.8}{+5.1} & \sig{+2.4} & \ci{+0.9}{+3.9} & 68 & \ci{37}{96} & \sig{+1.9} & \ci{+0.5}{+3.1} & +4.9 & +3.9\\
classone-2B & 0.23 & \sig{+3.2} & \ci{+1.3}{+5.1} & \sig{+3.2} & \ci{+1.3}{+4.9} & 99 & \ci{65}{162} & +0.3 & \ci{$-$1.0}{+1.5} & +9.0 & +8.2\\
\midrule
\grouprow{12}{Gemma-4-E4B (DI base 51.2, JevBench base 79.0)}{}
H2O-Lightning-E4B & 4.24 & \sig{+16.5} & \ci{+14.3}{+18.6} & \sig{+7.5} & \ci{+5.3}{+9.5} & 45 & \ci{35}{55} & \sig{+16.2} & \ci{+14.0}{+18.4} & +12.6 & +8.1\\
Smaug-Decisions & 3.87 & \sig{+14.3} & \ci{+12.2}{+16.4} & \sig{+7.8} & \ci{+5.9}{+9.9} & 55 & \ci{43}{66} & \sig{+13.7} & \ci{+11.6}{+15.9} & +5.9 & +5.7\\
nischay-1 & 1.46 & \sig{+11.4} & \ci{+9.5}{+13.4} & \sig{+7.9} & \ci{+6.0}{+9.9} & 69 & \ci{57}{81} & \sig{+6.5} & \ci{+4.9}{+8.1} & +5.2 & +5.1\\
linne-sprout & 1.45 & \sig{+9.4} & \ci{+7.4}{+11.4} & \sig{+7.0} & \ci{+5.2}{+9.0} & 74 & \ci{59}{92} & \sig{+9.8} & \ci{+7.8}{+11.9} & +7.0 & +5.1\\
cue-reflex & 0.70 & \sig{+9.0} & \ci{+7.0}{+11.1} & \sig{+7.7} & \ci{+5.8}{+9.6} & 86 & \ci{73}{101} & \sig{+7.6} & \ci{+5.8}{+9.6} & +7.5 & +6.8\\
basal-omni & 1.58 & \sig{+8.4} & \ci{+6.6}{+10.4} & \sig{+4.5} & \ci{+3.0}{+6.0} & 53 & \ci{40}{67} & \sig{+5.5} & \ci{+3.9}{+7.1} & +5.1 & +6.1\\
nirnaya-flash & 4.82 & \sig{+4.4} & \ci{+2.3}{+6.5} & \sig{+3.3} & \ci{+1.3}{+5.1} & 75 & \ci{40}{117} & \sig{+4.7} & \ci{+2.7}{+6.9} & +8.3 & +6.5\\
jez-large & 0.70 & \sig{+2.2} & \ci{+0.7}{+3.7} & \sig{+2.6} & \ci{+1.2}{+4.0} & \multicolumn{2}{c}{--} & +1.7 & \ci{0.0}{+3.3} & +0.5 & +1.9\\
\bottomrule
\end{longtable}
\endgroup

\begin{table}[!htb]
\centering\small
\caption{\textbf{Offset variants.} DI gain in points of the release, of its non-matrix parameters alone (norms,
embeddings), of the offset with one mean input for all positions, and of the position-aware offset used in the
paper (shaded). Point estimates; intervals for the last column in \cref{tab:releases}. \hib\,/\,\lob: higher\,/\,lower is better.}
\label{tab:releases-variants}
\begin{tabular}{@{}lrrr>{\columncolor{oursbg}}r@{}}
\toprule
\rowcolor{hdrbg} Release & Release \hib & Non-matrix only \hib & One-mean offset \hib & \cellcolor{oursbg}Offset \hib\\
\midrule
\grouprow{5}{Qwen3.5-4B}{}
StartLux-4B & +22.5 & +0.3 & +7.3 & +10.7\\
StartLux-4B-AT & +21.8 & +0.2 & +7.1 & +10.2\\
H2O-Lightning-4B & +21.2 & 0.0 & +6.0 & +8.9\\
karpo-4B & +20.0 & +0.9 & +4.7 & +5.7\\
zils-mini & +13.6 & +0.7 & +4.9 & +7.6\\
d3-mini & +13.3 & +0.7 & +4.6 & +8.0\\
EpochEcho-4B & +12.9 & 0.0 & +6.7 & +9.0\\
torchcast-4B & +11.8 & 0.0 & +6.3 & +8.4\\
rycerzes-4B & +11.8 & 0.0 & +8.7 & +10.3\\
nexus-4B & +8.1 & 0.0 & +4.4 & +5.7\\
akiray1-bbh & +5.6 & +0.2 & +0.1 & +0.9\\
Nullmega-LoRA & +5.4 & 0.0 & +5.0 & +5.2\\
Intern-Decision-4B & +4.4 & +0.1 & +2.0 & +1.8\\
\midrule
\grouprow{5}{Qwen3.5-2B}{}
joey-1 & +24.9 & +0.8 & $-$9.9 & $-$10.5\\
StartLux-2B & +24.3 & +0.3 & +6.7 & +8.1\\
karpo-2B & +21.7 & $-$0.3 & $-$2.9 & $-$2.3\\
KnowLine-2B & +21.4 & 0.0 & +8.2 & +8.7\\
d3-nano & +14.1 & +0.1 & +4.3 & +7.3\\
fort-1-2B & +9.3 & +0.1 & +4.1 & +6.4\\
Jeff-2B & +7.8 & +0.7 & +4.0 & +6.5\\
unee-2B & +7.2 & 0.0 & +4.6 & +5.8\\
JevK5-2B & +3.5 & 0.0 & +2.0 & +2.4\\
classone-2B & +3.2 & $-$0.2 & +3.4 & +3.2\\
\midrule
\grouprow{5}{Gemma-4-E4B}{}
H2O-Lightning-E4B & +16.5 & 0.0 & +5.9 & +7.5\\
Smaug-Decisions & +14.3 & $-$0.1 & +6.6 & +7.8\\
nischay-1 & +11.4 & $-$0.3 & +6.4 & +7.9\\
linne-sprout & +9.4 & 0.0 & +2.8 & +7.0\\
cue-reflex & +9.0 & 0.0 & +7.3 & +7.7\\
basal-omni & +8.4 & 0.0 & +3.9 & +4.5\\
nirnaya-flash & +4.4 & 0.0 & +3.3 & +3.3\\
jez-large & +2.2 & 0.0 & +1.9 & +2.6\\
\bottomrule
\end{tabular}
\end{table}

\begingroup\small
\begin{longtable}{@{}lrrrr@{}}
\caption{\textbf{The private two-task set.} Gain over the base in points of each release and of its offset on
the two private classification tasks (300 items each; tax type in both option orders). \hib\,/\,\lob: higher\,/\,lower is better.}\label{tab:private}\\
\toprule
\rowcolor{hdrbg} Release & Topic: release \hib & Topic: offset \hib & Tax type: release \hib & Tax type: offset \hib\\
\midrule
\endfirsthead
\toprule
\rowcolor{hdrbg} Release & Topic: release & Topic: offset & Tax type: release & Tax type: offset\\
\midrule
\endhead
\grouprow{5}{Qwen3.5-4B (base: topic 72.3, tax type 85.3)}{}
H2O-Lightning-4B & +5.3 & +1.7 & +2.7 & +9.0\\
StartLux-4B & +8.3 & +3.7 & +9.7 & $-$0.2\\
StartLux-4B-AT & +1.3 & +1.3 & +10.0 & $-$3.7\\
karpo-4B & $-$1.0 & $-$11.0 & +11.2 & $-$1.3\\
zils-mini & +4.7 & +2.7 & $-$5.0 & +3.0\\
d3-mini & +6.7 & +4.3 & $-$3.8 & $-$6.8\\
EpochEcho-4B & +4.0 & +3.0 & $-$9.0 & $-$17.2\\
torchcast-4B & +4.0 & +4.7 & $-$11.8 & $-$30.2\\
rycerzes-4B & +5.3 & +5.7 & $-$19.2 & $-$19.2\\
nexus-4B & +8.7 & +6.3 & +3.0 & $-$10.5\\
akiray1-bbh & +3.3 & +2.3 & +2.7 & +3.0\\
Nullmega-LoRA & +2.3 & +1.3 & +4.5 & +7.0\\
Intern-Decision-4B & +3.7 & 0.0 & +7.8 & +2.0\\
\midrule
\grouprow{5}{Qwen3.5-2B (base: topic 39.3, tax type 85.3)}{}
joey-1 & $-$3.7 & $-$38.3 & $-$8.7 & $-$35.3\\
StartLux-2B & +27.3 & +10.7 & +8.8 & +0.2\\
karpo-2B & +0.7 & $-$8.3 & $-$28.5 & $-$35.3\\
KnowLine-2B & +35.7 & +17.0 & +0.3 & $-$2.8\\
d3-nano & +26.7 & +11.0 & $-$24.2 & $-$3.5\\
fort-1-2B & +26.0 & +10.0 & $-$14.5 & +0.2\\
Jeff-2B & +20.0 & +6.3 & $-$22.8 & $-$23.7\\
unee-2B & +4.7 & +5.7 & $-$19.3 & $-$24.8\\
JevK5-2B & +4.0 & +9.0 & +2.2 & +3.7\\
classone-2B & +0.7 & +4.3 & +5.2 & +4.7\\
\midrule
\grouprow{5}{Gemma-4-E4B (base: topic 71.7, tax type 93.7)}{}
H2O-Lightning-E4B & +4.3 & $-$0.3 & $-$8.8 & +1.7\\
Smaug-Decisions & $-$16.7 & $-$21.7 & $-$7.8 & $-$7.5\\
nischay-1 & +5.0 & +2.0 & $-$2.3 & 0.0\\
linne-sprout & +3.0 & +2.0 & $-$1.3 & +0.8\\
cue-reflex & +2.3 & +1.0 & $-$0.3 & +0.3\\
basal-omni & 0.0 & +1.3 & $-$3.8 & +1.5\\
nirnaya-flash & $-$65.7 & $-$29.7 & +0.5 & 0.0\\
jez-large & +1.3 & +3.0 & +0.5 & +0.7\\
\bottomrule
\end{longtable}
\endgroup

\FloatBarrier
\subsection{Redundancy of the two parts (\cref{fig:rq1}, RQ1)}

For every release, \cref{tab:redundancy} counts the DI items that the base gets wrong and each model gets right.
The \mRem{} model fixes nearly as many items as the release for the releases with large gains, and most of the
items the \mOff{} model fixes are among them. Removing the offset from a release changes its DI average by $-1.2$ to
$+4.9$ points for the regular releases; the two large-update exceptions lose more ($+5.1$ and $+10.0$). The \mOff{}
model also reduces the label skew of the base, less than the release does.

\begingroup\small\setlength{\tabcolsep}{4pt}
\begin{longtable}{@{}lrr>{\columncolor{oursbg}}rrr@{\,}lrr>{\columncolor{oursbg}}r@{}}
\caption{\textbf{The two parts of a release are redundant.} Fixed: number of DI items that \mBase{} gets wrong
and the model gets right (an item scored in two option orders counts if its score rises). Also \mRem: percentage of
the items the \mOff{} fixes that the \mRem{} fixes too. \mRel{} $-$ \mRem: what removing the offset costs the
release, DI average in points with its 95\% interval. Label skew: total-variation distance between predicted and gold
label histograms, mean over DI benchmarks. $^\dagger$: exception. Shaded columns: the \mOff. \sig{Blue bold} (\signeg{red bold}): 95\% interval above (below) zero. \hib\,/\,\lob: higher\,/\,lower is better.}\label{tab:redundancy}\\
\toprule
\rowcolor{hdrbg} & \multicolumn{3}{c}{Fixed \hib} & Also & \multicolumn{2}{c}{\mRel{} $-$} & \multicolumn{3}{c}{Label skew \lob}\\
\rowcolor{hdrbg} Release & \mRel & \mRem & \cellcolor{oursbg}\mOff & \mRem{} (\%) & \multicolumn{2}{c}{\mRem} & \mRel & \mRem & \cellcolor{oursbg}\mOff\\
\midrule
\endfirsthead
\toprule
\rowcolor{hdrbg} Release & \mRel & \mRem & \cellcolor{oursbg}\mOff & Also (\%) & \multicolumn{2}{c}{\mRel{} $-$ \mRem} & \mRel & \mRem & \cellcolor{oursbg}\mOff\\
\midrule
\endhead
\grouprow{10}{Qwen3.5-4B (1022 DI items that \mBase{} gets wrong; label skew of \mBase{} 0.30)}{}
H2O-Lightning-4B & 586 & 565 & 341 & 86 & \sig{+1.0} & \ci{+0.2}{+1.8} & 0.10 & 0.09 & 0.17\\
StartLux-4B & 609 & 597 & 383 & 88 & +0.4 & \ci{$-$0.6}{+1.3} & 0.10 & 0.10 & 0.15\\
StartLux-4B-AT & 602 & 597 & 366 & 88 & +0.3 & \ci{$-$0.5}{+1.2} & 0.09 & 0.10 & 0.16\\
karpo-4B & 558 & 550 & 298 & 81 & \sig{+2.9} & \ci{+1.5}{+4.4} & 0.12 & 0.14 & 0.21\\
zils-mini & 449 & 448 & 319 & 81 & 0.0 & \ci{$-$1.4}{+1.5} & 0.14 & 0.14 & 0.18\\
d3-mini & 451 & 429 & 329 & 77 & +1.0 & \ci{$-$0.1}{+2.3} & 0.15 & 0.16 & 0.16\\
EpochEcho-4B & 454 & 425 & 366 & 79 & +1.1 & \ci{$-$0.1}{+2.4} & 0.14 & 0.14 & 0.16\\
torchcast-4B & 416 & 423 & 335 & 83 & $-$0.3 & \ci{$-$1.4}{+0.9} & 0.15 & 0.16 & 0.17\\
rycerzes-4B & 414 & 362 & 364 & 84 & \sig{+2.2} & \ci{+0.7}{+3.6} & 0.15 & 0.15 & 0.17\\
nexus-4B & 329 & 307 & 263 & 79 & \sig{+1.8} & \ci{+0.8}{+2.7} & 0.18 & 0.18 & 0.21\\
akiray1-bbh$^\dagger$ & 256 & 234 & 78 & 83 & \sig{+1.1} & \ci{+0.3}{+1.9} & 0.23 & 0.24 & 0.29\\
Nullmega-LoRA & 236 & 127 & 203 & 52 & \sig{+2.9} & \ci{+1.7}{+4.2} & 0.23 & 0.27 & 0.24\\
Intern-Decision-4B$^\dagger$ & 230 & 202 & 115 & 74 & +0.9 & \ci{$-$0.2}{+2.1} & 0.23 & 0.24 & 0.29\\
\midrule
\grouprow{10}{Qwen3.5-2B (1191 DI items that \mBase{} gets wrong; label skew of \mBase{} 0.36)}{}
joey-1$^\dagger$ & 728 & 624 & 442 & 83 & \sig{+10.0} & \ci{+8.0}{+12.0} & 0.12 & 0.30 & 0.67\\
StartLux-2B & 693 & 689 & 449 & 84 & +0.5 & \ci{$-$0.6}{+1.5} & 0.10 & 0.11 & 0.24\\
karpo-2B$^\dagger$ & 666 & 654 & 434 & 82 & \sig{+5.1} & \ci{+2.8}{+7.4} & 0.11 & 0.26 & 0.46\\
KnowLine-2B & 627 & 556 & 397 & 79 & \sig{+2.9} & \ci{+1.7}{+4.2} & 0.14 & 0.17 & 0.26\\
d3-nano & 534 & 514 & 413 & 75 & +0.5 & \ci{$-$1.0}{+1.8} & 0.17 & 0.19 & 0.25\\
fort-1-2B & 497 & 496 & 379 & 87 & $-$1.2 & \ci{$-$2.3}{0.0} & 0.23 & 0.23 & 0.24\\
Jeff-2B & 501 & 502 & 398 & 77 & $-$1.1 & \ci{$-$2.2}{+0.1} & 0.21 & 0.21 & 0.24\\
unee-2B & 467 & 456 & 393 & 79 & +0.8 & \ci{$-$0.4}{+2.0} & 0.23 & 0.25 & 0.26\\
JevK5-2B & 235 & 147 & 193 & 64 & \sig{+1.6} & \ci{+0.4}{+3.0} & 0.33 & 0.37 & 0.32\\
classone-2B & 304 & 118 & 244 & 42 & \sig{+2.9} & \ci{+1.2}{+4.5} & 0.28 & 0.34 & 0.29\\
\midrule
\grouprow{10}{Gemma-4-E4B (1006 DI items that \mBase{} gets wrong; label skew of \mBase{} 0.29)}{}
H2O-Lightning-E4B & 502 & 500 & 316 & 83 & +0.3 & \ci{$-$0.6}{+1.2} & 0.11 & 0.12 & 0.16\\
Smaug-Decisions & 473 & 458 & 339 & 80 & +0.6 & \ci{$-$0.5}{+1.6} & 0.12 & 0.12 & 0.18\\
nischay-1 & 392 & 268 & 310 & 58 & \sig{+4.9} & \ci{+3.3}{+6.5} & 0.15 & 0.23 & 0.17\\
linne-sprout & 384 & 376 & 290 & 89 & $-$0.4 & \ci{$-$1.6}{+0.7} & 0.16 & 0.16 & 0.17\\
cue-reflex & 363 & 316 & 318 & 84 & +1.4 & \ci{0.0}{+2.7} & 0.15 & 0.15 & 0.15\\
basal-omni & 317 & 230 & 200 & 82 & \sig{+2.9} & \ci{+1.6}{+4.2} & 0.21 & 0.22 & 0.24\\
nirnaya-flash & 357 & 333 & 283 & 73 & $-$0.3 & \ci{$-$1.5}{+0.8} & 0.23 & 0.23 & 0.24\\
jez-large & 187 & 162 & 155 & 69 & +0.5 & \ci{$-$0.7}{+1.7} & 0.26 & 0.26 & 0.26\\
\bottomrule
\end{longtable}
\endgroup

\FloatBarrier
\subsection{Shared offset, exchange tests and number of releases (\cref{fig:teaser}b, \cref{fig:rq2}c, RQ2)}

The \mShared{} is significant on all three bases and both benchmark groups (\cref{tab:shared}). In the exchange tests the
others' offset is never significantly worse than a release's own, and significantly better for d3-mini and karpo-4B,
whose own offsets are the weaker ones. Draws of one to four releases give $+5.9$ to $+10.5$ points; the two scaled
exceptions never gain (\cref{tab:kcurve}).

\begin{table}[!htb]
\centering\small
\caption{\textbf{The \mShared{} and the exchange test.} Top: gain in points of the untouched base plus the average
offset of $n$ releases, on the DI average and JevBench. Bottom: the base plus a release's own offset, plus the
average offset of the other releases, and their difference. 95\% paired intervals. \sig{Blue bold} (\signeg{red bold}): 95\% interval above (below) zero. \hib\,/\,\lob: higher\,/\,lower is better.}
\label{tab:shared}
\begin{tabular}{@{}lcr@{\,}lr@{\,}l@{}}
\toprule
\rowcolor{hdrbg} Base & $n$ & \multicolumn{2}{c}{\cellcolor{oursbg}DI gain \hib} & \multicolumn{2}{c}{\cellcolor{oursbg}JevBench gain \hib}\\
\midrule
\rowcolor{oursbg}Qwen3.5-4B & 8 & \phantom{\sig{+10.5}} & \llap{\sig{+10.5}\,}\ci{+8.4}{+12.6} & \phantom{\sig{+10.6}} & \llap{\sig{+10.6}\,}\ci{+6.1}{+15.3}\\
\rowcolor{oursbg}Qwen3.5-2B & 6 & \phantom{\sig{+8.8}} & \llap{\sig{+8.8}\,}\ci{+6.5}{+11.1} & \phantom{\sig{+7.4}} & \llap{\sig{+7.4}\,}\ci{+2.4}{+12.7}\\
\rowcolor{oursbg}Gemma-4-E4B & 6 & \phantom{\sig{+8.8}} & \llap{\sig{+8.8}\,}\ci{+6.9}{+10.8} & \phantom{\sig{+6.5}} & \llap{\sig{+6.5}\,}\ci{+2.7}{+10.3}\\
\bottomrule
\end{tabular}\\[6pt]
\begin{tabular}{@{}lr@{\,}lr@{\,}lr@{\,}l@{}}
\toprule
\rowcolor{hdrbg} Release & \multicolumn{2}{c}{\mOwn{} \hib} & \multicolumn{2}{c}{\cellcolor{oursbg}\mLoo{} \hib} & \multicolumn{2}{c}{Others $-$ own}\\
\midrule
\grouprow{7}{Qwen3.5-4B}{}
H2O-Lightning-4B & \sig{+8.9} & \ci{+6.9}{+11.0} & \sig{+10.2} & \ci{+8.1}{+12.3} & +1.3 & \ci{0.0}{+2.8}\\
d3-mini & \sig{+8.4} & \ci{+6.3}{+10.4} & \sig{+10.1} & \ci{+8.1}{+12.2} & \sig{+1.8} & \ci{+0.6}{+3.0}\\
karpo-4B & \sig{+4.8} & \ci{+2.7}{+6.8} & \sig{+10.3} & \ci{+8.1}{+12.4} & \sig{+5.5} & \ci{+3.9}{+7.2}\\
\midrule
\grouprow{7}{Qwen3.5-2B}{}
StartLux-2B & \sig{+8.1} & \ci{+5.8}{+10.3} & \sig{+8.5} & \ci{+6.4}{+10.7} & +0.4 & \ci{$-$0.7}{+1.5}\\
KnowLine-2B & \sig{+8.8} & \ci{+6.7}{+10.8} & \sig{+8.0} & \ci{+5.7}{+10.1} & $-$0.8 & \ci{$-$2.4}{+0.9}\\
d3-nano & \sig{+7.1} & \ci{+5.0}{+9.3} & \sig{+8.1} & \ci{+6.0}{+10.2} & +1.0 & \ci{$-$0.3}{+2.2}\\
\midrule
\grouprow{7}{Gemma-4-E4B}{}
H2O-Lightning-E4B & \sig{+7.5} & \ci{+5.3}{+9.5} & \sig{+8.8} & \ci{+6.9}{+10.7} & +1.4 & \ci{0.0}{+2.7}\\
nischay-1 & \sig{+7.9} & \ci{+6.0}{+9.8} & \sig{+9.0} & \ci{+7.0}{+10.9} & +1.1 & \ci{$-$0.3}{+2.4}\\
linne-sprout & \sig{+7.0} & \ci{+5.2}{+9.0} & \sig{+8.3} & \ci{+6.4}{+10.2} & +1.3 & \ci{$-$0.2}{+2.7}\\
\bottomrule
\end{tabular}
\end{table}

\begin{table}[!htb]
\centering\small
\caption{\textbf{How many releases the \mShared{} needs (Qwen3.5-4B), and scaled offsets of the two large-update
exceptions (Qwen3.5-2B).} Gain in points of the base plus the average offset of $k$ releases drawn at random
(seed 0; a draw can repeat), or plus a release's own offset multiplied by a strength; $k=8$: the average of all eight offsets, no random draw
(Draw --). 95\% paired intervals.
\sig{Blue bold} (\signeg{red bold}): 95\% interval above (below) zero. \hib\,/\,\lob: higher\,/\,lower is better.}
\label{tab:kcurve}
\begin{tabular}{@{}cclr@{\,}lr@{\,}l@{}}
\toprule
\rowcolor{hdrbg} $k$ & Draw & Releases averaged & \multicolumn{2}{c}{DI gain \hib} & \multicolumn{2}{c}{JevBench gain \hib}\\
\midrule
1 & 1 & rycerzes-4B & \sig{+10.3} & \ci{+8.4}{+12.3} & \sig{+8.3} & \ci{+3.0}{+13.7}\\
1 & 2 & rycerzes-4B & \sig{+10.3} & \ci{+8.4}{+12.3} & \sig{+8.3} & \ci{+3.0}{+13.7}\\
1 & 3 & H2O-Lightning-4B & \sig{+8.9} & \ci{+6.9}{+11.0} & \sig{+10.7} & \ci{+6.5}{+14.9}\\
2 & 1 & torchcast-4B, karpo-4B & \sig{+8.0} & \ci{+5.9}{+10.0} & \sig{+8.9} & \ci{+4.8}{+13.3}\\
2 & 2 & karpo-4B, nexus-4B & \sig{+5.9} & \ci{+4.0}{+8.0} & \sig{+8.6} & \ci{+4.6}{+12.9}\\
2 & 3 & torchcast-4B, nexus-4B & \sig{+8.4} & \ci{+6.4}{+10.4} & \sig{+10.8} & \ci{+6.7}{+15.2}\\
4 & 1 & StartLux-4B, torchcast-4B, d3-mini, rycerzes-4B & \sig{+10.4} & \ci{+8.4}{+12.5} & \sig{+10.3} & \ci{+5.7}{+15.0}\\
4 & 2 & EpochEcho-4B, karpo-4B, d3-mini, H2O-Lightning-4B & \sig{+8.4} & \ci{+6.2}{+10.6} & \sig{+10.3} & \ci{+6.0}{+14.9}\\
4 & 3 & torchcast-4B, karpo-4B, StartLux-4B, rycerzes-4B & \sig{+10.3} & \ci{+8.1}{+12.5} & \sig{+11.0} & \ci{+6.7}{+15.4}\\
\midrule 8 & -- & all eight & \sig{+10.5} & \ci{+8.4}{+12.6} & \sig{+10.6} & \ci{+6.1}{+15.3}\\
\bottomrule
\end{tabular}\\[6pt]
\begin{tabular}{@{}lcr@{\,}lr@{\,}l@{}}
\toprule
\rowcolor{hdrbg} Release & Strength & \multicolumn{2}{c}{DI gain \hib} & \multicolumn{2}{c}{JevBench gain \hib}\\
\midrule
joey-1 & 1 & \signeg{$-$11.2} & \ci{$-$13.9}{$-$8.4} & \signeg{$-$40.5} & \ci{$-$49.5}{$-$31.1}\\
joey-1 & 0.5 & \signeg{$-$3.8} & \ci{$-$6.5}{$-$1.2} & $-$2.8 & \ci{$-$9.3}{+3.8}\\
joey-1 & 0.25 & $-$0.9 & \ci{$-$3.5}{+1.8} & +2.9 & \ci{$-$2.1}{+7.8}\\
karpo-2B & 1 & \signeg{$-$2.9} & \ci{$-$5.7}{$-$0.3} & $-$1.0 & \ci{$-$7.4}{+5.2}\\
karpo-2B & 0.5 & +0.7 & \ci{$-$1.6}{+2.9} & +2.9 & \ci{$-$1.2}{+7.1}\\
karpo-2B & 0.25 & $-$0.1 & \ci{$-$1.6}{+1.4} & +1.6 & \ci{$-$1.6}{+4.6}\\
\bottomrule
\end{tabular}
\end{table}

\FloatBarrier
\subsection{Output-level corrections (\cref{fig:teaser}b, RQ2)}

\Cref{tab:calibration} compares the \mShared{} with four corrections of the base's label logits, paired on the same
items. The corrections that a single constant at the output can express (\mPrior, \mBias) gain at most 1.8 points.
The per-benchmark corrections, which use the identity of the benchmark, gain more on Qwen3.5-4B than on the other
bases and stay 5.5 to 7.7 points below the \mShared{} everywhere.

\begin{table}[!htb]
\centering\small
\caption{\textbf{The \mShared{} is not an output-level correction.} DI gain over \mBase{} in points, 95\% paired
intervals, on the item set of the \mShared{} runs. \mPrior: subtract each label position's mean log-probability
(label-free); \mBias: add a bias per label position fitted on gold labels; both use one vector per number of
options. \mPriorB, \mBiasB: the same, estimated per benchmark. Each is estimated on one half of the items and applied to
the other. Last column: the \mShared{} minus the row. Label skew: total-variation distance between predicted and
gold label histograms, mean over DI benchmarks. \sig{Blue bold} (\signeg{red bold}): 95\% interval above (below) zero. \hib\,/\,\lob: higher\,/\,lower is better.}
\label{tab:calibration}
\begin{tabular}{@{}llr@{\,}lr@{\,}l@{}}
\toprule
\rowcolor{hdrbg} Model & Acts on & \multicolumn{2}{c}{DI gain \hib} & \multicolumn{2}{c}{\mShared{} minus row \hib}\\
\midrule
\grouprow{6}{Qwen3.5-4B (label skew: \mBase{} 0.30, \mShared{} 0.17)}{}
\mPrior & label logits & +0.7 & \ci{$-$0.4}{+1.9} & \sig{+9.7} & \ci{+7.7}{+11.7}\\
\mBias & label logits & +0.1 & \ci{$-$1.3}{+1.4} & \sig{+10.4} & \ci{+8.3}{+12.4}\\
\mPriorB & label logits & \sig{+4.4} & \ci{+2.8}{+5.9} & \sig{+6.0} & \ci{+4.0}{+8.0}\\
\mBiasB & label logits & \sig{+5.0} & \ci{+2.8}{+7.0} & \sig{+5.5} & \ci{+3.3}{+7.6}\\
\mShared{} + \mPriorB & inside the network & \sig{+10.5} & \ci{+8.4}{+12.6} & \multicolumn{2}{c}{--}\\
\rowcolor{oursbg}\mShared & inside the network & \phantom{\sig{+10.5}} & \llap{\sig{+10.5}\,}\ci{+8.4}{+12.6} & \multicolumn{2}{c}{--}\\
\midrule
\grouprow{6}{Qwen3.5-2B (label skew: \mBase{} 0.36, \mShared{} 0.22)}{}
\mPrior & label logits & \sig{+1.8} & \ci{+0.3}{+3.2} & \sig{+7.0} & \ci{+5.0}{+9.2}\\
\mBias & label logits & +1.2 & \ci{$-$0.3}{+2.6} & \sig{+7.6} & \ci{+5.6}{+9.7}\\
\mPriorB & label logits & \sig{+2.6} & \ci{+0.3}{+4.5} & \sig{+6.2} & \ci{+4.1}{+8.5}\\
\mBiasB & label logits & \sig{+2.9} & \ci{+0.7}{+4.9} & \sig{+5.9} & \ci{+3.8}{+8.1}\\
\mShared{} + \mPriorB & inside the network & \sig{+8.8} & \ci{+6.7}{+11.0} & \multicolumn{2}{c}{--}\\
\rowcolor{oursbg}\mShared & inside the network & \phantom{\sig{+8.8}} & \llap{\sig{+8.8}\,}\ci{+6.5}{+11.1} & \multicolumn{2}{c}{--}\\
\midrule
\grouprow{6}{Gemma-4-E4B (label skew: \mBase{} 0.29, \mShared{} 0.16)}{}
\mPrior & label logits & +0.6 & \ci{$-$0.7}{+1.8} & \sig{+8.3} & \ci{+6.5}{+10.1}\\
\mBias & label logits & +0.2 & \ci{$-$1.2}{+1.7} & \sig{+8.6} & \ci{+6.8}{+10.5}\\
\mPriorB & label logits & +1.1 & \ci{$-$0.5}{+2.6} & \sig{+7.7} & \ci{+5.8}{+9.6}\\
\mBiasB & label logits & +1.3 & \ci{$-$0.5}{+3.1} & \sig{+7.5} & \ci{+5.6}{+9.5}\\
\mShared{} + \mPriorB & inside the network & \sig{+9.2} & \ci{+7.2}{+11.2} & \multicolumn{2}{c}{--}\\
\rowcolor{oursbg}\mShared & inside the network & \phantom{\sig{+8.8}} & \llap{\sig{+8.8}\,}\ci{+6.9}{+10.8} & \multicolumn{2}{c}{--}\\
\bottomrule
\end{tabular}
\end{table}

\FloatBarrier
\subsection{Profiles and direction (\cref{fig:rq2}a,b, RQ2)}

Uncorrected correlations of offset profiles look high (0.54 to 0.79 on average, \cref{tab:profiles}), and the
split-half estimate is lower; what matters is its ratio to the noise ceiling, 0.64 to 0.85 for the offset against
0.51 to 0.69 for whole releases and 0.23 to 0.30 for the extra gain, among regular releases. Offset shifts are most aligned from the middle of the network on;
on Gemma-4-E4B the first quarter of the layers shows little alignment (\cref{tab:direction}). The benchmarks the offset
helps most are listed in \cref{tab:benchgain}.

\begin{table}[!htb]
\centering\small
\caption{\textbf{Per-benchmark profiles agree across releases for the offset, more than for whole releases, and
little for the extra gain.} Pearson correlation over the 36 DI benchmarks of a per-benchmark gain: \mRel{},
\mRem{} and \mOff{} model minus \mBase; extra gain: \mRel{} minus \mOff. Reliability: one release on two
disjoint halves of the items (range over releases); naive: two releases on all items (inflated by the shared base
run); split-half: two releases on disjoint halves, mean (range) over pairs; noise ceiling: mean geometric mean of the
two reliabilities; ratio: split-half mean / noise-ceiling mean (1: the same profile up to noise). Releases with
reliability below 0.2 are left out; regular: without the four exceptions. 200 random
splits. Shaded cells: the \mOff.}
\label{tab:profiles}
\begin{tabular}{@{}lllccccc@{}}
\toprule
\rowcolor{hdrbg} Base & Profile & Releases & Reliability & Naive & Split-half & Noise ceiling & Ratio\\
\midrule
Qwen3.5-4B & \mRel & all (13) & 0.35--0.82 & 0.61 & 0.38 (0.14--0.79) & 0.60 & 0.64\\
Qwen3.5-4B & \mRel & regular (11) & 0.55--0.82 & 0.67 & 0.44 (0.19--0.79) & 0.64 & 0.69\\
Qwen3.5-4B & \mRem & all (13) & 0.36--0.82 & 0.60 & 0.37 (0.11--0.79) & 0.59 & 0.63\\
Qwen3.5-4B & \mRem & regular (11) & 0.36--0.82 & 0.64 & 0.42 (0.11--0.79) & 0.62 & 0.67\\
Qwen3.5-4B & Extra gain & all (13) & 0.29--0.79 & 0.14 & 0.09 ($-$0.32--0.74) & 0.50 & 0.18\\
Qwen3.5-4B & Extra gain & regular (11) & 0.29--0.79 & 0.20 & 0.12 ($-$0.32--0.74) & 0.52 & 0.23\\
Qwen3.5-4B & \mOff & all (12) & 0.25--0.67 & 0.68 & 0.38 ($-$0.08--0.64) & 0.52 & \cellcolor{oursbg}0.74\\
Qwen3.5-4B & \mOff & regular (11) & 0.35--0.67 & 0.79 & 0.46 (0.09--0.64) & 0.55 & \cellcolor{oursbg}0.85\\
\midrule
Qwen3.5-2B & \mRel & all (10) & 0.32--0.78 & 0.58 & 0.39 (0.07--0.74) & 0.64 & 0.61\\
Qwen3.5-2B & \mRel & regular (8) & 0.32--0.78 & 0.57 & 0.36 (0.07--0.74) & 0.61 & 0.58\\
Qwen3.5-2B & \mRem & all (9) & 0.40--0.79 & 0.58 & 0.39 (0.15--0.73) & 0.65 & 0.60\\
Qwen3.5-2B & \mRem & regular (7) & 0.40--0.79 & 0.64 & 0.43 (0.26--0.73) & 0.62 & 0.68\\
Qwen3.5-2B & Extra gain & all (10) & 0.28--0.86 & 0.26 & 0.19 ($-$0.30--0.70) & 0.56 & 0.34\\
Qwen3.5-2B & Extra gain & regular (8) & 0.28--0.78 & 0.23 & 0.15 ($-$0.10--0.70) & 0.51 & 0.30\\
Qwen3.5-2B & \mOff & all (10) & 0.28--0.77 & 0.54 & 0.23 ($-$0.01--0.46) & 0.47 & \cellcolor{oursbg}0.49\\
Qwen3.5-2B & \mOff & regular (8) & 0.28--0.62 & 0.70 & 0.31 (0.19--0.46) & 0.43 & \cellcolor{oursbg}0.73\\
\midrule
Gemma-4-E4B & \mRel & all (8) & 0.47--0.87 & 0.51 & 0.32 (0.02--0.66) & 0.64 & 0.51\\
Gemma-4-E4B & \mRel & regular (8) & 0.47--0.87 & 0.51 & 0.32 (0.02--0.66) & 0.64 & 0.51\\
Gemma-4-E4B & \mRem & all (8) & 0.32--0.84 & 0.50 & 0.30 (0.01--0.60) & 0.59 & 0.51\\
Gemma-4-E4B & \mRem & regular (8) & 0.32--0.84 & 0.50 & 0.30 (0.01--0.60) & 0.59 & 0.51\\
Gemma-4-E4B & Extra gain & all (7) & 0.24--0.76 & 0.21 & 0.12 ($-$0.13--0.69) & 0.44 & 0.27\\
Gemma-4-E4B & Extra gain & regular (7) & 0.24--0.76 & 0.21 & 0.12 ($-$0.13--0.69) & 0.44 & 0.27\\
Gemma-4-E4B & \mOff & all (8) & 0.34--0.81 & 0.62 & 0.30 (0.17--0.44) & 0.47 & \cellcolor{oursbg}0.64\\
Gemma-4-E4B & \mOff & regular (8) & 0.34--0.81 & 0.62 & 0.30 (0.17--0.44) & 0.47 & \cellcolor{oursbg}0.64\\
\bottomrule
\end{tabular}
\end{table}

\begin{table}[!htb]
\centering\small
\caption{\textbf{Offset shifts point the same way across releases.} Cosine between two models' mean change of
the residual state at the answer position, averaged over pairs: offset shift (\mOff{} minus \mBase) among all
decision releases and among regular ones; per prompt: the cosine on each prompt, averaged; extra shift: \mRel{}
minus \mOff; controls: range of the cosine between a non-decision fine-tune's offset shift and a regular
release's; size: norm of the mean offset shift over the mean norm of the base state.}
\label{tab:direction}
\begin{tabular}{@{}lccccccc@{}}
\toprule
\rowcolor{hdrbg} & \multicolumn{3}{c}{Offset shift} & \multicolumn{2}{c}{Extra shift} & Controls & \\
\rowcolor{hdrbg} Layer & All & \cellcolor{oursbg}Regular & Per prompt & Regular & Per prompt & vs regular & Size\\
\midrule
\grouprow{8}{Qwen3.5-4B (10 releases, 4 controls, 96 prompts)}{}
8 / 32 & 0.37 & \cellcolor{oursbg}0.46 & 0.46 & 0.33 & 0.29 & $-$0.06-$-$0.47 & 0.49\\
16 / 32 & 0.59 & \cellcolor{oursbg}0.65 & 0.62 & 0.49 & 0.37 & $-$0.05-$-$0.33 & 0.61\\
24 / 32 & 0.67 & \cellcolor{oursbg}0.70 & 0.68 & 0.48 & 0.35 & 0.02-$-$0.55 & 0.62\\
32 / 32 & 0.62 & \cellcolor{oursbg}0.63 & 0.60 & 0.58 & 0.43 & $-$0.31-$-$0.40 & 2.63\\
\midrule
\grouprow{8}{Qwen3.5-2B (8 releases, 4 controls, 96 prompts)}{}
6 / 24 & 0.66 & \cellcolor{oursbg}0.69 & 0.69 & 0.54 & 0.49 & 0.12-$-$0.68 & 0.76\\
12 / 24 & 0.73 & \cellcolor{oursbg}0.78 & 0.76 & 0.54 & 0.44 & 0.16-$-$0.75 & 1.14\\
18 / 24 & 0.77 & \cellcolor{oursbg}0.84 & 0.81 & 0.56 & 0.40 & 0.15-$-$0.75 & 1.35\\
24 / 24 & 0.84 & \cellcolor{oursbg}0.85 & 0.80 & 0.73 & 0.49 & $-$0.06-$-$0.73 & 7.07\\
\midrule
\grouprow{8}{Gemma-4-E4B (6 releases, 4 controls, 96 prompts)}{}
10 / 42 & 0.24 & \cellcolor{oursbg}0.24 & 0.24 & 0.37 & 0.30 & $-$0.13-$-$0.28 & 0.04\\
21 / 42 & 0.69 & \cellcolor{oursbg}0.69 & 0.65 & 0.46 & 0.33 & $-$0.19-$-$0.41 & 0.67\\
31 / 42 & 0.83 & \cellcolor{oursbg}0.83 & 0.76 & 0.48 & 0.29 & $-$0.16-$-$0.62 & 0.32\\
42 / 42 & 0.81 & \cellcolor{oursbg}0.81 & 0.77 & 0.37 & 0.23 & $-$0.62-$-$0.72 & 0.64\\
\bottomrule
\end{tabular}
\end{table}

\begingroup\small
\begin{longtable}{@{}lrrr@{}}
\caption{\textbf{Which benchmarks the offset helps.} Offset gain in points per DI benchmark, mean over the
releases of each base, sorted by the mean over bases. \hib\,/\,\lob: higher\,/\,lower is better.}\label{tab:benchgain}\\
\toprule
\rowcolor{hdrbg} Benchmark & Qwen3.5-4B \hib & Qwen3.5-2B \hib & Gemma-4-E4B \hib\\
\midrule
\endfirsthead
\toprule
\rowcolor{hdrbg} Benchmark & Qwen3.5-4B & Qwen3.5-2B & Gemma-4-E4B\\
\midrule
\endhead
BPoMP & +29.0 & +9.8 & +24.4\\
Home appliance simulator & +32.6 & +21.8 & +8.3\\
ContractNLI & +18.7 & +32.7 & +6.2\\
cfcolor & +13.2 & +15.3 & +19.8\\
BBH fixed-option tasks & +16.8 & +1.2 & +25.8\\
iSarcasmEval & +10.1 & +14.8 & +10.5\\
SATA-Bench & +13.1 & +5.7 & +13.5\\
POP909-CL & +19.2 & +0.8 & +8.3\\
MMLU-Pro & +12.6 & +1.5 & +10.8\\
New Yorker caption matching & +15.0 & $-$4.7 & +12.3\\
VAST & +8.5 & +7.2 & +4.0\\
ARC-Challenge & +5.4 & $-$1.8 & +14.2\\
When2Call MCQ & +4.6 & +1.0 & +11.5\\
Amazon ESCI & +2.1 & +10.5 & +4.4\\
BANKING77 & +3.6 & +16.5 & $-$3.3\\
MMLU & +4.2 & $-$0.8 & +11.9\\
CLadder & +16.4 & $-$10.5 & +7.5\\
ChessBench & +2.4 & +2.7 & +8.1\\
Habermas Machine & +5.5 & +9.8 & $-$4.0\\
RAGTruth & +1.0 & +0.2 & +9.7\\
NLI4CT & +1.9 & +6.8 & +1.7\\
MuSR & +6.2 & $-$2.5 & +5.4\\
FinEntity & +14.1 & $-$9.2 & +4.0\\
HLE & +7.3 & $-$2.7 & +2.9\\
CRUXEval & $-$4.5 & +8.8 & +1.3\\
Humicroedit & +5.1 & $-$0.8 & +1.0\\
GPQA Diamond & +4.4 & $-$4.8 & +5.4\\
BFCL & +1.3 & +4.2 & $-$0.8\\
SimpleBench & $-$8.5 & +6.0 & +5.0\\
RouterBench & +0.3 & +1.8 & +0.4\\
GSM8K & +3.6 & +4.3 & $-$7.9\\
ARC-Easy & $-$1.0 & $-$6.2 & +3.8\\
PhishNChips phishing decisions & $-$0.1 & $-$6.8 & +1.9\\
CLINC150+OOS & +0.1 & +2.3 & $-$8.1\\
HoVer claim verification & $-$5.5 & $-$5.3 & +3.3\\
SGD/SGD-X & $-$2.9 & $-$1.5 & $-$6.0\\
\bottomrule
\end{longtable}
\endgroup

\FloatBarrier
\subsection{Our adapters and the non-decision fine-tunes (\cref{fig:rq3}, RQ3)}

\begin{table}[!htb]
\centering\small\setlength{\tabcolsep}{3pt}
\caption{\textbf{Our label-only adapters stay at the offset.} Rank-64 LoRA on Qwen3.5-4B, full item set (7,356
evaluations; DI base 51.2). Gains in points with 95\% intervals; share
in percent. \sig{Blue bold} (\signeg{red bold}): 95\% interval above (below) zero. Shaded columns: the \mOff. \hib\,/\,\lob: higher\,/\,lower is better.}
\label{tab:adapters}
\begin{tabular}{@{}l>{\raggedright\arraybackslash}p{4.3cm}r@{\,}l>{\columncolor{oursbg}[\tabcolsep][0pt]}r@{}>{\columncolor{oursbg}[0pt][\tabcolsep]\,}lr@{\,}lrr@{}}
\toprule
\rowcolor{hdrbg} Model & Data & \multicolumn{2}{c}{DI gain \hib} & \multicolumn{2}{c}{\cellcolor{oursbg}Offset gain \hib} & \multicolumn{2}{c}{Share (\%)} & \multicolumn{2}{c}{JevBench}\\
\rowcolor{hdrbg} & & & & \cellcolor{oursbg} & \cellcolor{oursbg} & & & Model & Offset\\
\midrule
LoRA-8k & 8k items, 14 tasks & \sig{+9.9} & \ci{+8.2}{+11.6} & \sig{+7.4} & \ci{+5.8}{+9.0} & 75 & \ci{64}{85} & +6.9 & +6.4\\
LoRA-Varied & 32k items, 38 tasks & \sig{+9.0} & \ci{+7.2}{+10.8} & \sig{+7.4} & \ci{+5.9}{+9.0} & 83 & \ci{71}{95} & +6.1 & +9.1\\
LoRA-Clean & 32k items, 668 tasks, no benchmark family & \sig{+9.1} & \ci{+7.5}{+10.9} & \sig{+8.3} & \ci{+6.6}{+10.1} & 91 & \ci{78}{107} & +9.2 & +10.2\\
LoRA-Exposed & LoRA-Clean data with 11 benchmark families & \sig{+10.8} & \ci{+9.0}{+12.9} & \sig{+7.6} & \ci{+5.9}{+9.3} & 70 & \ci{60}{80} & +9.5 & +9.2\\
LoRA-Distill & LoRA-Clean items, H2O-Lightning-4B's answer distributions & \sig{+11.3} & \ci{+9.6}{+13.2} & \sig{+7.8} & \ci{+6.2}{+9.3} & 68 & \ci{59}{79} & +9.8 & +9.9\\
H2O-Lightning-4B & released; reference on the same item set & \sig{+19.9} & \ci{+18.3}{+21.5} & \sig{+7.7} & \ci{+6.1}{+9.4} & 39 & \ci{32}{46} & +12.9 & +11.0\\
\bottomrule
\end{tabular}
\end{table}

\begin{table}[!htb]
\centering\small
\caption{\textbf{Non-decision fine-tunes have no offset; excluded releases.} DI gain over the base in points,
95\% intervals. \sig{Blue bold} (\signeg{red bold}): 95\% interval above (below) zero. Shaded column: the \mOff. \hib\,/\,\lob: higher\,/\,lower is better.}
\label{tab:controls}
\begin{tabular}{@{}llr@{\,}l>{\columncolor{oursbg}[\tabcolsep][0pt]}r@{}>{\columncolor{oursbg}[0pt][\tabcolsep]\,}l@{}}
\toprule
\rowcolor{hdrbg} Model & Purpose or reason & \multicolumn{2}{c}{Release gain \hib} & \multicolumn{2}{c}{\cellcolor{oursbg}Offset gain \hib}\\
\midrule
\grouprow{6}{Non-decision fine-tunes, Qwen3.5-4B}{}
3D-Coder & code & +0.5 & \ci{$-$1.2}{+2.2} & \sig{+2.0} & \ci{+0.6}{+3.3}\\
Abliterated-4B & refusal removal & \signeg{$-$6.1} & \ci{$-$8.4}{$-$3.8} & \signeg{$-$2.2} & \ci{$-$3.8}{$-$0.5}\\
Distill-4B & reasoning distillation & \signeg{$-$3.5} & \ci{$-$5.8}{$-$1.4} & $-$0.5 & \ci{$-$1.8}{+0.9}\\
Super-Coder & code & +1.9 & \ci{$-$0.3}{+4.1} & +0.7 & \ci{$-$0.5}{+2.0}\\
\midrule
\grouprow{6}{Non-decision fine-tunes, Qwen3.5-2B}{}
SQL-Distill-2B & text-to-SQL distillation & \signeg{$-$3.9} & \ci{$-$6.1}{$-$1.7} & \signeg{$-$2.4} & \ci{$-$4.0}{$-$0.8}\\
MedVL-2B & medical vision-language & \sig{+2.8} & \ci{+1.0}{+4.7} & +0.7 & \ci{$-$0.6}{+2.0}\\
Abliterated-2B & refusal removal & $-$0.4 & \ci{$-$1.7}{+0.9} & 0.0 & \ci{$-$1.4}{+1.3}\\
Distill-2B & reasoning distillation & +0.5 & \ci{$-$1.7}{+2.6} & +0.9 & \ci{$-$0.9}{+2.6}\\
\midrule
\grouprow{6}{Non-decision fine-tunes, Gemma-4-E4B}{}
Bolt-E4B & general assistant & \sig{+2.6} & \ci{+1.1}{+4.2} & \sig{+1.9} & \ci{+0.7}{+3.1}\\
ContextPilot-E4B & context management & +0.4 & \ci{$-$0.9}{+1.7} & +0.6 & \ci{$-$0.5}{+1.6}\\
Arabic-E4B & Arabic & +1.6 & \ci{$-$0.3}{+3.8} & +0.2 & \ci{$-$1.2}{+1.6}\\
Abliterated-E4B & refusal removal & \signeg{$-$2.0} & \ci{$-$3.7}{$-$0.3} & \signeg{$-$1.5} & \ci{$-$3.0}{$-$0.1}\\
\midrule
\grouprow{6}{Releases run but not analysed, Qwen3.5-2B}{}
Jev-Style-2B-v3 & below the base under the common prompt & \signeg{$-$7.3} & \ci{$-$10.3}{$-$4.5} & \sig{+2.6} & \ci{+0.3}{+5.1}\\
Jev-Style-2B-v4 & below the base under the common prompt & \signeg{$-$11.7} & \ci{$-$14.4}{$-$8.9} & \signeg{$-$8.7} & \ci{$-$11.3}{$-$5.9}\\
Jev-Style-2B-v4.2 & below the base under the common prompt & \signeg{$-$9.8} & \ci{$-$12.5}{$-$7.0} & \signeg{$-$6.6} & \ci{$-$9.4}{$-$3.7}\\
\midrule
\grouprow{6}{Releases run but not analysed, Gemma-4-E4B}{}
Decision-1-E4B & no gain under the common readout & 0.0 & \ci{$-$1.6}{+1.6} & +1.0 & \ci{$-$0.4}{+2.5}\\
typed-decisions-de & no gain under the common readout & +0.4 & \ci{$-$1.1}{+1.7} & +0.7 & \ci{$-$0.5}{+1.7}\\
classone-E4B & below the base under the common prompt & \signeg{$-$15.1} & \ci{$-$17.4}{$-$12.7} & \signeg{$-$2.8} & \ci{$-$4.7}{$-$0.9}\\
Avrik-Lite & text weights identical to the base & 0.0 & \ci{0.0}{0.0} & 0.0 & \ci{0.0}{0.0}\\
\bottomrule
\end{tabular}
\end{table}

\begin{table}[!htb]
\centering\small\setlength{\tabcolsep}{3pt}
\caption{\textbf{The sample and the full item set agree.} DI gains in points on the 3,557-evaluation sample and
on all 7,356 evaluations (95\% intervals on the full set), and the offset share on the full set. \sig{Blue bold} (\signeg{red bold}): 95\% interval above (below) zero. \hib\,/\,\lob: higher\,/\,lower is better.}
\label{tab:fullset}
\begin{tabular}{@{}llrr@{\,}l>{\columncolor{oursbg}}r>{\columncolor{oursbg}[\tabcolsep][0pt]}r@{}>{\columncolor{oursbg}[0pt][\tabcolsep]\,}lr@{\,}l@{}}
\toprule
\rowcolor{hdrbg} & & \multicolumn{3}{c}{Release gain \hib} & \multicolumn{3}{c}{\cellcolor{oursbg}Offset gain \hib} & \multicolumn{2}{c}{Share}\\
\rowcolor{hdrbg} Release & Base & Sample & \multicolumn{2}{c}{Full set} & \cellcolor{oursbg}Sample & \multicolumn{2}{c}{\cellcolor{oursbg}Full set} & \multicolumn{2}{c}{(\%)}\\
\midrule
karpo-4B & Qwen3.5-4B & +20.0 & \sig{+19.5} & \ci{+17.8}{+21.3} & +5.7 & \sig{+4.9} & \ci{+3.1}{+6.7} & 25 & \ci{17}{33}\\
StartLux-4B & Qwen3.5-4B & +22.5 & \sig{+21.4} & \ci{+19.6}{+23.2} & +10.7 & \sig{+9.8} & \ci{+8.2}{+11.4} & 46 & \ci{40}{52}\\
d3-mini & Qwen3.5-4B & +13.3 & \sig{+12.2} & \ci{+10.2}{+14.1} & +8.0 & \sig{+7.4} & \ci{+5.8}{+9.1} & 61 & \ci{51}{72}\\
KnowLine-2B & Qwen3.5-2B & +21.4 & \sig{+21.4} & \ci{+19.7}{+23.1} & +8.7 & \sig{+7.7} & \ci{+6.2}{+9.4} & 36 & \ci{30}{42}\\
StartLux-2B & Qwen3.5-2B & +24.3 & \sig{+24.2} & \ci{+22.4}{+26.0} & +8.1 & \sig{+8.4} & \ci{+6.8}{+10.2} & 35 & \ci{29}{41}\\
Smaug-Decisions & Gemma-4-E4B & +14.3 & \sig{+14.3} & \ci{+12.9}{+15.9} & +7.8 & \sig{+7.4} & \ci{+5.9}{+8.9} & 52 & \ci{43}{60}\\
H2O-Lightning-E4B & Gemma-4-E4B & +16.5 & \sig{+16.0} & \ci{+14.6}{+17.4} & +7.5 & \sig{+6.9} & \ci{+5.5}{+8.5} & 43 & \ci{36}{51}\\
\bottomrule
\end{tabular}
\end{table}

\section{Further Analyses}\label{app:further}

\subsection{What a label-only adapter does}\label{app:mechanism}

On three bases, a rank-1 adapter trained on the \mLk{} data at a higher learning rate matches rank 64 within
1.6 points on the DI average (\cref{tab:mechanism}). Its position-aware offset keeps 75--100\% of its DI gain, and an
offset trained directly, with no weight update, 63--81\%. The adapter needs to act only on the options and the fixed
instruction to keep 81--97\% of its DI gain; acting only on the record and question keeps 22--45\%. On Qwen3.5-4B the
first half of the layers carries nearly all of it. These runs use a 6,551-item audit set in one option
order; \cref{tab:mechanism} lists the runs cited here.

\begingroup\small
\begin{longtable}{@{}lrrrr@{}}
\caption{\textbf{What a label-only adapter does, on three bases.} Our adapter on the \mLk{} data,
6,551-item audit set in the original option order. Accuracy in percent on the three trained tasks (ANLI, HellaSwag,
WinoGrande), the DI average (36 held-out benchmarks) and JevBench; share: DI gain as a percentage of the rank-1
adapter's DI gain (--: the base, which has no gain). Rank 1 uses its best learning rate. Rank 1 minus rank 64 on the DI average, points: Qwen3.5-4B $-$1.0 \ci{$-$2.2}{+0.2}; Nemotron-3-Nano-4B $-$1.5 \ci{$-$2.6}{$-$0.6}; LFM3-VL-2B +0.6 \ci{$-$0.5}{+1.8}. Shaded cells: the \mOff. \hib\,/\,\lob: higher\,/\,lower is better.}\label{tab:mechanism}\\
\toprule
\rowcolor{hdrbg} Model & Trained tasks \hib & DI \hib & JevBench \hib & Share (\%)\\
\midrule
\endfirsthead
\toprule
\rowcolor{hdrbg} Model & Trained tasks \hib & DI \hib & JevBench \hib & Share (\%)\\
\midrule
\endhead
\grouprow{5}{Qwen3.5-4B (rank-1 learning rate 0.001; trained offset 0.0001)}{}
\mBase & 55.6 & 51.1 & 76.8 & --\\
Rank 64 & 77.8 & 61.1 & 84.0 & 111\\
Rank 1 & 78.6 & 60.1 & 84.0 & 100\\
\cellcolor{oursbg}Offset of the rank-1 adapter & 72.2 & 57.9 & 86.5 & 75\\
Offset trained directly & 73.1 & 58.0 & 83.6 & 77\\
Offset trained directly, second seed & 75.6 & 57.0 & 84.6 & 66\\
Rank 1, active on record + question only & 69.2 & 53.5 & 81.6 & 27\\
Rank 1, active on options + instruction only & 71.1 & 58.4 & 86.0 & 81\\
Rank 1, layers 0--15 only & 78.3 & 59.9 & 85.2 & 97\\
\midrule
\grouprow{5}{Nemotron-3-Nano-4B (rank-1 learning rate 0.0003; trained offset 0.0001)}{}
\mBase & 53.3 & 50.7 & 79.1 & --\\
Rank 64 & 67.2 & 55.2 & 79.7 & 152\\
Rank 1 & 66.7 & 53.7 & 79.9 & 100\\
\cellcolor{oursbg}Offset of the rank-1 adapter & 63.3 & 53.7 & 80.9 & 100\\
Offset trained directly & 60.6 & 53.1 & 78.5 & 81\\
Offset trained directly, second seed & 60.6 & 53.1 & 77.5 & 81\\
Rank 1, active on record + question only & 60.3 & 52.0 & 80.2 & 45\\
Rank 1, active on options + instruction only & 63.6 & 53.6 & 81.5 & 97\\
\midrule
\grouprow{5}{LFM3-VL-2B (rank-1 learning rate 0.001; trained offset 1e-05)}{}
\mBase & 45.3 & 44.4 & 69.6 & --\\
Rank 64 & 61.1 & 51.7 & 78.0 & 93\\
Rank 1 & 61.1 & 52.3 & 76.7 & 100\\
\cellcolor{oursbg}Offset of the rank-1 adapter & 57.8 & 51.5 & 78.0 & 91\\
Offset trained directly & 57.8 & 49.7 & 74.6 & 67\\
Offset trained directly, second seed & 58.3 & 49.3 & 74.2 & 63\\
Rank 1, active on record + question only & 46.4 & 46.1 & 72.1 & 22\\
Rank 1, active on options + instruction only & 59.4 & 50.9 & 78.6 & 83\\
\bottomrule
\end{longtable}
\endgroup

\FloatBarrier
\subsection{Merging releases}\label{app:merges}

If the extra gains of different releases were independent improvements, merging the releases would add them. We
test this with weight-space merges of releases of one base and no training: the base plus a weighted sum of the
releases' weight changes (\cref{tab:merges}).

\textbf{The gains do not add.} On Qwen3.5-4B, the equal-weight average of two, three and five releases is 1.9, 3.1
and 4.7 points below H2O-Lightning-4B alone on the DI average. On Qwen3.5-2B, the average of joey-1 and KnowLine-2B
is 1.8 points above joey-1 alone, although the two releases gain 24.9 and 21.5 points over the base; the average of
five is 1.8 points below it. On both bases the result falls as more releases enter the average. This agrees with
RQ4: what a release adds beyond the plateau is specific to that release.

\textbf{Where a second release helps depends on the layers.} Adding joey-1 at half weight to KnowLine-2B raises the
DI average by 4.2 points when it enters layers 0--11 and leaves it unchanged in layers 12--23, as for our adapter,
whose gain sits in the first half of the layers (\cref{app:mechanism}). On Qwen3.5-4B, adding d3-mini to
H2O-Lightning-4B moves the DI average in neither half and raises JevBench by 1.9 points in layers 16--31. The table
shows one pair of releases per base and both halves of each split.

\begin{table}[!htb]
\centering\small
\caption{\textbf{Merging releases of one base.} Weight-space merges of released models, with no training:
the base plus a weighted sum of the releases' weight changes. Accuracy in percent on the DI average
(36 held-out benchmarks) and JevBench, 3,557 evaluations per model; $\Delta$: difference in points
from the reference named in the group row, with its 95\% interval. Levels of single releases differ from
\cref{tab:releases} by at most 0.3 points (separate runs). \sig{Blue bold} (\signeg{red bold}): 95\% interval above (below) zero. \hib\,/\,\lob: higher\,/\,lower is better.}
\label{tab:merges}
\begin{tabular}{@{}lrr@{\,}lrr@{\,}l@{}}
\toprule
\rowcolor{hdrbg} Model & DI \hib & \multicolumn{2}{c}{$\Delta$ DI} & JevBench \hib & \multicolumn{2}{c}{$\Delta$ JevBench}\\
\midrule
\grouprow{7}{Qwen3.5-4B: single releases and equal-weight averages, against H2O-Lightning-4B}{}
\mBase & 50.4 & \signeg{$-$21.1} & \ci{$-$23.4}{$-$18.9} & 76.8 & \signeg{$-$13.2} & \ci{$-$18.7}{$-$8.0}\\
H2O-Lightning-4B & 71.5 & \multicolumn{2}{c}{--} & 90.0 & \multicolumn{2}{c}{--}\\
d3-mini & 63.4 & \signeg{$-$8.1} & \ci{$-$10.2}{$-$6.1} & 86.6 & $-$3.4 & \ci{$-$6.9}{+0.5}\\
torchcast-4B & 62.2 & \signeg{$-$9.3} & \ci{$-$11.2}{$-$7.3} & 92.1 & +2.1 & \ci{$-$0.4}{+4.7}\\
Average of 2 (H2O-Lightning-4B, d3-mini) & 69.6 & \signeg{$-$1.9} & \ci{$-$3.2}{$-$0.6} & 89.4 & $-$0.6 & \ci{$-$3.6}{+2.3}\\
Average of 3 (+ torchcast-4B) & 68.4 & \signeg{$-$3.1} & \ci{$-$4.7}{$-$1.5} & 89.4 & $-$0.6 & \ci{$-$3.2}{+1.9}\\
Average of 5 (+ nexus-4B, EpochEcho-4B) & 66.8 & \signeg{$-$4.7} & \ci{$-$6.4}{$-$2.9} & 89.8 & $-$0.2 & \ci{$-$3.2}{+2.9}\\
\midrule
\grouprow{7}{Qwen3.5-4B: H2O-Lightning-4B plus $0.5\times$ d3-mini in half of the layers, against H2O-Lightning-4B}{}
d3-mini added in layers 0--15 & 71.9 & +0.4 & \ci{$-$0.5}{+1.4} & 89.7 & $-$0.3 & \ci{$-$1.9}{+1.3}\\
d3-mini added in layers 16--31 & 71.4 & $-$0.2 & \ci{$-$0.8}{+0.5} & 92.0 & \sig{+1.9} & \ci{+0.6}{+3.6}\\
\midrule
\grouprow{7}{Qwen3.5-2B: single releases and equal-weight averages, against joey-1}{}
\mBase & 43.8 & \signeg{$-$24.9} & \ci{$-$27.3}{$-$22.7} & 70.3 & \signeg{$-$12.3} & \ci{$-$17.4}{$-$6.9}\\
joey-1 & 68.7 & \multicolumn{2}{c}{--} & 82.6 & \multicolumn{2}{c}{--}\\
KnowLine-2B & 65.3 & \signeg{$-$3.4} & \ci{$-$5.4}{$-$1.4} & 82.8 & +0.3 & \ci{$-$3.9}{+4.9}\\
d3-nano & 58.0 & \signeg{$-$10.8} & \ci{$-$12.9}{$-$8.7} & 80.4 & $-$2.1 & \ci{$-$6.3}{+2.3}\\
Average of 2 (joey-1, KnowLine-2B) & 70.6 & \sig{+1.8} & \ci{+0.3}{+3.2} & 85.7 & +3.1 & \ci{$-$0.1}{+6.3}\\
Average of 3 (+ d3-nano) & 70.2 & +1.5 & \ci{$-$0.3}{+3.0} & 85.5 & +3.0 & \ci{$-$0.4}{+6.4}\\
Average of 5 (+ fort-1-2B, Jeff-2B) & 66.9 & $-$1.8 & \ci{$-$3.8}{+0.1} & 83.6 & +1.0 & \ci{$-$2.9}{+5.3}\\
\midrule
\grouprow{7}{Qwen3.5-2B: KnowLine-2B plus $0.5\times$ joey-1 in half of the layers, against KnowLine-2B}{}
joey-1 added in layers 0--11 & 69.5 & \sig{+4.2} & \ci{+2.4}{+6.0} & 83.3 & +0.4 & \ci{$-$3.2}{+3.9}\\
joey-1 added in layers 12--23 & 64.9 & $-$0.4 & \ci{$-$1.8}{+1.1} & 84.7 & +1.9 & \ci{$-$0.7}{+4.7}\\
\bottomrule
\end{tabular}
\end{table}

\FloatBarrier
\subsection{Corpus overlap and the evaluation protocol}\label{app:protocol}

\begin{table}[!htb]
\centering\small
\caption{\textbf{Text overlap with the public training corpus, and where the extra gain lies.} Left: share of a
benchmark's evaluation items (percent) that share a 13-word sequence with a training row of the public decision
corpus; 38 further benchmarks below 5\%; 441 of all evaluation items in total. Right: extra
gain (\mRel{} minus \mOff, points, mean over benchmarks) of each regular release on the 7
benchmarks with at least 20\% overlap and on all other DI and trained-task benchmarks; higher
on the overlap benchmarks for 21 of 27 releases (means
+6.9 and +5.0).}
\label{tab:overlap}
\begin{tabular}[t]{@{}lr@{}}
\toprule
\rowcolor{hdrbg} Benchmark & Overlap (\%)\\
\midrule
ContractNLI & 98 \\
ANLI & 69 \\
NLI4CT & 42 \\
RAGTruth response-level hallucination & 36 \\
SATA-Bench & 32 \\
HoVer claim verification & 26 \\
WinoGrande & 24 \\
RouterBench & 11 \\
MMLU & 10 \\
Amazon ESCI & 5\\
\bottomrule
\end{tabular}\hspace{2em}
\begin{tabular}[t]{@{}lrr@{}}
\toprule
\rowcolor{hdrbg} Release & Overlap bench. & Others\\
\midrule
\grouprow{3}{Qwen3.5-4B}{}
H2O-Lightning-4B & +13.7 & +12.5\\
StartLux-4B & +14.8 & +11.3\\
StartLux-4B-AT & +10.1 & +11.3\\
karpo-4B & +20.6 & +13.0\\
zils-mini & +7.3 & +6.1\\
d3-mini & +8.2 & +5.3\\
EpochEcho-4B & +5.2 & +4.1\\
torchcast-4B & +6.9 & +2.8\\
rycerzes-4B & +0.4 & +1.5\\
nexus-4B & +1.1 & +2.6\\
Nullmega-LoRA & +0.8 & +0.1\\
\midrule
\grouprow{3}{Qwen3.5-2B}{}
StartLux-2B & +19.2 & +15.3\\
KnowLine-2B & +14.3 & +12.5\\
d3-nano & +7.9 & +6.7\\
fort-1-2B & +3.6 & +2.7\\
Jeff-2B & +5.6 & +0.3\\
unee-2B & +3.1 & +0.9\\
JevK5-2B & +1.2 & +1.2\\
classone-2B & $-$1.0 & $-$0.2\\
\midrule
\grouprow{3}{Gemma-4-E4B}{}
H2O-Lightning-E4B & +14.2 & +8.2\\
Smaug-Decisions & +11.2 & +5.6\\
nischay-1 & +6.9 & +3.2\\
linne-sprout & +7.3 & +1.6\\
cue-reflex & $-$0.1 & +1.2\\
basal-omni & 0.0 & +4.3\\
nirnaya-flash & +1.8 & +0.9\\
jez-large & +1.8 & $-$0.6\\
\bottomrule
\end{tabular}
\end{table}

\begin{table}[!htb]
\centering\small
\caption{\textbf{Option order changes regression verdicts in both directions.} Releases whose list of
regressions (a benchmark at least 5 points below the base) differs between the
original option order and the average over both orders, on the four position-skewed benchmarks: 11
of 31 releases; 3 regressions removed and 8 added by averaging (totals
91 and 96). --: no regression that only this order finds. Single-class slices (share of the most frequent gold label):
ACOS: 99\% of 120 items; BRIGHT: 100\% of 115 items; ToolRet: 100\% of 117 items.}
\label{tab:protocol}
\begin{tabular}{@{}llp{4.3cm}p{5.2cm}@{}}
\toprule
\rowcolor{hdrbg} Release & Base & Only in the original order & Only with orders averaged\\
\midrule
H2O-Lightning-4B & Qwen3.5-4B & private: tax type & --\\
zils-mini & Qwen3.5-4B & -- & private: tax type\\
EpochEcho-4B & Qwen3.5-4B & -- & private: tax type\\
torchcast-4B & Qwen3.5-4B & -- & private: tax type\\
Nullmega-LoRA & Qwen3.5-4B & Amazon ESCI & --\\
d3-nano & Qwen3.5-2B & -- & private: tax type\\
fort-1-2B & Qwen3.5-2B & -- & private: tax type\\
Jeff-2B & Qwen3.5-2B & -- & private: tax type\\
classone-2B & Qwen3.5-2B & RAGTruth & --\\
H2O-Lightning-E4B & Gemma-4-E4B & -- & private: tax type\\
Smaug-Decisions & Gemma-4-E4B & -- & private: tax type\\
\bottomrule
\end{tabular}
\end{table}

\FloatBarrier

\section{Qualitative Examples}\label{app:qualitative}

Items drawn at random (seed 0) from Qwen3.5-4B's sample, among public-benchmark items that show one behaviour each:
the base is wrong and the \mShared{} and almost every release's offset are right (228 candidate items, two shown);
the base and H2O-Lightning-4B's own offset are wrong and only H2O-Lightning-4B's full update and at most two other
releases are right (59 items, two shown); and the base answers the same position in both option orders on a skewed
benchmark (53 items, one shown). Each card shows the item as every model sees it on the left, with the correct option marked
\exgold{like this}, and on the right the answer and its probability for four models, \qok{blue} if correct and
\qbad{orange} if wrong, followed by how many of the other regular releases on Qwen3.5-4B answer correctly with their
full update and with their offset alone. Texts are shown as stored, shortened with ``…''.

\begin{excard}{A}{JevBench-original}{A constant is enough}{oursfg}
\exhead{Record}
\exrecord{The final signed order says courier, despite an old email saying post.}
\exhead{Question}
Extract the final confirmed delivery method. Ignore cancelled plans and hypothetical alternatives. Choose unknown if no final method is confirmed.\par\vspace{3pt}
\exhead{Options}
{\footnotesize\raggedright \exgold{A.~courier: Courier delivery}\\ B.~pickup: Customer pickup\\ C.~post: Postal service\\ D.~unknown: No final confirmed method\par}
\tcblower
\exhead{Answer (probability)}
\begin{tabular}{@{}l@{\hspace{6pt}}l@{}}
\mBase & \qbad{D (58\%)}\\
\mShared & \qok{A (95\%)}\\
H2O-Lightning-4B: \mOff & \qok{A (92\%)}\\
H2O-Lightning-4B: \mRel & \qok{A (92\%)}\\
\end{tabular}
\par\vspace{4pt}\exhead{Other 10 releases, correct}
\begin{tabular}{@{}l@{\hspace{6pt}}l@{}}
\mRel & \exbar{10}{10}\\
\mOff & \exbar{10}{10}\\
\end{tabular}
\par\vspace{5pt}{\raggedright\textbf{Reading.} The base is wrong; a constant that depends on nothing in the item makes it right, here and for 10 of 10 other releases' offsets (\cref{prop:exact}c).\par}
\end{excard}

\begin{excard}{B}{FinEntity}{A constant is enough}{oursfg}
\exhead{Record}
\exrecord{\{ "document": "\textbackslash{}"We're seeing economic conditions deteriorate in China, the United States and Europe. That's definitely weighing on Asian manufacturing activity,\textbackslash{}" said Toru Nishihama, chief economist at Dai-ichi Life Research Institute.", "task": "FinEntity entity-given sentiment" \}}
\exhead{Question}
Classify the sentiment toward the supplied financial entity span. The entity span is given explicitly as 187:219 (Dai-ichi Life Research Institute); do not extract or alter it. Choose exactly one of Negative, Neutral, or Positive.\par\vspace{3pt}
\exhead{Options}
{\footnotesize\raggedright A.~Negative: Negative\\ \exgold{B.~Neutral: Neutral}\\ C.~Positive: Positive\par}
\tcblower
\exhead{Answer (probability)}
\begin{tabular}{@{}l@{\hspace{6pt}}l@{}}
\mBase & \qbad{A (72\%)}\\
\mShared & \qok{B (66\%)}\\
H2O-Lightning-4B: \mOff & \qok{B (72\%)}\\
H2O-Lightning-4B: \mRel & \qok{B (89\%)}\\
\end{tabular}
\par\vspace{4pt}\exhead{Other 10 releases, correct}
\begin{tabular}{@{}l@{\hspace{6pt}}l@{}}
\mRel & \exbar{9}{10}\\
\mOff & \exbar{8}{10}\\
\end{tabular}
\par\vspace{5pt}{\raggedright\textbf{Reading.} The base is wrong; a constant that depends on nothing in the item makes it right, here and for 8 of 10 other releases' offsets (\cref{prop:exact}c).\par}
\end{excard}

\begin{excard}{C}{MMLU-Pro}{Only the full release is right}{qcwrongfg}
\exhead{Record}
\exrecord{A solution contains 1 mg per ml of myosin and 10\textasciicircum{}14 latex particles per ml. When a given volume of this solution is dried on a grid and viewed under the electron microscope, a typical field contains 122 protein molecules and 10 latex particles. Calculate the molecular weight of myosin.}
\exhead{Question}
Which option is the correct answer?\par\vspace{3pt}
\exhead{Options}
{\footnotesize\raggedright A.~A: 5.93 × 10\textasciicircum{}6 g / mole;\enspace B.~B: 7.93 × 10\textasciicircum{}5 g / mole;\enspace C.~C: 4.93 × 10\textasciicircum{}6 g / mole;\enspace D.~D: 2.93 × 10\textasciicircum{}5 g / mole;\enspace E.~E: 4.93 × 10\textasciicircum{}4 g / mole;\enspace F.~F: 5.93 × 10\textasciicircum{}5 g / mole;\enspace G.~G: 3.93 × 10\textasciicircum{}6 g / mole;\enspace H.~H: 3.93 × 10\textasciicircum{}5 g / mole;\enspace I.~I: 6.93 × 10\textasciicircum{}5 g / mole;\enspace \exgold{J.~J: 4.93 × 10\textasciicircum{}5 g / mole}\par}
\tcblower
\exhead{Answer (probability)}
\begin{tabular}{@{}l@{\hspace{6pt}}l@{}}
\mBase & \qbad{A (46\%)}\\
\mShared & \qbad{F (35\%)}\\
H2O-Lightning-4B: \mOff & \qbad{F (30\%)}\\
H2O-Lightning-4B: \mRel & \qok{J (70\%)}\\
\end{tabular}
\par\vspace{4pt}\exhead{Other 10 releases, correct}
\begin{tabular}{@{}l@{\hspace{6pt}}l@{}}
\mRel & \exbar{2}{10}\\
\mOff & \exbar{1}{10}\\
\end{tabular}
\par\vspace{5pt}{\raggedright\textbf{Reading.} No offset reaches this item, and only 2 of 10 other releases do: the extra gain of \cref{sec:rq4} is specific to a release.\par}
\end{excard}

\begin{excard}{D}{JevBench-hard}{Only the full release is right}{qcwrongfg}
\exhead{Record}
\exrecord{\{ "event": "The reply arrived Wednesday March 11, 2026 at 13:30 New York local time.", "rules": "A support window is measured in New York local time. It opens at 09:00 on the first business day after a notice is received and lasts exactly 52 hours of elapsed time. Business days only determine opening; after opening, weekends do not pause it. Notice was received Friday March 6, 2026 at 18:10 New York time. Saturday …}
\exhead{Question}
Apply all stated timing, unit, inclusion, correction, and threshold rules; select the exact band.\par\vspace{3pt}
\exhead{Options}
{\footnotesize\raggedright A.~before\_open: The event occurred before the window opened.\\ B.~within\_window: The event occurred from opening through the exact …\\ \exgold{C.~late\_by\_under\_2h: The event occurred after closing but less than two …}\\ D.~late\_by\_2h\_or\_more: The event was at least two hours late.\par}
\tcblower
\exhead{Answer (probability)}
\begin{tabular}{@{}l@{\hspace{6pt}}l@{}}
\mBase & \qbad{A (46\%)}\\
\mShared & \qbad{B (42\%)}\\
H2O-Lightning-4B: \mOff & \qbad{B (49\%)}\\
H2O-Lightning-4B: \mRel & \qok{C (29\%)}\\
\end{tabular}
\par\vspace{4pt}\exhead{Other 10 releases, correct}
\begin{tabular}{@{}l@{\hspace{6pt}}l@{}}
\mRel & \exbar{1}{10}\\
\mOff & \exbar{0}{10}\\
\end{tabular}
\par\vspace{5pt}{\raggedright\textbf{Reading.} No offset reaches this item, and only 1 of 10 other releases do: the extra gain of \cref{sec:rq4} is specific to a release.\par}
\end{excard}

\begin{excard}{E}{iSarcasmEval}{Answering by position}{naivefg}
\exhead{Record}
\exrecord{5 Guys is the most overrated food place I've ever been to , absolute rip off}
\exhead{Question}
Is this text intended to be sarcastic?\par\vspace{3pt}
\exhead{Options}
{\footnotesize\raggedright \exgold{A.~no: No}\\ B.~yes: Yes\par}
\tcblower
\exhead{Answer (probability)}
\begin{tabular}{@{}l@{\hspace{6pt}}l@{}}
\mBase & \qok{A (65\%)}\\
\mShared & \qok{A (80\%)}\\
H2O-Lightning-4B: \mOff & \qok{A (72\%)}\\
H2O-Lightning-4B: \mRel & \qok{A (87\%)}\\
\end{tabular}\par\vspace{4pt}\exhead{Options reversed}
\begin{tabular}{@{}l@{\hspace{6pt}}l@{}}
\mBase & \qbad{A (62\%)}\\
H2O-Lightning-4B: \mOff & \qok{B (82\%)}\\
H2O-Lightning-4B: \mRel & \qok{B (86\%)}\\
\end{tabular}
\par\vspace{4pt}\exhead{Other 10 releases, correct}
\begin{tabular}{@{}l@{\hspace{6pt}}l@{}}
\mRel & \exbar{10}{10}\\
\mOff & \exbar{10}{10}\\
\end{tabular}
\par\vspace{5pt}{\raggedright\textbf{Reading.} The base keeps the same position when the options are reversed, so the original order alone measures a position preference (\cref{prop:skew}).\par}
\end{excard}

\section{LLM Usage}\label{app:llm}

\paragraph{LLMs as research assistance.} An AI assistant (Claude), following instructions prepared by the authors, helped draft and edit the text from the authors' experiment reports, wrote the scripts that regenerate all tables and figures from the result files, and checked every number against those files and every reference against its source. The authors reviewed and revised all of this output. The research questions, experiments, analysis code and conclusions are the authors'. The proofs were drafted by the authors and formalized with the assistant's help in Lean~4 with Mathlib, where every result is machine-checked, so their correctness does not depend on the assistant. The authors take full responsibility for the paper's content.

\section{Artifact Licenses}\label{app:licenses}

\begingroup\small
\begin{longtable}{@{}p{4.4cm}lp{3.9cm}p{4.3cm}c@{}}
\caption{\textbf{Artifacts and licenses}, as stated on each model's Hugging Face page (API, retrieved
2026-11-19) or in each repository's license file. $^\ast$no license tag on the model page; used for analysis only, no
weights redistributed. $^\ddagger$gated, non-commercial license; used for analysis only. $^\dagger$the dataset card
names per-source licenses of its constituent datasets. Source --: not public. No third-party weights or data are redistributed with this
paper; the code downloads them.}\label{tab:licenses}\\
\toprule
\rowcolor{hdrbg} Name & Type & License & Use & Source\\
\midrule
\endfirsthead
\toprule
\rowcolor{hdrbg} Name & Type & License & Use & Source\\
\midrule
\endhead
\grouprow{5}{Bases}{}
Qwen3.5-4B \citep{qwen2026qwen35} & model & apache-2.0 & base & \href{https://huggingface.co/Qwen/Qwen3.5-4B}{Link}\\
Qwen3.5-2B \citep{qwen2026qwen35} & model & apache-2.0 & base & \href{https://huggingface.co/Qwen/Qwen3.5-2B}{Link}\\
Gemma-4-E4B \citep{gemma2026gemma4} & model & apache-2.0 & base & \href{https://huggingface.co/google/gemma-4-E4B-it}{Link}\\
Nemotron-3-Nano-4B \citep{taghibakhshi2025elastic} & model & nvidia-nemotron-open-model-license & base, mechanism runs only & \href{https://huggingface.co/nvidia/NVIDIA-Nemotron-3-Nano-4B-BF16}{Link}\\
LFM3-VL-2B \citep{liquidai2026lfm3} & model & lfm1.0 & base, mechanism runs only & \href{https://huggingface.co/LiquidAI/LFM3-VL-2B}{Link}\\
\midrule
\grouprow{5}{Released decision models}{}
H2O-Lightning-4B \citep{h2o2026h2olightning4b} & model & apache-2.0 & audited release & \href{https://huggingface.co/h2oai/h2o-lightning-4b}{Link}\\
StartLux-4B \citep{zhuang2026startlux} & model & cc-by-nc-4.0 & audited release & \href{https://huggingface.co/startlux-models/StartLux-Decision-v1.1-4B-Instant}{Link}\\
StartLux-4B-AT \citep{zhuang2026startlux} & model & cc-by-nc-4.0 & audited release & \href{https://huggingface.co/startlux-models/StartLux-Decision-v1.1-4B-AutoThink}{Link}\\
karpo-4B \citep{machinepulse2026karpo4b} & model & apache-2.0 & audited release & \href{https://huggingface.co/MachinePulseAI/karpo-decision-4B}{Link}\\
zils-mini \citep{qhoenix2026zilsmini} & model & apache-2.0 & audited release & \href{https://huggingface.co/qhoenix/zils-mini-decision-t6p}{Link}\\
d3-mini \citep{vllm2026d3mini} & model & apache-2.0 & audited release & \href{https://huggingface.co/vllm-sr/d3-mini}{Link}\\
EpochEcho-4B \citep{epochecho2026epochecho4b} & model & apache-2.0 & audited release & \href{https://huggingface.co/EpochEcho/decision-4b-preview}{Link}\\
torchcast-4B \citep{torchcast2026torchcast4b} & model & apache-2.0 & audited release & \href{https://huggingface.co/torchcast-ai/torchcast-decision-4b}{Link}\\
rycerzes-4B \citep{rycerzes2026rycerzes4b} & model & apache-2.0 & audited release & \href{https://huggingface.co/rycerzes/Qwen3.5-4B-Decision}{Link}\\
nexus-4B \citep{zackyalgiffari2026nexus4b} & model & apache-2.0 & audited release & \href{https://huggingface.co/zackyalgiffari/nexus-decision-4b}{Link}\\
akiray1-bbh \citep{akiray2026akiray1bbh} & model & none stated$^\ast$ & audited release & \href{https://huggingface.co/akiray1/qwen3.5-4b-decision-bbh-1m-cosine-epoch2}{Link}\\
Nullmega-LoRA \citep{nullmega2026nullmegalora} & model & none stated$^\ast$ & audited release & \href{https://huggingface.co/Nullmega/decision-qwen3.5-4B_lora}{Link}\\
Intern-Decision-4B \citep{zhu2026interndecision} & model & apache-2.0 & audited release & \href{https://huggingface.co/internlm/Intern-Decision-4B}{Link}\\
joey-1 \citep{base2026joey1} & model & mit & audited release & \href{https://huggingface.co/baselabs/joey-1}{Link}\\
StartLux-2B \citep{zhuang2026startlux} & model & cc-by-nc-4.0 & audited release & \href{https://huggingface.co/startlux-models/StartLux-Decision-v1.1-2B-Instant}{Link}\\
karpo-2B \citep{machinepulse2026karpo2b} & model & apache-2.0 & audited release & \href{https://huggingface.co/MachinePulseAI/karpo-decision-2B}{Link}\\
KnowLine-2B \citep{pelaai2026knowline2b} & model & apache-2.0 & audited release & \href{https://huggingface.co/PelaAI/KnowLine-2B-Gen1}{Link}\\
d3-nano \citep{vllm2026d3nano} & model & apache-2.0 & audited release & \href{https://huggingface.co/vllm-sr/d3-nano}{Link}\\
fort-1-2B \citep{teximal2026fort12b} & model & apache-2.0 & audited release & \href{https://huggingface.co/teximal/fort-1-2b}{Link}\\
Jeff-2B \citep{jefflegacy2026jeff2b} & model & apache-2.0 & audited release & \href{https://huggingface.co/jeff-legacy/Jeff-Qwen3.5-2B}{Link}\\
unee-2B \citep{mumthas2026unee2b} & model & apache-2.0 & audited release & \href{https://huggingface.co/uneeverse/unee-2b}{Link}\\
JevK5-2B \citep{alibiserikbay2026jevk52b} & model & apache-2.0 & audited release & \href{https://huggingface.co/alibiserikbay/JevK5-2B}{Link}\\
classone-2B \citep{gonzaga2026classone2b} & model & apache-2.0 & audited release & \href{https://huggingface.co/devops-thiago/classone-qwen3.5-2b}{Link}\\
H2O-Lightning-E4B \citep{h2o2026h2olightninge4b} & model & h2o-model-non-commercial-licence-v1.1$^\ddagger$ & audited release & \href{https://huggingface.co/h2oai/h2o-lightning-e4b}{Link}\\
Smaug-Decisions \citep{abacus2026smaugdecisions} & model & apache-2.0 & audited release & \href{https://huggingface.co/abacusai/Smaug-Decisions}{Link}\\
nischay-1 \citep{ajeet2026nischay1} & model & apache-2.0 & audited release & \href{https://huggingface.co/Ajeet-001/nischay-1}{Link}\\
linne-sprout \citep{pnpm2026linnesprout} & model & apache-2.0 & audited release & \href{https://huggingface.co/PNPM244/linne-sprout-1-0}{Link}\\
cue-reflex \citep{oidlabs2026cuereflex} & model & apache-2.0 & audited release & \href{https://huggingface.co/oidlabs/cue-reflex-4B}{Link}\\
basal-omni \citep{kinas2026basalomni} & model & apache-2.0 & audited release & \href{https://huggingface.co/Remek/basal-omni-1.0-mini}{Link}\\
nirnaya-flash \citep{darkrider2026nirnayaflash} & model & apache-2.0 & audited release & \href{https://huggingface.co/Darkrider/nirnaya-flash-1.0}{Link}\\
jez-large \citep{pragmaticfrontier2026jezlarge} & model & apache-2.0 & audited release & \href{https://huggingface.co/pragmaticfrontier/jez-large-1.0.0}{Link}\\
\midrule
\grouprow{5}{Releases run but not analysed}{}
Decision-1-E4B \citep{worthify2026decision1e4b} & model & apache-2.0 & run, not analysed & \href{https://huggingface.co/Worthify/Decision-1-E4B}{Link}\\
typed-decisions-de \citep{johannhartmann2026typeddecisionsde} & model & apache-2.0 & run, not analysed & \href{https://huggingface.co/johannhartmann/gemma-4-E4B-it-unsloth-bnb-4bit_qlora_typed-decisions-de}{Link}\\
classone-E4B \citep{gonzaga2026classonee4b} & model & apache-2.0 & run, not analysed & \href{https://huggingface.co/devops-thiago/classone-gemma4-e4b}{Link}\\
Avrik-Lite \citep{avrik2026avriklite} & model & apache-2.0 & run, not analysed & \href{https://huggingface.co/avrik-ai/Avrik-Lite-4B}{Link}\\
Jev-Style-2B-v3 \citep{yan2026jevstyle2bv3} & model & apache-2.0 & run, not analysed & \href{https://huggingface.co/chaoliangUNSW/Jev-Style-2B-Decision-v3}{Link}\\
Jev-Style-2B-v4 \citep{yan2026jevstyle2bv4} & model & apache-2.0 & run, not analysed & \href{https://huggingface.co/chaoliangUNSW/Jev-Style-2B-Decision-v4}{Link}\\
Jev-Style-2B-v4.2 \citep{yan2026jevstyle2bv42} & model & apache-2.0 & run, not analysed & \href{https://huggingface.co/chaoliangUNSW/Jev-Style-2B-Decision-v4.2}{Link}\\
\midrule
\grouprow{5}{Non-decision fine-tunes}{}
3D-Coder \citep{ccore20263dcoder} & model & apache-2.0 & non-decision control & \href{https://huggingface.co/ccore/Qwen3.5-4B-3D-Coder}{Link}\\
Abliterated-4B \citep{huihuiai2026abliterated4b} & model & apache-2.0 & non-decision control & \href{https://huggingface.co/huihui-ai/Huihui-Qwen3.5-4B-abliterated}{Link}\\
Distill-4B \citep{xtoolsat2026distill4b} & model & apache-2.0 & non-decision control & \href{https://huggingface.co/xtools-at/DeepSeek-V4-Pro-Qwen3.5-4B-UNCENSORED-heretic}{Link}\\
Super-Coder \citep{jica2026supercoder} & model & none stated$^\ast$ & non-decision control & \href{https://huggingface.co/jica98/qwen3.5-4B-super-coder}{Link}\\
SQL-Distill-2B \citep{leuchtmann2026sqldistill2b} & model & apache-2.0 & non-decision control & \href{https://huggingface.co/AlioLeuchtmann/qwen-3.5-sql-27B-Distill-2B}{Link}\\
MedVL-2B \citep{openmed2026medvl2b} & model & apache-2.0 & non-decision control & \href{https://huggingface.co/OpenMed/Qwen3.5-2B-MedVL}{Link}\\
Abliterated-2B \citep{huihuiai2026abliterated2b} & model & apache-2.0 & non-decision control & \href{https://huggingface.co/huihui-ai/Huihui-Qwen3.5-2B-abliterated}{Link}\\
Distill-2B \citep{emperoai2026distill2b} & model & apache-2.0 & non-decision control & \href{https://huggingface.co/empero-ai/Qwen3.8-2B-Distill}{Link}\\
Bolt-E4B \citep{nvidia2026bolt} & model & openmdw-1.1 & non-decision control & \href{https://huggingface.co/nvidia/NVIDIA-Nemotron-3.5-Bolt-E4B-BF16}{Link}\\
ContextPilot-E4B \citep{pan2026contextpilot} & model & other & non-decision control & \href{https://huggingface.co/tencent/ContextPilot-E4B}{Link}\\
Arabic-E4B \citep{asif2026arabice4b} & model & apache-2.0 & non-decision control & \href{https://huggingface.co/noman-asif/Gemma-4-E4B-Arabic-It}{Link}\\
Abliterated-E4B \citep{huihuiai2026abliteratede4b} & model & apache-2.0 & non-decision control & \href{https://huggingface.co/huihui-ai/Huihui-gemma-4-E4B-it-abliterated}{Link}\\
\midrule
\grouprow{5}{Data and benchmarks}{}
Decision Index (public suite 0.3) \citep{decisionindex} & benchmark & MIT (kit) & evaluation only & \href{https://github.com/apolinario/decision-index}{Link}\\
JevBench \citep{jevbench} & benchmark & MIT & evaluation only & \href{https://github.com/fstandhartinger/jevbench}{Link}\\
tasksource typed decisions \citep{sileo2023tasksource} & dataset & other$^\dagger$ & training data of our adapters; overlap scan & \href{https://huggingface.co/datasets/tasksource/tasksource-jev-typed-decisions}{Link}\\
Private two-task set & dataset & private & evaluation only; no item shown & --\\
\bottomrule
\end{longtable}
\endgroup

\end{document}